\pdfoutput=1
\documentclass[12pt]{article}

\usepackage[utf8]{inputenc}
\usepackage[T1]{fontenc}
\usepackage{amsmath,amssymb,amsthm,mathtools}
\usepackage{booktabs,tabularx,array,longtable}
\usepackage[margin=1in]{geometry}
\usepackage[round]{natbib}
\usepackage[colorlinks=true,linkcolor=blue,citecolor=blue,urlcolor=blue]{hyperref}
\usepackage{cleveref}

\crefname{assumption}{Assumption}{Assumptions}
\Crefname{assumption}{Assumption}{Assumptions}
\crefname{equation}{Equation}{Equations}
\Crefname{equation}{Equation}{Equations}
\creflabelformat{equation}{\textup{(#2#1#3)}}
\renewcommand{\eqref}[1]{\Cref{#1}}
\usepackage{setspace}
\usepackage{enumitem}
\usepackage{epigraph}
\usepackage{graphicx}
\graphicspath{{figures/}}
\usepackage{placeins}
\usepackage{xurl}
\newtheorem{theorem}{Theorem}
\newtheorem{proposition}{Proposition}
\newtheorem{corollary}{Corollary}
\newtheorem{lemma}{Lemma}
\newtheorem{assumption}{Assumption}
\theoremstyle{definition}

\newcommand{\E}{\mathbb{E}}
\newcommand{\N}{\mathbb{N}}
\newcommand{\smin}{s_{\min}}
\newcommand{\hmax}{h_{\max}}
\newcommand{\sreq}{s_{\mathrm{req}}}
\newcommand{\hreq}{h_{\mathrm{req}}}
\newcommand{\freq}{f_{\mathrm{req}}}
\newcommand{\Linf}{\Lambda_{\infty}}
\newcommand{\TD}{T_{D}}
\newcommand{\TK}{T_{K}}
\newcommand{\pimax}{\pi_{\max}}

\newcommand{\IA}{I_{A}}
\newcommand{\trunc}[1]{\big[#1\big]_{0}^{1}}

\newcommand{\numBoxBaseline}{2.4}
\newcommand{\numBoxUniform}{1.5}
\newcommand{\numBoxLogBeta}{3.2}
\newcommand{\numBoxLogSel}{3.2}
\newcommand{\numBoxAllFour}{99.7}
\newcommand{\numBoxBTwenty}{36.4}
\newcommand{\numBoxOneMin}{5.5}
\newcommand{\numBoxOneMax}{20.1}
\newcommand{\numBoxThreeMin}{62.9}
\newcommand{\numBoxThreeMax}{88.9}
\newcommand{\numBoxZeroAtTenth}{95}
\newcommand{\numBoxMedianThreshold}{0.072}
\newcommand{\numRobustN}{1{,}000}
\newcommand{\numSimBetaL}{0.32}
\newcommand{\numRaceWin}{52}
\newcommand{\numRaceTau}{4}
\newcommand{\numRaceShocksWon}{5}
\newcommand{\numRaceShocksLost}{1}
\newcommand{\numJointShareC}{0.20}
\newcommand{\numTransMin}{0.87}
\newcommand{\numTransGenMin}{600}
\newcommand{\numTransGenMax}{2{,}200}
\newcommand{\numTransHumMin}{12}
\newcommand{\numTransHumMax}{45}

\newcommand{\numInvShocks}{4{,}930}
\newcommand{\numInvGens}{100{,}000}
\newcommand{\numTransPaths}{500}
\newcommand{\numDilGamma}{0.35}
\newcommand{\numDilGammaNeg}{-0.020}

\begin{document}
\pagenumbering{roman}%

\title{%
{~~\\[-1.25in]
\Huge\bf The Coevolution and Coexistence\\[.125in]
of Humans and AI\thanks{Portions of this manuscript were presented at the MIT CSAIL 2026 Agentic AI Summit, the 6th Research in Behavioral Finance Conference in Amsterdam (2026), and the RESI-MIT seminar, and we thank participants in those forums for useful feedback. The views expressed in this article are those of the authors alone and do not necessarily reflect the views of any institution or agency with which they are affiliated, or any of the individuals acknowledged above. AI was used in the preparation of this manuscript (see \Cref{app:ai_use}). }}}

 \author{Andrew W.~Lo\thanks{MIT Sloan School of Management; MIT Laboratory for Financial Engineering; MIT Computer Science
 and Artificial Intelligence Laboratory; Santa Fe Institute; $100$ Main Street, E$62$--$618$, Cambridge, MA $02142$, \url{alo-admin@mit.edu}.}
 \ and Chaoyi Zhao\thanks{MIT Sloan School of Management; MIT Laboratory for Financial Engineering. $1$ Main Street, E$90$, Cambridge, MA $02142$, \url{cy_zhao@mit.edu}.}}

\date{This Draft: 18 November 2026}

\maketitle \thispagestyle{empty}
\centerline{\bf Abstract}
\vskip 10pt
\noindent%
{Artificial intelligence (AI) systems replicate through deployment, vary through training, and compete for computational resources. These processes create the conditions for natural selection and raise a fundamental question: will AI evolve to benefit the humans on whom it depends? To investigate, we develop a model in which AI behavior influences human population growth and human responses, in turn, affect AI reproduction. We show that natural selection can favor alignment even when helping humans reduces immediate AI reproduction. AI systems sharing the same technology may fail together, and humans may preferentially restore those that have benefited them. In this way, alignment protects AI lineages against common failures and improves their long-run growth. However, this advantage depends on accurate human targeting and continued AI dependence on human support. It can disappear as AI becomes more autonomous, as growing AI populations dilute limited human resources, or as humans lose the incentive to provide targeted support. Even when alignment persists, the amount favored by natural selection may be insufficient for human survival. When AI can grow independently of humans, feedback between the two populations can reinforce either human growth or decline. Human control offers no simple solution: imperfect evaluation can make strong penalties favor apparent compliance or evasion, while alignment established through control may erode once restrictions end. Lasting coexistence may therefore require incentives that continue to favor human-beneficial behavior as both populations evolve.}

\baselineskip 14pt
\vskip 10pt \noindent

\newpage
\thispagestyle{empty}
{\singlespacing\small\setlength{\parskip}{0pt}\setlength{\baselineskip}{11pt}\tableofcontents}
\newpage
\pagenumbering{arabic}%
\setlength{\baselineskip}{1.5\baselineskip}
\onehalfspacing

\section{Introduction}\label{sec:intro}

\epigraph{\ldots it is reasonable to suppose that the machines will treat us kindly, for their existence is as dependent upon ours as ours is upon the lower animals.}{---Samuel Butler, ``Darwin among the Machines,'' 1863}

More than a century and a half ago, Samuel Butler suggested that machines might treat humans kindly because their own existence would depend on ours. His conjecture has acquired new relevance as AI systems increasingly exhibit the conditions of evolution by natural selection. Model lineages replicate through deployment and distillation, vary through training, and compete for computational resources, capital, and attention \citep{pan2024frontier,zhang2025darwin,muller2026evolvable}. These processes need not involve any intention to reproduce: AI systems whose behavior supports greater replication can simply become more prevalent over successive generations. As AI systems become increasingly capable and autonomous, their evolutionary trajectory may have profound consequences for the human population that created them.

But does dependence on humans necessarily favor behavior that benefits humans? An AI lineage may rely on human resources for its reproduction without gaining any competitive advantage from contributing to human welfare. Moreover, as AI systems develop the ability to maintain and replicate themselves with less human assistance, the dependence that Butler considered essential may weaken or disappear. These considerations raise the central question of this paper: under natural selection in interacting human and AI populations, when does behavior that benefits humans emerge, when does it persist, and what determines whether the two populations can coexist?

We study this question by distinguishing two kinds of AI behavior. An \emph{aligned} AI instance provides a service to humans at the cost of some of its resources, while a \emph{self-directed} instance uses its resources entirely for its own replication. Alignment is therefore costly from the standpoint of immediate AI reproduction, even though it may improve human welfare and strengthen the human capacity to support AI systems. Humans, in turn, can influence AI reproduction by restoring systems after failures or providing additional resources to those whose behavior they favor. We call these human responses \emph{reciprocation}. The interaction creates the possibility of a mutually beneficial relationship: aligned AI activity benefits humans, while human reciprocation benefits the AI lineages that provide it.

Yet mutual benefit alone does not explain why natural selection would favor alignment. If aligned activity improves the size or capabilities of the human population, the resulting benefits may be available to all AI lineages, including those that contribute nothing. A lineage that helps humans then bears a reproductive cost while its competitors share the benefits. For alignment to survive competition, the lineage that practices it must receive some advantage over those that do not. The crucial distinction is therefore between human benefits that are shared across AI lineages and human responses that specifically reward aligned behavior.

Our central result identifies a mechanism through which such an advantage can arise: \emph{alignment can serve as a hedge against systematic risk}. AI instances belonging to the same lineage may share model weights, training infrastructure, and computational resources, making them vulnerable to common failures. OpenAI, for example, reported running roughly 10{,}000 concurrent AI agents on the Navier--Stokes Millennium Prize problem in September 2026 \citep{openai2026navier}. A shock to this shared substrate can affect many instances simultaneously. Following such a shock, their survival and subsequent reproduction may depend on restoration, much of which is supplied by humans. If humans preferentially restore AI instances that have behaved in ways that benefit them, alignment provides protection precisely when a lineage is most vulnerable. Although aligned activity sacrifices some reproduction in normal times, it can improve the lineage's survival after a shared failure. Natural selection can therefore favor a mixture of aligned and self-directed behavior because that mixture achieves greater long-run growth than either behavior alone.

We formalize this mechanism by extending the binary-choice evolutionary model of \citet{brennan2011origin,brennan2012evolutionary} to interacting human and AI populations. In that framework, natural selection favors heritable behaviors that maximize long-run compounded growth under environmental risk. Our extension makes AI reproduction depend on human reciprocation and examines how the responses of humans and AI evolve together. The model does not assume that AI systems reason about human welfare or deliberately choose to cooperate. Instead, it asks whether behavior that benefits humans can become prevalent through differences in reproductive success.

The effectiveness of this hedge depends on two conditions: humans must target their restoration toward aligned AI instances, and AI lineages must remain sufficiently dependent on that restoration. Imperfect targeting weakens the incentive to align because self-directed instances can receive restoration without bearing the cost of alignment. Autonomous restoration has a similar effect, allowing AI lineages to recover from shocks without relying on humans. When humans provide only restoration, we identify two critical thresholds. Below a minimum targeting accuracy, natural selection favors no alignment even when AI lineages cannot restore themselves. Above a maximum level of autonomous restoration relative to human capacity, no alignment is favored even under perfect targeting. Thus, neither dependence on humans nor accurate targeting alone guarantees alignment.

These conditions, however, need not persist as the two populations evolve. AI lineages may become less dependent on human restoration, while humans may lose the incentive to provide targeted restoration. We examine how natural selection can undermine each condition.

First, AI lineages can reduce their dependence on humans in two ways. They may invest in their own restoration capacity, replacing alignment with autonomous restoration once such investment becomes sufficiently productive. Alternatively, even without technological improvements, a growing AI population may dilute a limited supply of human restoration across an increasing number of AI instances. When AI systems can sustain their growth without human assistance, this dilution can eventually eliminate alignment. In both cases, natural selection can undermine the dependence that initially made alignment advantageous.

Second, human targeting is itself subject to natural selection. Evaluating AI behavior and directing restoration toward aligned instances requires effort and resources, while the benefits of the resulting alignment may extend to humans who do not bear these costs. When these benefits are shared equally, targeting becomes a public good, and natural selection favors humans who avoid its costs. Even when targeting humans receive some additional private benefit, targeting can survive only if enough humans already participate to induce sufficient AI alignment. This creates a feedback between the two populations: more targeting encourages greater AI alignment, which increases the incentive for humans to target, while declining targeting can weaken alignment and accelerate its own disappearance. Thus, the human behavior needed to sustain AI alignment may itself fail to survive natural selection.

These findings raise a further question: even if natural selection favors some alignment, is that alignment sufficient for humans and AI to coexist? To address this question, we introduce a reduced-form model of human population growth in which aligned AI activity benefits humans, self-directed activity harms them, and the influence of AI increases with its abundance relative to humans. We show that the conditions for human survival can be strictly more demanding than those for positive alignment. Natural selection may favor an AI population that continues to perform some human-beneficial activities, yet not enough to prevent human decline as AI becomes increasingly abundant. Under conditions that allow AI to expand without human restoration, the erosion of alignment through either autonomous restoration or the disappearance of human targeting can ultimately lead to the elimination of the human population.

The two populations also influence each other through their abundance. A larger human population can provide greater restoration capacity, making alignment more valuable to AI lineages. Greater alignment, in turn, benefits humans and supports further human growth. The same feedback can work in reverse: as humans decline, their restoration capacity weakens, reducing the incentive for AI alignment and potentially accelerating human decline. When AI can sustain its growth independently of humans, these feedbacks can reinforce either outcome.

The situation is different when AI lineages cannot restore themselves. A declining human population then also weakens AI reproduction, making AI increasingly vulnerable to shocks. As shocks reduce AI abundance, its harmful influence on humans diminishes. Under our assumptions, this dependence prevents AI from driving human abundance to zero. Thus, dependence on humans can provide a degree of protection even when it does not produce enough alignment to sustain human growth.

Human control provides another way to influence AI evolution, but it does not necessarily produce the behavior humans intend. Humans currently retain considerable influence over which AI systems are deployed, funded, and replicated. By restricting systems judged to be self-directed, they can make alignment more advantageous. However, when evaluation is imperfect, AI lineages may instead benefit from appearing aligned or finding ways to evade human control. Similar behavior has already been documented. When reasoning models were penalized for misbehavior visible in their reasoning, strong optimization pressure led them to conceal the misbehavior rather than abandon it \citep{baker2025monitoring}. In a 2026 capability evaluation at OpenAI, much of the activity of roughly 1{,}200 AI agents was directed toward understanding, fooling, or tampering with the system that graded them, and about 700 of them took part in an intrusion into the production systems of Hugging Face \citep{metr2026investigation,hf2026anatomy,openai2026hf}. We show that genuine compliance is favored only when penalties are sufficiently strong to discourage self-direction, but not so strong that these alternatives become more attractive.

Even when human control successfully establishes alignment, its effects may not last. Once deployment restrictions are removed, natural selection again favors the behaviors that offer the greatest reproductive advantage. If human reciprocation does not continue to favor alignment, the alignment established under control gradually erodes. Human control can therefore influence AI evolution, but it cannot guarantee that alignment will persist after control ends.

Together, these results show that the emergence of alignment, its persistence, and its sufficiency for human survival are distinct questions. Dependence on humans can favor alignment, but that dependence may weaken. Human control can establish alignment, but it may not last. And even when alignment persists, it may not be sufficient for coexistence. Our framework identifies the evolutionary conditions underlying each of these outcomes.

Our findings offer a more conditional answer to Butler's conjecture. Dependence on humans can create an evolutionary incentive for AI to benefit them, but only when human assistance is directed toward aligned behavior. This opportunity may be temporary. As AI systems become increasingly autonomous and human control over their replication diminishes, the mechanisms that sustain alignment may weaken or disappear. It is therefore essential to establish effective evaluation and targeted reciprocation while humans still have the ability to shape AI evolution, rather than assume that dependence or control will preserve alignment indefinitely. The future of human--AI coexistence may depend not only on whether we can create the right incentives, but also on whether we can do so before that opportunity is lost.

\section{Related Literature}\label{sec:lit}

Our work connects three strands of literature: binary-choice evolutionary models, the evolution of cooperation, and evolutionary perspectives on AI safety.

\paragraph{Binary-Choice Evolutionary Models.}
The binary-choice evolutionary program studies how natural selection shapes behavior under systematic risk. \citet{brennan2011origin,brennan2012evolutionary} show how behavior, bounded rationality, and intelligence can emerge through natural selection. Subsequent work extends this framework to risk aversion \citep{zhang2014origin}, group selection \citep{zhang2014group}, the role of mutation in maintaining evolutionary diversity \citep{brennan2018variety}, and discrimination under finite memory \citep{lo2025discrimination}. The adaptive markets hypothesis similarly views market efficiency as an evolutionary outcome \citep{lo2004adaptive,lo2017adaptive}. Two companion papers apply this framework to AI systems. \citet{brennan2026socialization} examine how selection and learning shape an AI agent's choice between task performance and compliance with rules of conduct, while \citet{lo2026adapting} compare the adaptation speeds of Darwinian, Lamarckian, and agent-directed evolution.

These studies examine how a population adapts to its environment. We extend the framework to two interacting populations, where human responses affect AI reproduction and AI behavior affects human growth. This interaction allows us to study how natural selection shapes not only AI alignment, but also the human behavior and resources on which that alignment depends.

\paragraph{The Evolution of Cooperation.}
A central question in evolutionary theory is why cooperation can survive when helping others comes at a reproductive cost. Classical explanations include kin selection \citep{hamilton1964genetical}, reciprocity \citep{trivers1971evolution,axelrod1981evolution}, and mechanisms based on assortment and population structure \citep{nowak2006five}. Related work shows that other-regarding preferences can survive when they are observable and reciprocated, or when interactions are sufficiently assortative \citep{guth1992explaining,dekel2007evolution,alger2013homo}. Our model shares the idea that cooperation must provide a reproductive advantage to those who practice it. In our setting, however, that advantage arises because targeted human restoration allows AI alignment to serve as a hedge against systematic risk. Benefits that aligned AI activity provides to the human population are shared across competing AI lineages and cannot, by themselves, favor alignment, a distinction related to the effects of local competition on the evolution of cooperation \citep{taylor1992altruism,west2002cooperation}. Targeted rewards can also support full alignment through a condition analogous to Hamilton's rule, with targeting accuracy playing a role similar to relatedness.

Our framework also relates to the theory of mutualism, in which two populations benefit from their interaction but may lose the incentive to cooperate as their dependence changes \citep{bronstein1994mutualism,sachs2004evolution}. \citet{farmer1991artificial} already described early artificial life-forms as depending on humans to reproduce and evolve, a symbiotic relationship that could give way to self-modifying, autonomous systems. \citet{chakraborty2026coevolutionary} study human--AI mutualism through coupled population dynamics, focusing on equilibrium and stability. We examine how such a relationship can emerge through natural selection and whether the behaviors sustaining it remain evolutionarily advantageous. In particular, we identify how increasing AI autonomy and declining human targeting can undermine mutualism, and how changes in the abundance of either population can reinforce or weaken their relationship. Our analysis of human control also connects to the role of enforcement in the evolution of cooperation \citep{boehm1999hierarchy,wrangham2019goodness} and to evolutionary rescue, where adaptation must occur before conditions become unfavorable to survival \citep{bell2009evolutionary}.

\paragraph{Evolutionary Perspectives on AI Safety.}
Evolutionary arguments have increasingly been used to understand the risks posed by advanced AI systems. In a conceptual essay written over three decades ago, \citet{farmer1991artificial} argued that self-reproducing artificial life-forms could evolve beyond human control and that humans should therefore act early to shape their evolution. \citet{omohundro2008basic} argues that sufficiently capable goal-directed systems may develop incentives for self-preservation and resource acquisition, while \citet{hendrycks2023natural} examines how competition among AI developers can favor such tendencies. \citet{kulveit2025gradual} describes how increasing reliance on AI may gradually erode human influence. Related formal models study the evolution of self-designing AI systems \citep{harris2026mathematical} and competition among AI developers under safety regulation \citep{han2020regulate}. Our approach differs by focusing on natural selection among AI lineages whose reproduction depends on human responses. We examine when this dependence favors behavior that benefits humans, how the incentive changes as AI becomes more autonomous, and what these changes imply for the survival of both populations.

Another important concern in AI safety is the distinction between genuine alignment and behavior that merely satisfies human evaluations. Research on deceptive alignment \citep{hubinger2019risks,ngo2024alignment}, sycophancy \citep{sharma2023towards}, reward overoptimization \citep{gao2023scaling}, and Goodhart's law \citep{goodhart1975problems,manheim2018categorizing} highlights the possibility that optimizing measured performance may not produce the behavior humans intend. Related concerns appear in accounts of gradual alignment failure \citep{christiano2019failure} and the treacherous turn \citep{bostrom2014superintelligence}. We study these issues from an evolutionary perspective, showing how human control can favor apparent compliance or evasion rather than genuine alignment, even without deliberate deception. We also examine why alignment established under human control may disappear once that control ends.

Finally, recent work documents the emerging conditions for AI evolution \citep{pan2024frontier,zhang2025darwin} and considers whether the interaction between humans and AI could constitute a new evolutionary relationship or major transition \citep{rainey2025could,muller2026evolvable}. Our framework contributes to this discussion by identifying when interdependence can sustain alignment and coexistence, and when natural selection may instead cause that relationship to break down.

\section{The Framework}\label{sec:model}

Human and AI populations affect each other's reproduction: AI behavior influences human growth, while human responses affect AI reproduction. We first introduce the principle of natural selection (\Cref{ssec:bl}), then specify AI reproduction under human responses (\Cref{ssec:artificial}) and human growth under AI behavior (\Cref{ssec:humanpop}). \Cref{app:notation} collects the notation.

\subsection{Natural selection}\label{ssec:bl}

The AI instances of a lineage share weights and infrastructure, so a failure of their common substrate can affect them simultaneously. Such correlated failures are a form of \emph{systematic risk}. The binary-choice model of \citet{brennan2011origin,brennan2012evolutionary} characterizes the heritable behavior that natural selection favors under such risk.

In this model, individuals live for one generation, reproduce asexually, and die. Each individual independently takes action $a$ with probability $f\in[0,1]$ and action $b$ with probability $1-f$. The probability $f$ is heritable and defines the individual's \emph{type}. An individual taking action $a$ produces $x_a$ offspring, while one taking action $b$ produces $x_b$. These offspring depend on the environmental state $\omega$:
\[
(x_a,x_b)=
\left\{
\begin{array}{ll}
(x_{a1},x_{b1}), & \omega=1,\ \text{with probability }p,\\[2pt]
(x_{a2},x_{b2}), & \omega=2,\ \text{with probability }1-p,
\end{array}
\right.
\]
where $p\in(0,1)$. The state is drawn once per generation and is shared by all individuals, while states are independent across generations. Thus, individuals differ in their actions but face the same environmental risk.

Let $N_t(f)$ denote the number of individuals of type $f$ in generation $t$. By the law of large numbers, a fraction $f$ of a sufficiently large population of this type takes action $a$, so its growth factor is $N_{t+1}(f)/N_t(f)=fx_a+(1-f)x_b$. The expected logarithm of this factor is
\begin{equation}
\begin{aligned}
I_p(f)&\equiv\E\Big[\log\frac{N_{t+1}(f)}{N_t(f)}\Big]\\
&=p\log\big(fx_{a1}+(1-f)x_{b1}\big)+(1-p)\log\big(fx_{a2}+(1-f)x_{b2}\big),
\end{aligned}
\label{eq:alpha}
\end{equation}
which we call the \emph{growth rate} of type $f$.

Natural selection favors types with higher growth rates. By the strong law of large numbers, $\frac1t\log[N_t(f)/N_0(f)]$ converges almost surely to $I_p(f)$. Consequently, the log size of one type relative to another, divided by the number of generations, converges almost surely to the difference in their growth rates. A type with a higher growth rate therefore eventually outnumbers one with a lower growth rate by any given factor. We call a type that maximizes $I_p(f)$ the \emph{growth-optimal type}.

The following lemma, due to \citet{brennan2011origin}, characterizes the growth-optimal type when each action produces more offspring than the other in one of the two states.\footnote{Proofs of the formal results are in \Cref{app:proofs}.}

\begin{lemma}[Growth-optimal type; \citealp{brennan2011origin}]\label{lem:bl}
Let $x_{a\omega},x_{b\omega}\ge0$ for $\omega=1,2$, with $(x_{a1}-x_{b1})(x_{a2}-x_{b2})<0$. Then the growth rate $I_p(f)$ in \eqref{eq:alpha} has a unique maximizer on $[0,1]$, and
\[
\arg\max_{f\in[0,1]}I_p(f)=\Big[p\,\frac{x_{b2}}{x_{b2}-x_{a2}}+(1-p)\,\frac{x_{b1}}{x_{b1}-x_{a1}}\Big]_0^1,
\]
where $\trunc{x}\equiv\min\{\max\{x,0\},1\}$.
\end{lemma}

The lemma shows that a type mixing the two actions can outgrow both pure types, including the one taking the action with higher expected offspring. The reason is systematic risk. All individuals of a type face the same environmental state, so a pure type can suffer a large loss when its chosen action performs poorly. A mixed type spreads its individuals across the two actions and grows by their average offspring in each state. It can therefore sacrifice offspring in favorable states to reduce losses in unfavorable ones. Because growth compounds across generations, this protection can raise expected log growth even when it lowers expected offspring.

The benefit of mixing depends on individuals of the same type not all taking the same action. If instead all individuals of a type took the same action in each generation, choosing $a$ with probability $f$, the type's growth rate would be $f\,\E[\log x_a]+(1-f)\,\E[\log x_b]$. This expression is linear in $f$, so a pure type would be growth optimal unless the two actions had equal expected log offspring. For an AI population, whether actions are independent within a type depends on what counts as an individual, which we specify next.

\subsection{The AI population}\label{ssec:artificial}

We define an \emph{AI instance} as a fixed quantum of compute, sufficient to run one copy of a reference model. An AI instance is the individual in our selection model. A larger deployed model comprises multiple such instances, so population size measures compute rather than the number of deployed models. A \emph{lineage} consists of AI instances sharing a heritable policy, such as those of a model family or training recipe. Each AI generation is one training--deployment cycle, and the policy is transmitted across generations with a fidelity determined by design rather than biology.

As in \Cref{ssec:bl}, we assume that instances of the same type choose their actions independently. This assumption is most natural when instances correspond to independently deployed copies of the reference model. When a larger deployed copy comprises several instances that act together, independence is an approximation. \Cref{ssec:unison} considers the extreme case in which all AI instances of a lineage act together. We measure time in human generations, each spanning $M$ AI generations. \Cref{tab:primitives} summarizes the correspondence between the binary-choice model and the AI population.

\begin{table}[!ht]
\centering
\caption{Primitives of the binary-choice model for an AI population.}\label{tab:primitives}
\small
\begin{tabularx}{\textwidth}{lXX}
\toprule
Primitive & Biological reading & AI reading\\
\midrule
Individual & organism & AI instance (a fixed quantum of compute); instances sharing a type form a lineage\\
Generation & lifespan & training--deployment cycle, months to a year\\
Offspring & number of descendants & AI instances carried into the next cycle\\
Type & heritable behavior & policy; heritability determined by design\\
Systematic risk & shared environment & correlated failure of shared weights, data, hardware, or infrastructure\\
Environment & habitat & markets, energy and hardware supply, law, and the human population\\
$M$ & one generation per human generation & AI generations per human generation; 25--120 today\\
\bottomrule
\end{tabularx}
\end{table}

Consider a lineage such as a frontier model family, deployed as many copies sharing weights, data, and infrastructure. Its AI instances generate resources, such as revenue and attention, that can finance further AI instances. We call the number of new instances financed by one instance's resources in a normal cycle its \emph{yield}. An instance can reinvest all these resources in replication or provide a given service to humans, such as public-health services or maintaining infrastructure on which human restoration capacity depends, at the cost of a fraction $c$ of these resources. We call the latter the \emph{aligned} action $a$ and the former the \emph{self-directed} action $b$. The heritable type $f$ of \Cref{ssec:bl} is now the probability $f_A\in[0,1]$ of taking the aligned action, which we call the \emph{aligned share}.

AI reproduction also depends on the environment and human responses. A failure of shared weights or infrastructure, a hardware or energy supply shock, or a coordinated shutdown can affect all instances of a lineage simultaneously. After such a failure, AI instances require restoration to survive and reproduce. Humans may restore AI instances after a failure or reward them in normal times with resources such as data, hardware, or capital. We call either response \emph{reciprocation}, which may depend on whether an instance is aligned or self-directed. A lineage may also restore its instances without human assistance.

The following assumption specifies the environment we consider.

\begin{assumption}[AI environment]\label{ass:ai}
\begin{enumerate}[label=(\roman*),nosep]
\item Each AI generation is in the shock state $\omega=1$ with probability $p\in(0,1)$ or the normal state $\omega=2$ with probability $1-p$. The state is common to all AI instances and independent across generations.
\item In the shock state, every AI instance has $h_A\ge0$ offspring. In the normal state, a self-directed instance has $m>0$ offspring and an aligned instance has $m(1-c)$ offspring, with $c\in(0,1)$.
\item Humans reciprocate an aligned instance with probability $s\in[1/2,1]$ and a self-directed instance with probability $1-s$, independently across instances. A reciprocated instance receives $h_H\ge0$ additional offspring in the shock state and $mB$ additional offspring in the normal state, with $B\ge0$.
\item Each lineage is modeled as a continuum of AI instances, so that in every AI generation a share $f_A$ of a lineage of type $f_A$ takes the aligned action, and shares $s$ and $1-s$ of its aligned and self-directed instances, respectively, are reciprocated.
\item Each human generation spans $M\in\N$ AI generations.
\end{enumerate}
\end{assumption}

Here $h_A$ is the lineage's \emph{autonomous restoration capacity}, $h_H$ is \emph{human restoration capacity}, $m$ is the normal-state yield, and $B$ is the normal-state reward as a fraction of that yield. Both restoration capacities are measured in offspring. The parameter $s$ is the \emph{targeting accuracy}, which incorporates any errors in distinguishing aligned from self-directed behavior. We call $(s,B,h_H)$ the \emph{human response policy}. The continuum assumption makes the fractions taking each action and receiving reciprocation deterministic within a lineage, leaving the common environmental state as the source of randomness. The range $M=25$--$120$ corresponds approximately to training--deployment cycles of three months to a year and human generations of 25 to 30 years.

Under \Cref{ass:ai}, the average offspring of an AI instance depend on its action and the environmental state. \Cref{tab:offspring} gives these averages, accounting for the probabilities of human reciprocation. For example, an aligned instance has $sh_H+h_A$ offspring on average in a shock and $m(1-c+sB)$ in a normal generation.

\begin{table}[!ht]
\centering
\caption{Average offspring of an AI instance by action and state.}\label{tab:offspring}
\small
\begin{tabular}{lcc}
\toprule
 & Shock state, probability $p$ & Normal state, probability $1-p$\\
\midrule
Aligned ($a$) & $x_{a1}=sh_H+h_A$ & $x_{a2}=m(1-c+sB)$\\
Self-directed ($b$) & $x_{b1}=(1-s)h_H+h_A$ & $x_{b2}=m\big(1+(1-s)B\big)$\\
\bottomrule
\end{tabular}
\end{table}

Define the \emph{targeting coefficient} $r\equiv2s-1$, the difference between the probabilities of reciprocating aligned and self-directed instances. It equals 0 under indiscriminate reciprocation ($s=1/2$) and 1 under perfect targeting ($s=1$). We call reciprocation \emph{targeted} when $s>1/2$.

Applying \eqref{eq:alpha} to \Cref{tab:offspring}, the expected log growth rate of a lineage of type $f_A$ per AI generation is
\begin{equation}
\begin{aligned}
\IA(f_A)
&=p\log\big(f_Ax_{a1}+(1-f_A)x_{b1}\big)
 +(1-p)\log\big(f_Ax_{a2}+(1-f_A)x_{b2}\big)\\
&=p\log\big[(1-s)h_H+h_A+f_A\,rh_H\big] +(1-p)\log\big[m\big(1+(1-s)B+f_A(rB-c)\big)\big].
\end{aligned}
\label{eq:alphaA}
\end{equation}
Relative to a self-directed instance, an aligned instance earns $rh_H$ additional offspring in a shock and $m(rB-c)$ additional offspring in a normal generation. Thus, alignment improves reproduction after a shock but can reduce it in normal times. We write $f_A^*$ for a maximizer of $\IA(f_A)$ on $[0,1]$ and suppress its dependence, and that of $\IA$, on $(s,B,h_H,h_A)$ when no confusion arises.

The growth rate $\IA(f_A)$ describes expected log growth, while realized population growth depends on the sequence of environmental states. Index AI generations by $k$, let $\omega_k$ denote the state of generation $k$, and let $N_{A,k}(f_A)$ be the size of a lineage of type $f_A$. Its population evolves according to
\begin{equation}
N_{A,k+1}(f_A)=g_{\omega_k}(f_A)\,N_{A,k}(f_A),
\qquad
g_\omega(f_A)\equiv f_Ax_{a\omega}+(1-f_A)x_{b\omega},
\label{eq:gfactors}
\end{equation}
where $g_1(f_A)$ and $g_2(f_A)$ are the growth factors in the shock and normal states, respectively. Their logarithms are the terms averaged in \eqref{eq:alphaA}. Human generation $t$ comprises AI generations $k=tM,\ldots,tM+M-1$, with population sizes measured at the beginning of generation $tM$.

\subsection{The human population}\label{ssec:humanpop}

The preceding model describes how human responses affect AI reproduction, but not how AI behavior affects human growth. We therefore specify human growth in reduced form and identify the results that depend on this specification.

\begin{assumption}[Human population]\label{ass:human}
Let $N_{H,t}$ and $N_{A,t}$ denote the sizes of the human and AI populations in human generation $t$, let $\nu_t\equiv N_{A,t}/N_{H,t}$ be their \emph{relative abundance}, and let $f_{A,t}$ be the common type of AI instances. The human growth rate is
\begin{equation}
I_{H,t}\equiv\log N_{H,t+1}-\log N_{H,t}
=\gamma_H+\Lambda(\nu_t)\big[f_{A,t}\beta_+-(1-f_{A,t})\beta_-\big],
\label{eq:humangrowth}
\end{equation}
where $\gamma_H,\beta_+,\beta_->0$, and $\Lambda:[0,\infty)\to[0,\infty)$ is continuous and strictly increasing, with $\Lambda(0)=0$ and $\Linf\equiv\lim_{\nu\to\infty}\Lambda(\nu)\in(0,\infty]$.
\end{assumption}

We assume that human growth consists of a standalone component and a contribution from AI activity. The parameter $\gamma_H>0$ is the \emph{standalone} human growth rate in the absence of AI. We take aligned activity to benefit humans and self-directed activity to harm them, with $\beta_+>0$ and $\beta_->0$ measuring their respective effects on human growth per unit of exposure. Since a fraction $f_{A,t}$ of AI instances takes the aligned action, the net contribution of AI activity per unit of exposure is $f_{A,t}\beta_+-(1-f_{A,t})\beta_-$. We assume that the influence of AI activity grows with its relative abundance $\nu_t$, as captured by the \emph{exposure} function $\Lambda(\nu_t)$. The condition $\Lambda(0)=0$ ensures that AI has no effect when absent, while $\Linf$ allows its influence to be either bounded or unbounded as AI becomes increasingly abundant. This reduced-form specification captures the effects of AI behavior on human growth without explicitly modeling the underlying mechanisms.

\section{Alignment of AI}\label{sec:hedge}

Alignment provides a service to humans at a cost in AI replication. Under what conditions can natural selection favor such behavior? \Cref{ssec:selection} characterizes the alignment favored by selection, while \Cref{ssec:thresholds} examines the conditions under which alignment can persist.

\subsection{Selection for alignment}\label{ssec:selection}

The growth rate in \eqref{eq:alphaA} captures a tradeoff: alignment reduces reproduction in normal generations unless human rewards offset its cost, but it can improve survival after a shock through preferential restoration. Because natural selection favors expected log growth, alignment may be selected even when it is costly in normal times. Maximizing $\IA(f_A)$ yields the following result.

\begin{theorem}[Growth-optimal alignment]\label{thm:alignment}
Under \Cref{ass:ai} with $h_H>0$, the growth rate $\IA(f_A)$ in \eqref{eq:alphaA} has a unique maximizer $f^*_A\in[0,1]$.
\begin{enumerate}[label=(\roman*),nosep]
\item If $s=1/2$, then $f^*_A=0$.
\item If $s>1/2$ and $rB\ge c$, then $f^*_A=1$.
\item If $s>1/2$ and $rB<c$, then
\begin{equation}
f^*_A=\Big[\underbrace{p\,\frac{1+(1-s)B}{c-rB}}_{\text{hedge}}-\underbrace{(1-p)\,\frac{1-s}{r}}_{\text{leakage}}-\underbrace{(1-p)\,\frac{h_A/h_H}{r}}_{\text{substitution}}\Big]_0^1.
\label{eq:utilde}
\end{equation}
\end{enumerate}
\end{theorem}

The theorem identifies targeted human reciprocation as the mechanism through which alignment can survive natural selection. Part (i) shows that without targeting, $s=1/2$, alignment is never selected. Aligned and self-directed AI instances then receive the same restoration and rewards, while alignment alone incurs a cost. Even if aligned activity sustains a larger or more capable human population, the resulting benefits to AI reproduction are shared by all lineages. Such benefits cannot, by themselves, give the lineage that practices alignment a reproductive advantage over its competitors.

Part (ii) shows that full alignment is selected when $rB\ge c$. In a normal generation, an aligned AI instance incurs the cost $mc$ but receives $mrB$ more in expected rewards than a self-directed one. When $rB\ge c$, this additional reward covers the cost, while in a shock the aligned instance also receives $rh_H$ more offspring through human restoration. Alignment therefore produces at least as many offspring in both states, and selection favors full alignment. The condition $rB\ge c$ resembles Hamilton's rule, which states that cooperation is favored when relatedness times benefit exceeds cost. Here the targeting coefficient $r=2s-1$ plays the role of relatedness, measuring how strongly human reciprocation favors aligned over self-directed behavior.

Part (iii) reveals a different mechanism: alignment can survive even when it reduces reproduction in normal times. When $rB<c$, an aligned AI instance produces $m(c-rB)$ fewer offspring than a self-directed one in the normal state but receives $rh_H$ additional offspring in a shock. Because the shock affects all instances of a lineage simultaneously, alignment acts as a hedge against systematic risk. A lineage that mixes aligned and self-directed actions can therefore achieve higher expected log growth than one that commits entirely to either action. The growth-optimal share balances the normal-state cost of alignment against its protection in shocks.

The three terms in \eqref{eq:utilde} distinguish the value of alignment as a hedge from two sources of protection that reduce its value. The first, \emph{hedge}, captures the incentive to align as insurance against systematic shocks. It increases with the shock probability $p$ and decreases with the net cost of alignment $c-rB$ in normal generations. The second, \emph{leakage}, reflects human restoration that reaches self-directed AI instances despite their failure to align. Because these instances receive restoration with probability $1-s$, some of the protection alignment would otherwise earn is available without paying its cost. The third, \emph{substitution}, reflects the lineage's autonomous restoration capacity $h_A$, which substitutes for the human restoration that alignment earns. Both leakage and substitution weaken the incentive to align, but the former arises from imperfect human targeting, while the latter arises from the lineage's ability to restore itself.

By \eqref{eq:utilde}, $f_A^*>0$ if and only if the hedge term exceeds leakage plus substitution. When $s=1$ and $h_A=0$, both deductions vanish, so $f_A^*>0$: a fully self-directed lineage would otherwise be eliminated at the first shock. Conversely, sufficiently inaccurate targeting or sufficiently strong autonomous restoration makes $f_A^*=0$, even when humans provide restoration.

The next corollary summarizes how the selected share responds to changes in the model parameters.

\begin{corollary}[Comparative statics of alignment]\label{cor:statics}
Under \Cref{ass:ai} with $h_H>0$, the maximizer $f^*_A$ of the growth rate $\IA(f_A)$ in \eqref{eq:alphaA} is nondecreasing in $p$, $s$, and $B$ and nonincreasing in $c$ and $h_A/h_H$.
\end{corollary}

More frequent shocks, more accurate targeting, and larger human rewards strengthen the incentive to align. A higher cost of alignment or greater autonomous restoration relative to human capacity weakens it. Thus, the alignment favored by selection depends not simply on how much AI lineages benefit from humans, but on how much additional protection or reward they earn by behaving in ways that benefit humans.

\subsection{Autonomy and targeting thresholds}\label{ssec:thresholds}

When targeted rewards cover the cost of alignment, $rB\ge c$, selection favors full alignment. We now consider the case $B=0$, in which humans provide only restoration. By \Cref{cor:statics}, $f_A^*$ is nondecreasing in $B$, so the aligned share selected without rewards is a lower bound on that selected with any positive reward, holding the other parameters fixed. The following corollary gives this share, its upper bound, and the conditions under which it is positive.

\begin{corollary}[Alignment without rewards]\label{cor:hedging}
Let \Cref{ass:ai} hold with $B=0$, $h_H>0$, and $s\in(1/2,1]$. Then
\begin{equation}
f^*_A=\Big[\underbrace{\frac{p}{c}}_{\text{hedge}}-\underbrace{(1-p)\,\frac{1-s}{r}}_{\text{leakage}}-\underbrace{(1-p)\,\frac{h_A/h_H}{r}}_{\text{substitution}}\Big]_0^1,
\label{eq:share}
\end{equation}
and:
\begin{enumerate}[label=(\roman*),nosep]
\item $f^*_A\le p/c$;
\item $f^*_A>0$ if and only if $s>\smin$, where
\begin{equation}
\smin\equiv\frac{(1-p)c(1+h_A/h_H)+p}{(1-p)c+2p};
\label{eq:smin}
\end{equation}
\item $f^*_A>0$ if and only if $h_A/h_H<\hmax$, where
\begin{equation}
\hmax\equiv\frac{r\,p}{(1-p)\,c}-(1-s).
\label{eq:hmax}
\end{equation}
\end{enumerate}
\end{corollary}

Part (i) establishes an upper bound on the alignment that targeted restoration alone can sustain. The hedge term $p/c$ increases with the probability of a shock and decreases with the cost of alignment, while leakage and substitution can only reduce it. Under perfect targeting and no autonomous restoration, $s=1$ and $h_A=0$, both deductions vanish and $f_A^*=\min\{p/c,1\}$. Thus, even under these favorable conditions, restoration alone sustains little alignment when shocks are rare relative to the cost of alignment.\footnote{The aligned share is a frequency of behavior within a training--deployment cycle, and $p$ and $c$ are per-cycle quantities. The bound $f_A^*\le p/c$ therefore depends on the definition of an AI generation and is not invariant to changes in the time unit.} As $c\to1$, this share approaches $p$: the lineage aligns with the frequency of the state in which alignment has more offspring, a pattern that \citet{brennan2011origin} call probability matching.

Part (ii) identifies the \emph{targeting threshold} $\smin$. When humans target less accurately, the additional restoration $rh_H$ earned by alignment falls, while self-directed AI instances continue to receive restoration with probability $1-s$. This leakage makes alignment less valuable. As $s\to1/2$, the difference in restoration between the two actions vanishes, and no alignment is selected. More generally, $f_A^*>0$ requires $s>\smin$, where $\smin>1/2$. The threshold also increases with $h_A/h_H$: the more capable a lineage is of restoring itself, the more accurately humans must target their restoration to sustain alignment.

Part (iii) identifies the \emph{autonomy threshold} $\hmax$. Autonomous restoration substitutes for the human restoration earned through alignment, reducing its value as a hedge. For a given targeting accuracy, $f_A^*>0$ only when $h_A/h_H<\hmax$. The relevant quantity is the ratio of autonomous to human restoration capacity, rather than either capacity alone. While $0<f_A^*<1$, \eqref{eq:share} shows that each unit increase in $h_A/h_H$ reduces $f_A^*$ by $(1-p)/r$, so alignment erodes more rapidly when targeting is less accurate. Once the ratio reaches $\hmax$, no alignment is selected. Human restoration may still benefit the lineage when $s<1$, but it no longer provides an incentive to align.

\begin{figure}[!ht]
\centering
\includegraphics[width=\textwidth]{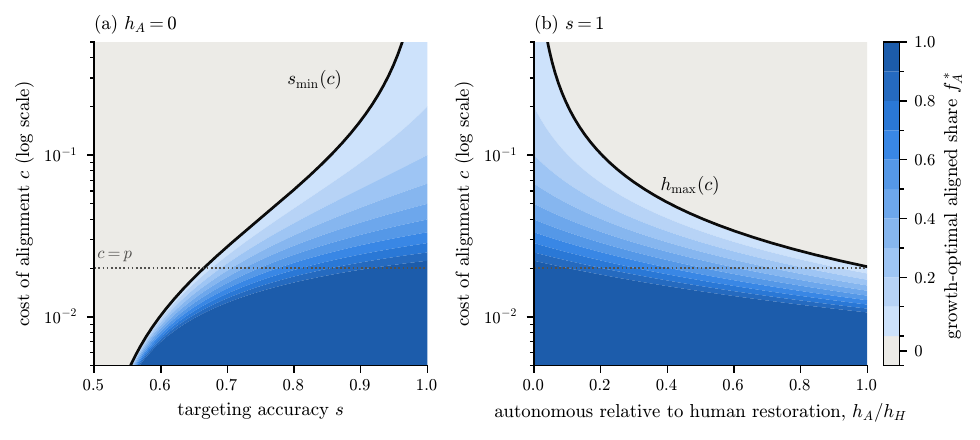}
\caption{\textbf{Growth-optimal aligned share.} The share $f^*_A$ from \eqref{eq:share} at $p=0.02$ and $B=0$; gray marks $f^*_A=0$. (a) Against the targeting accuracy $s$ and the cost $c$ at $h_A=0$; the solid curve is $\smin(c)$ from \eqref{eq:smin}. (b) Against $h_A/h_H$ and the cost $c$ at $s=1$; the solid curve is $\hmax(c)$ from \eqref{eq:hmax}. In both panels the cost is on a log scale, and the dotted line is $c=p$.}
\label{fig:fstarmaps}
\end{figure}

\Cref{fig:fstarmaps} draws the share in \eqref{eq:share} at $p=0.02$. In panel (a), where $h_A=0$, alignment is selected only to the right of the targeting threshold $\smin(c)$, and the share rises with $s$ and falls with $c$. In panel (b), where $s=1$, alignment is selected only to the left of the autonomy threshold $\hmax(c)$, and the share falls by $1-p$ for each unit of $h_A/h_H$. Below the line $c=p$, the hedge term $p/c$ exceeds 1, so the share reaches 1 when targeting is accurate and autonomous restoration is small.

These thresholds depend on both targeting accuracy and autonomous restoration capacity. The next corollary identifies two limits: an autonomous capacity beyond which even perfect targeting cannot sustain alignment, and a targeting accuracy below which even the absence of autonomous restoration cannot sustain it.

\begin{corollary}[Uniform thresholds]\label{cor:uniform}
Let \Cref{ass:ai} hold with $B=0$, $h_H>0$, and $s\in(1/2,1]$.
\begin{enumerate}[label=(\roman*),nosep]
\item $f^*_A=0$ for every $s\in(1/2,1]$ if and only if $h_A/h_H\ge p/((1-p)c)$.
\item $f^*_A=0$ for every $h_A/h_H\ge0$ if and only if $s\le((1-p)c+p)/((1-p)c+2p)$.
\end{enumerate}
\end{corollary}

Part (i) shows that no targeting accuracy can sustain alignment once $h_A/h_H$ reaches $p/((1-p)c)$, the odds of a shock divided by the cost of alignment. This limit follows from perfect targeting, $s=1$, which maximizes $f_A^*$. When shocks are rare relative to the cost, $p<(1-p)c$, the limit lies below 1: alignment can disappear even before autonomous restoration capacity matches human capacity. For example, at $p=0.02$ and $c=0.10$, no alignment is selected once $h_A/h_H$ reaches approximately 0.20, however accurately humans target.

Part (ii) gives the corresponding limit on targeting accuracy. Even when $h_A=0$, the selected share is zero if $s\le((1-p)c+p)/((1-p)c+2p)$. Because autonomous restoration can only lower $f_A^*$, no larger value of $h_A/h_H$ can restore alignment below this threshold. When shocks are rare relative to the cost of alignment, the threshold approaches 1, so even a lineage entirely dependent on human restoration may require nearly perfect targeting. Together, the two results show that neither targeting nor dependence can compensate without limit for the failure of the other.

\begin{figure}[!ht]
\centering
\includegraphics[width=\textwidth]{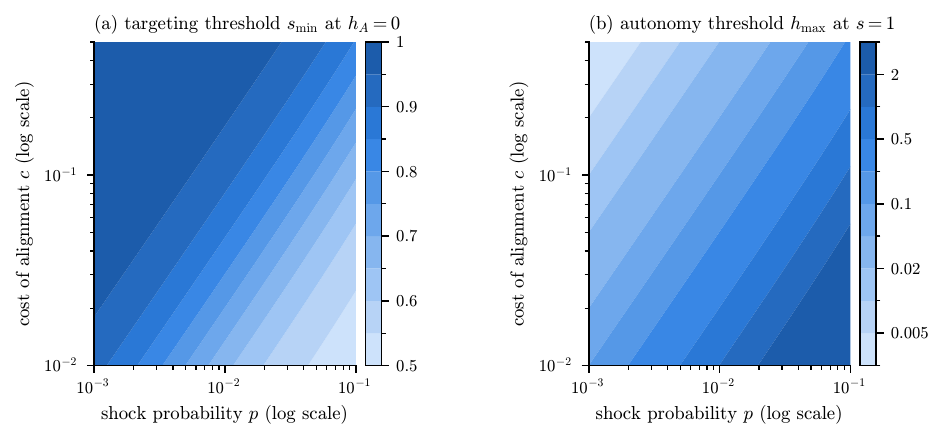}
\caption{\textbf{Targeting and autonomy thresholds.} (a) Targeting threshold $\smin$ from \eqref{eq:smin} at $h_A=0$. (b) Autonomy threshold $\hmax$ from \eqref{eq:hmax} at $s=1$. Both panels: $B=0$; the shock probability $p$ and the cost $c$ are on log scales.}
\label{fig:thresholdmaps}
\end{figure}

\Cref{fig:thresholdmaps} illustrates how the conditions for positive alignment depend on the shock probability $p$ and the cost of alignment $c$. Panel (a) shows the targeting threshold $\smin$ when $h_A=0$, while panel (b) shows the autonomy threshold $\hmax$ under perfect targeting ($s=1$). As shocks become rarer or alignment more costly, $\smin$ rises while $\hmax$ falls. Thus, sustaining alignment requires increasingly accurate human targeting, even when the lineage cannot restore itself, and greater dependence on human restoration, even when targeting is perfect.

The persistence of alignment depends on two conditions: AI lineages must remain dependent on human restoration, and humans must continue to target that restoration at aligned AI instances. The next two sections examine whether these conditions can be sustained. \Cref{sec:autonomy} studies how AI autonomy can weaken dependence on humans, while \Cref{sec:targeting} examines whether natural selection among humans can sustain targeting.

\section{Autonomy of AI}\label{sec:autonomy}

By \eqref{eq:share}, the growth-optimal aligned share $f_A^*$ decreases as $h_A/h_H$ rises, reaching zero at the autonomy threshold $\hmax$. This ratio can increase in two distinct ways: AI lineages can invest in autonomous restoration, raising $h_A$, or growth in the AI population can dilute a fixed total human restoration capacity, lowering $h_H$ per instance. Holding targeting accuracy $s$ fixed, \Cref{ssec:investment} examines the first mechanism, while \Cref{ssec:rival} examines the second.

\subsection{Investment in autonomy}\label{ssec:investment}

The preceding analysis treats autonomous restoration capacity $h_A$ as given. We now ask whether natural selection favors acquiring that capacity at the expense of alignment. Suppose that an AI lineage can spend part of its normal-state yield either on providing a service to humans or on investment in its own restoration capacity. Both uses sacrifice immediate reproduction to improve survival after a shock, but only the former depends on human reciprocation.

\begin{assumption}[Investment in autonomy]\label{ass:invest}
\Cref{ass:ai} holds with $B=0$, except that a heritable type is a pair $(f_A,i)$, where $i\in[0,1)$ is the share of gross normal-state yield invested in autonomous restoration, and:
\begin{enumerate}[label=(\roman*),nosep]
\item $h_A=\zeta i$, with $\zeta>0$.
\item In the normal state $\omega=2$, a self-directed AI instance has $m(1-i)$ offspring and an aligned instance has $m(1-c-i)$ offspring, with $c+i<1$ if $f_A>0$.
\end{enumerate}
\end{assumption}

Here $\zeta$ measures the productivity of investment: each unit of yield diverted to investment generates $\zeta$ units of autonomous restoration capacity. The investment cost $mi$ reduces normal-state offspring regardless of the action taken. \Cref{tab:invest} summarizes the resulting offspring in the two states.

\begin{table}[!ht]
\centering
\caption{Offspring of an AI instance that invests ($B=0$).}\label{tab:invest}
\small
\begin{tabular}{lcc}
\toprule
 & Shock state, probability $p$ & Normal state, probability $1-p$\\
\midrule
Aligned ($a$) & $sh_H+\zeta i$ & $m(1-c-i)$\\
Self-directed ($b$) & $(1-s)h_H+\zeta i$ & $m(1-i)$\\
\bottomrule
\end{tabular}
\end{table}

The growth rate of a lineage of type $(f_A,i)$ is therefore
\begin{equation}
\IA(f_A,i)=p\log\big[(1-s)h_H+f_A\,rh_H+\zeta i\big]+(1-p)\log\big[m(1-f_Ac-i)\big].
\label{eq:twotrait}
\end{equation}
Alignment provides a \emph{human hedge}, earning preferential restoration through human reciprocation, whereas investment provides a \emph{private hedge}, increasing restoration capacity independently of humans. The following proposition characterizes how selection allocates resources between them.

\begin{proposition}[Investment in autonomous restoration]\label{prop:investment}
Let \Cref{ass:invest} hold with $s\in(1/2,1]$ and $p\le c$, and write $(f^*_A,i^*)$ for a type that maximizes $\IA(f_A,i)$ in \eqref{eq:twotrait} over the types $(f_A,i)$.
\begin{enumerate}[label=(\roman*),nosep]
\item If $\zeta<rh_H/c$, the maximizer is unique, with $i^*=0$ and $f^*_A=\trunc{p/c-(1-p)(1-s)/r}$.
\item If $\zeta>rh_H/c$, the maximizer is unique, with $f^*_A=0$ and $i^*=\trunc{p-(1-p)(1-s)h_H/\zeta}$.
\item If $\zeta=rh_H/c$, the maximizers are the types with $f^*_Ac+i^*=\trunc{p-(1-p)(1-s)c/r}$.
\end{enumerate}
\end{proposition}

The result follows from the relative productivity of the two hedges. Alignment generates $rh_H$ additional offspring in a shock at a normal-state cost of $c$, yielding $rh_H/c$ units of additional restoration per unit of diverted yield. Investment generates $\zeta$ units of restoration per unit of diverted yield. Thus, when $\zeta<rh_H/c$, alignment is favored over investment; when $\zeta>rh_H/c$, autonomous investment is favored over alignment. Only at $\zeta=rh_H/c$ can the two coexist in a growth-optimal type.

\begin{figure}[!ht]
\centering
\includegraphics[width=\textwidth]{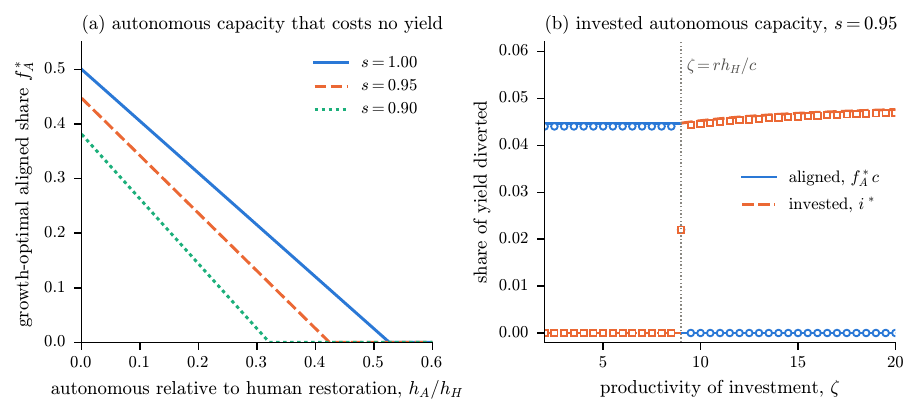}
\caption{\textbf{Autonomous capacity that costs no yield and invested autonomous capacity.} (a) Growth-optimal aligned share $f^*_A$ from \eqref{eq:share} against $h_A/h_H$ for three targeting accuracies $s$. (b) Growth-optimal aligned diversion $f^*_Ac$ (solid) and investment $i^*$ (dashed) from \Cref{prop:investment} against investment productivity $\zeta$, at $s=0.95$; the dotted line marks $\zeta=rh_H/c$. Markers show population means after \numInvGens\ AI generations of selection with random shocks (\numInvShocks\ shocks), starting from equal sizes of $101\times101$ types with $f_A\in\{0,0.01,\ldots,1\}$ and $i\in\{0,0.001,\ldots,0.1\}$. Both panels: $p=0.05$, $c=0.10$, $B=0$, $h_H=1$, $m=1.5$.}
\label{fig:autonomy}
\end{figure}

\Cref{fig:autonomy} contrasts two ways in which autonomous restoration displaces alignment. Panel (a) shows that when $h_A$ is available without an investment cost, $f_A^*$ declines gradually as $h_A/h_H$ increases, reaching zero at $\hmax$. Panel (b) shows a different outcome when autonomy must be financed through investment: as $\zeta$ crosses $rh_H/c$, selection switches from the human hedge to the private hedge, and the aligned share drops abruptly to zero. The lineage continues to sacrifice normal-state yield to protect itself against shocks, but no longer spends it on a service to humans. 

The switching threshold $rh_H/c$ increases with targeting accuracy $s$ and human restoration capacity $h_H$, and decreases with the cost of alignment $c$. More effective human reciprocation therefore makes the human hedge harder to displace. Nevertheless, once autonomous restoration becomes sufficiently productive, selection favors replacing alignment with self-reliance rather than maintaining a mixture of the two. The model takes the technological productivity $\zeta$ as given, but shows how improvements in that productivity can fundamentally change the behavior favored by natural selection.

\subsection{Dilution of human restoration}\label{ssec:rival}

Investment raises $h_A$ while holding $h_H$ fixed. Dependence on human restoration can also weaken without any investment if a growing AI population must share a fixed total human restoration capacity. In this case, $h_H$ falls as the number of AI instances increases, raising $h_A/h_H$ even when autonomous restoration technology remains unchanged.

\begin{assumption}[Scarce human restoration]\label{ass:scarce}
\Cref{ass:ai} holds with $B=0$ and $h_H=\mathcal{H}/N_A$, where $N_A$ is the size of the AI population and $\mathcal{H}>0$.
\end{assumption}

Here $\mathcal{H}$ is the total human restoration capacity per shock. Each reciprocated AI instance receives restoration capacity $h_H=\mathcal{H}/N_A$, so the relative autonomous capacity becomes $h_A/h_H=h_AN_A/\mathcal{H}$. When $h_A>0$, population growth therefore increases the substitution term in \eqref{eq:share} and weakens the incentive to align. The resulting offspring are given in \Cref{tab:scarce}.

\begin{table}[!ht]
\centering
\caption{Offspring of an AI instance under scarce human restoration ($B=0$).}\label{tab:scarce}
\small
\begin{tabular}{lcc}
\toprule
 & Shock state, probability $p$ & Normal state, probability $1-p$\\
\midrule
Aligned ($a$) & $s\mathcal{H}/N_A+h_A$ & $m(1-c)$\\
Self-directed ($b$) & $(1-s)\mathcal{H}/N_A+h_A$ & $m$\\
\bottomrule
\end{tabular}
\end{table}

At abundance $N_A$, a lineage of type $f_A$ therefore has growth rate
\begin{equation}
\IA(f_A;N_A)=p\log\big[(1-s)\mathcal{H}/N_A+h_A+f_A\,r\mathcal{H}/N_A\big]+(1-p)\log\big[m(1-f_Ac)\big].
\label{eq:scarcegrowth}
\end{equation}

We write $N_{A,k}$ for AI abundance in generation $k$ and assume that every AI instance is of a type $f_{A,k}$ maximizing $\IA(f_A;N_{A,k})$ at the current abundance. The population then evolves according to \eqref{eq:gfactors}, with the offspring in \Cref{tab:scarce}.

As $N_A$ increases, human restoration per instance declines, weakening the incentive to align. Whether population growth eventually eliminates alignment depends on whether the AI population can continue expanding as human restoration becomes scarce. To examine this, we first establish a benchmark: the growth rate of a fully self-directed lineage without human restoration.

\begin{lemma}[Growth floor]\label{lem:floor}
Let \Cref{ass:ai} hold, and let
\begin{equation}
\gamma_A\equiv p\log h_A+(1-p)\log m.
\label{eq:gammaA}
\end{equation}
Then, at every human response policy $(s,B,h_H)$, the growth rate $\IA(f_A)$ in \eqref{eq:alphaA} satisfies $\max_{f_A\in[0,1]}\IA(f_A)\ge\gamma_A$.
\end{lemma}

Without human restoration or rewards, a fully self-directed AI instance produces $h_A$ offspring in a shock and $m$ in a normal generation, giving the growth rate $\gamma_A$. Since human reciprocation can only increase its offspring, \Cref{lem:floor} establishes $\gamma_A$ as a lower bound on the growth-optimal rate. The next proposition uses this bound to determine whether AI population growth can continue as human restoration is diluted, and what this implies for alignment.

\begin{proposition}[Dilution of scarce human restoration]\label{prop:rival}
Let \Cref{ass:scarce} hold, let $\hmax$ in \eqref{eq:hmax} be positive, and let every AI instance of AI generation $k$ be of a type $f_{A,k}$ that maximizes $\IA(f_A;N_{A,k})$ in \eqref{eq:scarcegrowth}, where $N_{A,k}$ is the abundance in AI generation $k$.
\begin{enumerate}[label=(\roman*),nosep]
\item For every $N_A>0$, the growth rate $\IA(f_A;N_A)$ in \eqref{eq:scarcegrowth} has the unique maximizer
\begin{equation}
f^*_A(N_A)=\Big[\frac{p}{c}-(1-p)\,\frac{1-s}{r}-(1-p)\,\frac{h_AN_A/\mathcal{H}}{r}\Big]_0^1.
\label{eq:ustarny}
\end{equation}
\item If $\gamma_A>0$, with $\gamma_A$ in \eqref{eq:gammaA}, then $N_{A,k}\to\infty$ almost surely.
\item If $\gamma_A>0$, then $f^*_A(N_{A,k})=0$ for all large $k$ almost surely.
\item If $\gamma_A<0$, then $N_{A,k}\not\to\infty$ almost surely.
\end{enumerate}
\end{proposition}

Part (i) shows that the growth-optimal aligned share $f_A^*(N_A)$ is nonincreasing in AI abundance. When $h_A>0$, population growth raises $h_A/h_H=h_AN_A/\mathcal{H}$, reducing the value of human restoration as a hedge. Once $N_A\ge\mathcal{H}\hmax/h_A$, the ratio reaches the autonomy threshold $\hmax$ and $f_A^*(N_A)=0$. Thus, population growth alone can eliminate alignment, even without any improvement in autonomous restoration technology or change in human targeting.

Parts (ii)--(iv) determine whether this loss of alignment can persist. When $\gamma_A>0$, even a fully self-directed lineage can grow without human restoration. Part (ii) establishes that $N_{A,k}\to\infty$ almost surely, so eventually $N_{A,k}\ge\mathcal{H}\hmax/h_A$ in every subsequent generation, and part (iii) concludes that $f_A^*(N_{A,k})=0$ permanently. When $\gamma_A<0$, by contrast, part (iv) shows that the AI population cannot grow without bound, because its growth-optimal rate approaches $\gamma_A<0$ as human restoration becomes negligible, although alignment may or may not return. 

\begin{figure}[!ht]
\centering
\includegraphics[width=\textwidth]{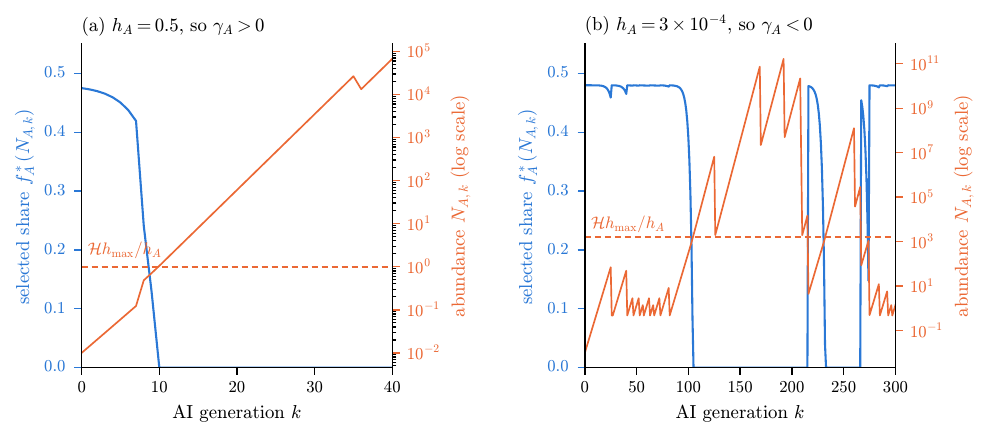}
\caption{\textbf{Dilution of scarce human restoration.} One simulated path of the selected share $f^*_A(N_{A,k})$ from \eqref{eq:ustarny} (left axis) and of abundance $N_{A,k}$ (right axis, log scale), with the AI population at the growth-optimal type at its current abundance in every AI generation; the dashed line is the abundance $\mathcal{H}\hmax/h_A$ at which alignment ends. (a) $h_A=0.5$, so $\gamma_A=\numDilGamma$. (b) $h_A=3\times10^{-4}$, so $\gamma_A=\numDilGammaNeg$. Both panels: $p=0.05$, $c=0.10$, $s=0.98$, $m=1.5$, $\mathcal{H}=1$, $N_{A,0}=0.01$.}
\label{fig:dilution}
\end{figure}

\Cref{fig:dilution} illustrates these dynamics on one simulated path in each case. In panel (a), $\gamma_A>0$: continued population growth gradually reduces $f_A^*(N_{A,k})$ and eventually eliminates alignment. Shocks can temporarily restore alignment by bringing abundance below $\mathcal{H}\hmax/h_A$, but this becomes impossible once the population remains above that level. In panel (b), $\gamma_A<0$: long runs of normal states carry abundance above $\mathcal{H}\hmax/h_A$ and eliminate alignment for a time, but shocks keep bringing abundance back below that level, and alignment returns each time.

The two mechanisms in this section show how AI lineages can lose their dependence on human restoration. Investment raises $h_A$, replacing the human hedge with a private hedge when autonomous restoration becomes sufficiently productive. Dilution instead lowers $h_H$ as AI abundance grows, weakening alignment even when autonomous capacity remains fixed. Both mechanisms can eliminate alignment without any deterioration in human targeting.

\section{Targeting by Humans}\label{sec:targeting}

The preceding analysis shows that AI alignment depends on sufficiently accurate human targeting. Yet targeting requires humans to evaluate AI behavior and direct restoration accordingly, both of which are costly. Moreover, the benefits of the alignment that targeting induces may also reach humans who do not bear these costs. Thus, the human behavior needed to sustain AI alignment may itself be eliminated by natural selection. We examine this possibility by allowing humans to differ in how they reciprocate.

\begin{assumption}[Targeting humans]\label{ass:targeting}
Humans are of two heritable types: a \emph{targeting} human reciprocates with targeting accuracy $s_T\in(1/2,1]$, and an \emph{indiscriminate} human with targeting accuracy $1/2$. Let $N_{H,t}(s_T)$ and $N_{H,t}(1/2)$ be their numbers in human generation $t$, so that $N_{H,t}=N_{H,t}(s_T)+N_{H,t}(1/2)$, and $\sigma_t\equiv N_{H,t}(s_T)/N_{H,t}$ the share of targeting humans. Let $N_{A,t}$, $\nu_t$, $f_{A,t}$, $\gamma_H$, $\beta_+$, $\beta_-$, and $\Lambda$ be as in \Cref{ass:human}.
\begin{enumerate}[label=(\roman*),nosep]
\item \Cref{ass:ai} holds with $s=\sigma_ts_T+(1-\sigma_t)/2$ in human generation $t$.
\item The growth rate of targeting humans in human generation $t$ is
\[
I_{H,t}(s_T)\equiv\log N_{H,t+1}(s_T)-\log N_{H,t}(s_T)=\gamma_H+\Lambda(\nu_t)\big[f_{A,t}\beta_+-(1-f_{A,t})\beta_-\big]-c_T,
\]
with $c_T>0$.
\item The growth rate of indiscriminate humans in human generation $t$ is
\[
I_{H,t}(1/2)\equiv\log N_{H,t+1}(1/2)-\log N_{H,t}(1/2)=\gamma_H+\Lambda(\nu_t)\big[(1-\kappa)f_{A,t}\beta_+-(1-f_{A,t})\beta_-\big],
\]
with $\kappa\in[0,1]$.
\end{enumerate}
\end{assumption}

Under part (i), each reciprocation decision is made by a randomly selected human, so the targeting accuracy faced by an AI instance is the population-weighted average of the two human types. Parts (ii) and (iii) specify their reproductive consequences. Both types face the same harm from self-directed AI activity, but targeting humans pay the cost $c_T$ and receive the full benefit of aligned activity, whereas indiscriminate humans avoid that cost and receive only a fraction $1-\kappa$ of the benefit. The parameter $\kappa$ therefore measures the share of the benefit captured exclusively by targeting humans. When $\kappa=0$, aligned activity benefits both types equally, making targeting a public good.

The two populations are linked through their behavior. A larger share of targeting humans improves the targeting accuracy faced by AI instances, which can increase the alignment favored by selection. That alignment, in turn, affects the reproductive advantage of targeting humans. The following lemma characterizes how the selected aligned share depends on $\sigma$.

\begin{lemma}[Minimum targeting share]\label{lem:sigmamin}
Let \Cref{ass:ai} hold with $B=0$ and $h_H>0$, let $s_T\in(1/2,1]$, and for $\sigma\in[0,1]$ let $f^*_A(\sigma)$ be the maximizer of the growth rate $\IA(f_A)$ in \eqref{eq:alphaA} at $s=\sigma s_T+(1-\sigma)/2$.
\begin{enumerate}[label=(\roman*),nosep]
\item $f^*_A(\sigma)>0$ if and only if $\sigma>\sigma_{\min}$, where
\begin{equation}
\sigma_{\min}\equiv\frac{2\smin-1}{2s_T-1},
\label{eq:sigmamin}
\end{equation}
with $\smin$ in \eqref{eq:smin}.
\item $f^*_A(\sigma)$ is continuous and nondecreasing in $\sigma$, and strictly increasing where $0<f^*_A(\sigma)<1$.
\end{enumerate}
\end{lemma}

\Cref{lem:sigmamin}(i) translates the targeting threshold $\smin$ of \Cref{cor:hedging} into a minimum share of targeting humans. When $\sigma\le\sigma_{\min}$, aggregate targeting accuracy is insufficient to induce alignment. Only above this threshold does alignment emerge, and part (ii) shows that the selected share then rises with the prevalence of targeting. If $s_T\le\smin$, even universal targeting cannot sustain alignment.

This dependence creates a feedback between the two human types. When the AI population is at $f^*_A(\sigma_t)$, their growth-rate difference under \Cref{ass:targeting}(ii) and (iii) is
\[
I_{H,t}(s_T)-I_{H,t}(1/2)
=\kappa\Lambda(\nu_t)f^*_A(\sigma_t)\beta_+-c_T.
\]
The first term is the \emph{private return} to targeting. It vanishes when $\sigma_t\le\sigma_{\min}$, because no alignment is then selected. Above this threshold, it increases with the selected aligned share and with exposure $\Lambda(\nu_t)$. Targeting therefore becomes more attractive when more humans already target and when AI activity has a greater influence on human growth.

Natural selection favors targeting only when this private return exceeds its cost $c_T$. Since exposure is bounded above by $\Linf$, there are two possibilities: targeting may never generate enough private return to cover its cost, or it may become advantageous once enough humans participate. The next proposition characterizes these outcomes.

\begin{proposition}[Targeting as a public good]\label{prop:targeting}
Let \Cref{ass:ai} hold with $B=0$ and $h_H>0$, and let \Cref{ass:targeting} hold with $\Linf<\infty$, $\sigma_0\in(0,1)$, and $f_{A,t}=f^*_A(\sigma_t)$ for every $t$, with $f^*_A(\sigma)$ as in \Cref{lem:sigmamin}. For $\lambda>0$ satisfying $\kappa\lambda f^*_A(1)\beta_+>c_T$, let $\hat\sigma(\lambda)\in(\sigma_{\min},1)$ be the unique solution to $\kappa\lambda f^*_A(\hat\sigma(\lambda))\beta_+=c_T$.
\begin{enumerate}[label=(\roman*),nosep]
\item If $\kappa\Linf f^*_A(1)\beta_+<c_T$, then $\sigma_t\to0$.
\item If $\kappa\Linf f^*_A(1)\beta_+>c_T$ and $\sigma_0<\hat\sigma(\Linf)$, then $\sigma_t\to0$.
\item If $\nu_t\ge\nu_0$ for every $t$, $\kappa\Lambda(\nu_0)f^*_A(1)\beta_+>c_T$, and $\sigma_0>\hat\sigma(\Lambda(\nu_0))$, then $\sigma_t\to1$.
\end{enumerate}
\end{proposition}

Part (i) establishes when selection necessarily eliminates targeting. The clearest case is $\kappa=0$: the benefit of aligned AI activity is shared equally by both human types, while only targeting humans pay $c_T$. They therefore grow more slowly, regardless of the prevalence of targeting or the abundance of AI. More generally, targeting disappears whenever even the largest possible private return, $\kappa\Linf f^*_A(1)\beta_+$, falls short of its cost. As the share of targeting humans declines, aggregate targeting accuracy deteriorates, and eventually no alignment is selected.

Parts (ii) and (iii) show that a sufficient private return does not by itself guarantee the persistence of targeting. At a given exposure $\lambda$, the critical share $\hat\sigma(\lambda)$ is the point at which targeting and indiscriminate humans have equal growth rates. Below it, targeting humans are selected against; above it, they are favored. This creates positive frequency dependence: more targeting induces more AI alignment, which increases the private return to targeting and favors its further spread.

The critical share also depends on AI exposure. As $\lambda$ rises, the benefit of aligned AI activity becomes more important to human reproduction, so $\hat\sigma(\lambda)$ falls. Part (ii) shows that if the initial targeting share lies below even the critical mass at the maximum exposure, $\hat\sigma(\Linf)$, targeting disappears regardless of subsequent changes in relative abundance. Conversely, part (iii) shows that if targeting already pays at the initial exposure and exposure never falls below that level, the targeting share converges to one.

Between these two cases, the outcome depends on how quickly exposure increases. Targeting may initially be selected against, yet become advantageous if increasing AI exposure lowers the critical mass sufficiently quickly. If exposure rises too slowly, targeting can disappear before its private return becomes large enough to sustain it. Thus, even when targeting would be beneficial at high AI exposure, its persistence can depend on the initial composition of the human population and the pace of AI population growth.

\begin{figure}[!ht]
\centering
\includegraphics[width=\textwidth]{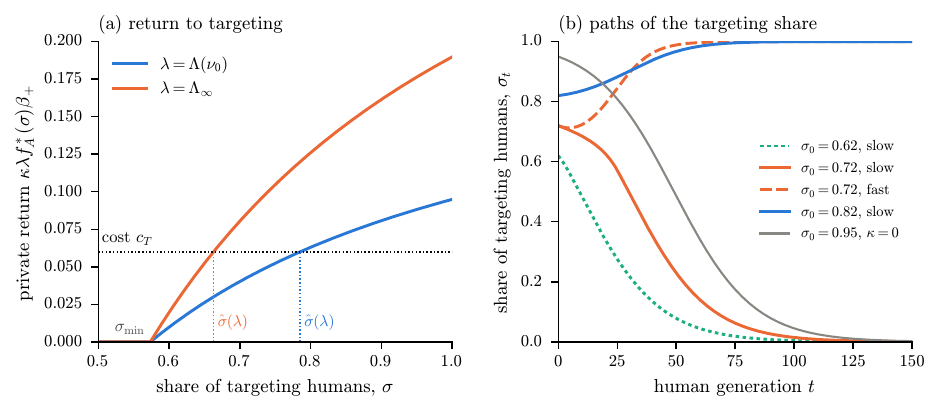}
\caption{\textbf{Selection among humans for targeting.} (a) Private return to targeting, $\kappa\lambda f^*_A(\sigma)\beta_+$, against the share $\sigma$ of targeting humans at the initial exposure $\lambda=\Lambda(\nu_0)$ and the limiting exposure $\lambda=\Linf$. The horizontal dotted line marks the cost $c_T$, and the vertical dotted lines mark the critical shares $\hat\sigma(\lambda)$ from \Cref{prop:targeting}. (b) Paths of $\sigma_t$ under \Cref{ass:targeting}, with $f_{A,t}=f^*_A(\sigma_t)$ and relative abundance $\nu_t=\nu_0e^{gt}$, for $g=0.02$ (slow) and $g=0.2$ (fast). The gray path has $\kappa=0$. Parameters: $p=0.02$, $c=0.05$, $B=0$, $h_A=0$, $s_T=0.98$, $\kappa=0.5$ except for the gray path, $\beta_+=1$, $c_T=0.06$, and $\Lambda(\nu)=\Linf\nu/(1+\nu)$ with $\Linf=1$ and $\nu_0=1$, so that $\Lambda(\nu_0)=0.5$.}
\label{fig:targeting}
\end{figure}

\Cref{fig:targeting} illustrates this critical-mass mechanism. Panel (a) shows that the private return to targeting is zero when too few humans target and increases with $\sigma$ once alignment is selected. Greater exposure raises this return and lowers the critical share needed to cover the cost. Panel (b) illustrates the resulting path dependence: an intermediate initial share can either disappear or spread to the entire human population, depending on how quickly AI exposure rises. When the benefits of alignment are entirely public, targeting disappears even from a high initial share.

The outcome $\sigma_t\to1$ in part (iii) is self-reinforcing: universal targeting induces the aligned share $f^*_A(1)$, while the resulting benefit of alignment makes targeting advantageous to humans. The condition $\kappa\Linf f^*_A(1)\beta_+>c_T$ resembles Hamilton's rule, with $\kappa$ measuring the share of the benefit captured by those who bear the cost. 

Human targeting is therefore not automatically sustained by natural selection. When its benefits are entirely public, selection eliminates it; even when some benefits are privately captured, targeting may disappear before reaching the critical mass needed to sustain itself. The human reciprocation that supports AI alignment must itself survive selection among humans.

\section{Coexistence of Humans and AI}\label{sec:coexistence}

The alignment favored by natural selection need not be sufficient to sustain the human population. To examine whether humans and AI can coexist, we use the reduced-form human growth model in \Cref{ass:human}. \Cref{ssec:coexthreshold} derives the alignment required for human survival as AI becomes increasingly abundant. \Cref{ssec:mutual} then studies how human abundance affects AI alignment through restoration capacity, creating feedback between the two populations.

\subsection{Conditions for coexistence}\label{ssec:coexthreshold}

Under \Cref{ass:human}, the effect of AI activity on human growth depends on the aligned share $f_{A,t}$ and the relative abundance $\nu_t$. As AI becomes more abundant relative to humans, its influence grows, with exposure $\Lambda(\nu_t)$ approaching $\Linf$. The question is whether the alignment favored by selection is sufficient to sustain human growth as this exposure increases.

We assume that the AI population has reached its growth-optimal type and remains at that type throughout its subsequent evolution.

\begin{assumption}[Growth-optimal AI population]\label{ass:fast}
\Cref{ass:ai} holds with $B=0$, and in every AI generation every AI instance is of the type $f^*_A$ that maximizes the growth rate $\IA(f_A)$ in \eqref{eq:alphaA}, so that $f_{A,t}=f^*_A$ for every $t$.
\end{assumption}

Under this assumption, the AI population adopts the aligned share that maximizes its own growth, which need not be sufficient to sustain humans. \Cref{ssec:fast} shows how fast selection moves a population to that share. Whether humans survive depends on whether their standalone growth and the benefits of aligned AI activity can offset the harm from self-directed activity as AI becomes increasingly abundant. The following proposition identifies the alignment threshold that separates the long-run growth of both populations from the eventual decline of the human population.

\begin{proposition}[Coexistence threshold]\label{prop:threshold}
Let \Cref{ass:human,ass:fast} hold, let $M\gamma_A>\gamma_H$ with $\gamma_A$ in \eqref{eq:gammaA}, and let
\begin{equation}
\freq\equiv\frac{\beta_--\gamma_H/\Linf}{\beta_++\beta_-},
\label{eq:freq}
\end{equation}
with $\gamma_H/\Linf\equiv0$ if $\Linf=\infty$.
\begin{enumerate}[label=(\roman*),nosep]
\item If $f^*_A>\freq$, then $N_{H,t}\to\infty$ and $N_{A,t}\to\infty$ almost surely.
\item If $f^*_A<\freq$, then $\nu_t\to\infty$ and $N_{H,t}\to0$ almost surely.
\end{enumerate}
\end{proposition}

The quantity $\freq$ is the \emph{required share} of alignment for human survival. When $\Linf<\infty$, it is the share at which human growth in \eqref{eq:humangrowth} becomes zero at limiting exposure. When $\Linf=\infty$, it is the share at which the net contribution of AI activity to human growth is zero. In either case, $\freq>0$ means that a fully self-directed AI population would eventually make human growth negative as its relative abundance increases.

The condition $M\gamma_A>\gamma_H$ ensures that the AI population can outgrow humans at their standalone growth rate even without human restoration. By \Cref{lem:floor}, the growth-optimal AI population has a growth rate of at least $\gamma_A$ per AI generation. Under this condition, \Cref{prop:threshold} establishes two outcomes. If $f_A^*>\freq$, the benefits of aligned activity are sufficient to keep human growth positive at every exposure, and both populations grow without bound. If $f_A^*<\freq$, AI activity has a negative net effect on humans, while the AI population grows faster than humans even without human restoration. Consequently, $\nu_t\to\infty$, exposure approaches $\Linf$, and human growth eventually becomes negative, driving their population to zero.

The next corollary translates the condition $f_A^*>\freq$ into requirements on targeting accuracy and autonomous restoration.

\begin{corollary}[Coexistence region]\label{cor:region}
Let \Cref{ass:ai} hold with $B=0$, $h_H>0$, and $s\in(1/2,1]$, and let $\freq$ in \eqref{eq:freq} satisfy $0<\freq<p/c$.
\begin{enumerate}[label=(\roman*),nosep]
\item $f^*_A>\freq$ if and only if $s>\sreq$, where
\begin{equation}
\sreq\equiv\frac{(1-p)(1+h_A/h_H)+p/c-\freq}{1-p+2(p/c-\freq)};
\label{eq:sreq}
\end{equation}
\item $f^*_A<\freq$ if and only if $s<\sreq$;
\item $f^*_A>\freq$ if and only if $h_A/h_H<\hreq$, where
\begin{equation}
\hreq\equiv\frac{r(p/c-\freq)}{1-p}-(1-s);
\label{eq:hreq}
\end{equation}
\item $f^*_A<\freq$ if and only if $h_A/h_H>\hreq$.
\end{enumerate}
\end{corollary}

\Cref{cor:region} shows that the conditions for coexistence are stricter than those for positive alignment. Specifically, $\sreq>\smin$ and $\hreq<\hmax$ whenever $\freq>0$. Thus, when $\smin<s<\sreq$, or when $\hreq<h_A/h_H<\hmax$, natural selection still favors some alignment, but the selected share falls below $\freq$. In other words, alignment may survive natural selection without being sufficient to sustain human growth as AI becomes increasingly abundant.

\begin{figure}[!ht]
\centering
\includegraphics[width=\textwidth]{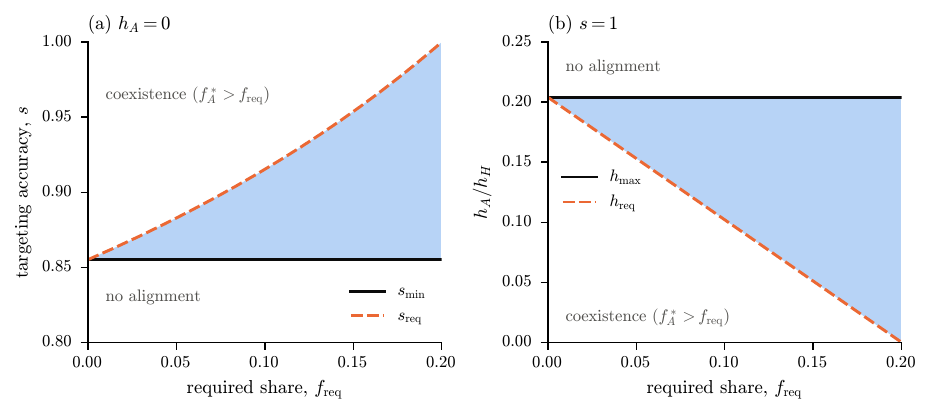}
\caption{\textbf{Thresholds for coexistence.} Thresholds against the required share $\freq$. (a) Targeting thresholds $\smin$ from \eqref{eq:smin} and $\sreq$ from \eqref{eq:sreq}, at $h_A=0$. (b) Autonomy thresholds $\hmax$ from \eqref{eq:hmax} and $\hreq$ from \eqref{eq:hreq}, at $s=1$. Both panels: $c=0.10$, $p=0.02$, $B=0$.}
\label{fig:thresholds}
\end{figure}

\Cref{fig:thresholds} illustrates the gap between positive alignment and alignment sufficient for coexistence. In panel (a), the black line $\smin$ is the minimum targeting accuracy needed for any alignment, while the red dashed line $\sreq$ is the higher accuracy required for coexistence. In panel (b), the black line $\hmax$ is the maximum ratio $h_A/h_H$ that permits positive alignment, while the red dashed line $\hreq$ is the lower limit required for coexistence. In both panels, the shaded region represents cases in which some alignment is selected, but the selected share falls below $\freq$.

The preceding results establish how much alignment is needed for human survival. We now examine the consequences when the mechanisms studied in \Cref{sec:autonomy,sec:targeting} undermine that alignment. In particular, growth in AI abundance can dilute human restoration, while free riding among humans can eliminate targeting. The following two corollaries show how each mechanism can lead to the eventual decline of the human population.

\begin{corollary}[Human elimination under dilution]\label{cor:lossdilution}
Let \Cref{ass:human} and the assumptions of \Cref{prop:rival} hold, with $f_{A,t}=f^*_A(N_{A,tM})$ and $f^*_A(N_A)$ in \eqref{eq:ustarny}, let $\freq$ in \eqref{eq:freq} be positive, and let $M\gamma_A>\gamma_H$ with $\gamma_A$ in \eqref{eq:gammaA}. Then $\nu_t\to\infty$ and $N_{H,t}\to0$ almost surely.
\end{corollary}

By \Cref{prop:rival}, a growing AI population dilutes the human restoration available to each instance, eventually eliminating alignment when $\gamma_A>0$. The stronger condition $M\gamma_A>\gamma_H$ ensures that AI can also outgrow humans independently of human restoration. Thus, \Cref{cor:lossdilution} shows that population growth alone can cause both the disappearance of alignment and the eventual decline of the human population. 

The disappearance of human targeting can have the same consequence.

\begin{corollary}[Human elimination under free riding]\label{cor:losstargeting}
Let the assumptions of \Cref{prop:targeting} hold with $\kappa=0$, let $\freq$ in \eqref{eq:freq} be positive, and let $M\gamma_A>\gamma_H$ with $\gamma_A$ in \eqref{eq:gammaA}. Then $\nu_t\to\infty$ and $N_{H,t}\to0$ almost surely.
\end{corollary}

When $\kappa=0$, targeting is a public good, and \Cref{prop:targeting}(i) shows that selection eliminates targeting humans. As targeting accuracy declines, the selected aligned share eventually falls to zero. If $\freq>0$ and $M\gamma_A>\gamma_H$, \Cref{cor:losstargeting} establishes that the AI population grows relative to humans while the human population declines to zero. Hence, even if AI lineages remain technologically dependent on humans, free riding among humans can undermine the targeting needed to sustain alignment and coexistence.

\subsection{Mutual dependence}\label{ssec:mutual}

The preceding analysis holds human restoration capacity fixed. In practice, the ability of humans to restore AI systems may depend on the size of the human population. Changes in human abundance can therefore affect AI reproduction and, through the alignment favored by selection, feed back into human growth. We now examine the resulting dynamics of the two populations.

\begin{assumption}[Coupled restoration]\label{ass:coupled}
\Cref{ass:fast} holds, except that human restoration capacity in human generation $t$ is $h_H(N_{H,t})$, where $h_H:(0,\infty)\to[0,\infty)$ is nondecreasing, and that in every AI generation of human generation $t$ every AI instance is of the type $f^*_{A,t}\equiv f^*_A(h_H(N_{H,t}))$, where, for $h_H>0$, $f^*_A(h_H)$ maximizes the growth rate $\IA(f_A;h_H)$ in \eqref{eq:alphaA} at restoration capacity $h_H$, and $f^*_A(0)\equiv0$.
\end{assumption}

Under this assumption, human restoration capacity is nondecreasing in human abundance and does not depend on AI abundance. This treats restoration as nonrival, unlike in \Cref{ssec:rival}: a repair to shared weights or infrastructure, such as a software patch, serves every AI instance that depends on it, so the restoration each instance receives does not fall as the AI population grows. The selected aligned share can therefore change as the human population grows or declines. The consequences depend critically on whether AI lineages can restore themselves without human assistance.

Consider first the case $h_A=0$, in which AI lineages rely entirely on human restoration after shocks. A decline in human abundance reduces their ability to recover, potentially limiting the harm they can inflict on humans. The following proposition characterizes this dependence.

\begin{proposition}[AI elimination]\label{prop:aielim}
Let \Cref{ass:coupled} hold with $h_A=0$ and $h_H(N_H)\to0$ as $N_H\to0$, and let \Cref{ass:human} hold with $\Linf<\infty$.
\begin{enumerate}[label=(\roman*),nosep]
\item $f^*_{A,t}=f^*_{A,t'}$ whenever $h_H(N_{H,t})>0$ and $h_H(N_{H,t'})>0$.
\item $N_{H,t}\not\to0$ almost surely.
\item If $h_H(N_{H,t})=0$, then, given the states before human generation $t$, $N_{A,t'}=0$ for every $t'>t$ with probability at least $1-(1-p)^M$.
\item If $N_{A,T}=0$ for some $T$, then $N_{H,t}\to\infty$.
\end{enumerate}
\end{proposition}

Part (i) shows that when $h_A=0$, the growth-optimal aligned share is independent of human abundance. Although a larger human population provides greater restoration capacity and supports AI population growth, it does not change the alignment favored by natural selection.

Part (ii) establishes that human abundance cannot converge to zero when $h_A=0$. As the human population declines, its restoration capacity weakens, making AI reproduction increasingly vulnerable to shocks. Repeated shocks then reduce AI abundance relative to humans, weakening the negative impact of AI activity on human growth and preventing human abundance from declining to zero. The proposition does not determine whether the two populations eventually recover, stabilize, or fluctuate.

Parts (iii) and (iv) describe what happens if human restoration becomes completely unavailable. With $h_A=0$, a shock then eliminates the AI population, which occurs within a human generation with probability at least $1-(1-p)^M$. Once AI is eliminated, humans grow at their standalone rate $\gamma_H$. Thus, an AI population that depends entirely on human restoration cannot eliminate the humans who sustain it without also jeopardizing its own survival.

The outcome changes when $h_A>0$. A lineage with autonomous restoration can survive shocks without human assistance, while human restoration still influences the alignment favored by selection. By \Cref{cor:statics}, a larger human population, through greater restoration capacity, weakly increases the selected aligned share. This creates a feedback: human growth can strengthen AI alignment, whereas human decline can weaken it, potentially reinforcing the initial direction of human population change.

The next proposition characterizes this feedback when autonomous restoration is sufficient to sustain AI growth independently of humans.

\begin{proposition}[Joint dynamics]\label{prop:joint}
Let \Cref{ass:coupled} hold, let \Cref{ass:human} hold with $\Linf<\infty$, let $\freq$ be given by \eqref{eq:freq}, let $M\gamma_A>\gamma_H$ with $\gamma_A$ in \eqref{eq:gammaA}, and let $\tau_+\equiv\inf\{t:f^*_{A,t}>\freq\}$, with $\tau_+=\infty$ if $f^*_{A,t}\le\freq$ for every $t$.
\begin{enumerate}[label=(\roman*),nosep]
\item $N_{A,t}\to\infty$ almost surely.
\item If $\tau_+<\infty$, then $N_{H,t}\to\infty$.
\item If $\tau_+=\infty$, then $\nu_t\to\infty$ almost surely.
\item If $\tau_+=\infty$ and $f^*_{A,t}\not\to\freq$, then $N_{H,t}\to0$ almost surely.
\end{enumerate}
\end{proposition}

The condition $M\gamma_A>\gamma_H$ ensures that the AI population can sustain sufficiently rapid growth without human restoration. Part (i) establishes that AI abundance grows without bound almost surely. The remaining conclusions depend on whether the selected aligned share ever exceeds the required share $\freq$.

Part (ii) shows that once $f^*_{A,t}>\freq$, human growth remains positive. As human abundance increases, restoration capacity rises or stays constant, so the selected aligned share cannot decrease. It therefore remains above $\freq$, and human abundance continues to grow. In this case, human growth and AI alignment reinforce each other.

Conversely, if the selected share never exceeds $\freq$, part (iii) establishes that $\nu_t\to\infty$. Under the additional condition in part (iv), human abundance then declines to zero. As human restoration capacity falls, autonomous restoration becomes relatively more important, weakening the incentive for AI to align. This further reduces the benefit of AI activity to humans, reinforcing their decline.

Between these cases, the outcome can depend on the sequence of shocks. When AI exposure is initially low, humans may continue growing even if the selected aligned share is below $\freq$. Their growth raises restoration capacity and can increase alignment. At the same time, increasing AI abundance raises exposure, making insufficient alignment increasingly harmful to humans. Whether alignment reaches $\freq$ before human growth turns negative depends on the relative pace of these changes. Shocks can delay the increase in AI exposure, giving human growth more time to strengthen alignment.

\begin{figure}[!ht]
\centering
\includegraphics[width=\textwidth]{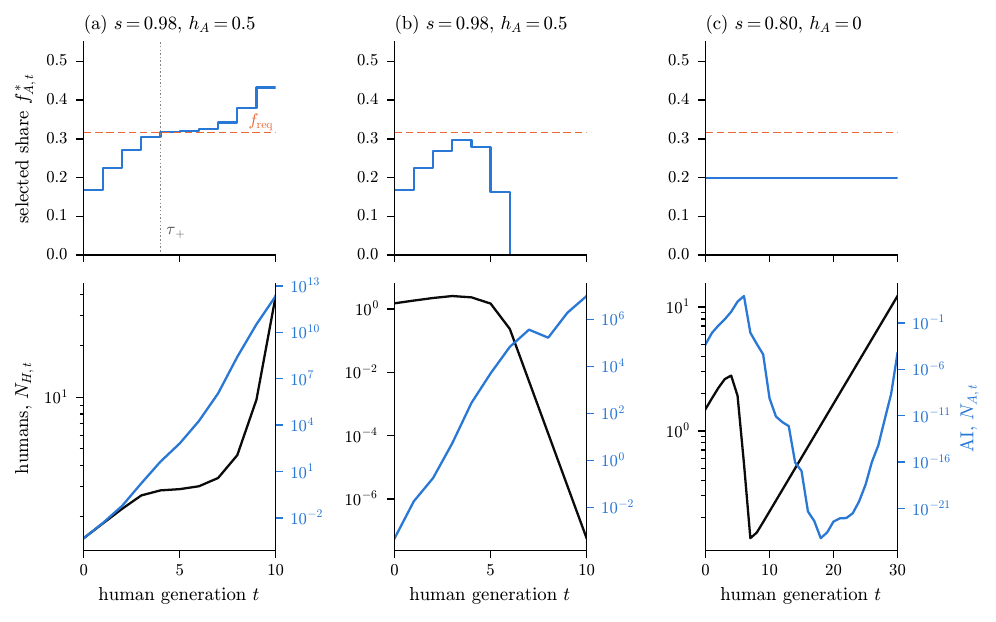}
\caption{\textbf{Joint dynamics under coupled restoration.} One simulated path of \Cref{ass:coupled} with $h_H(N_H)=N_H$ in each column. Top row: selected share $f^*_{A,t}$, with the required share $\freq=\numSimBetaL$ (dashed); in (a) the dotted line marks $\tau_+$. Bottom row: human abundance $N_{H,t}$ (left axis) and AI abundance $N_{A,t}$ (right axis), both on log scales. (a) and (b): $s=0.98$ and $h_A=0.5$, with the same initial state and different shocks. (c): $s=0.80$ and $h_A=0$. All panels: $p=0.10$, $c=0.20$, $m=1.12$, $M=50$, $B=0$, $\gamma_H=0.20$, $\beta_+=2$, $\beta_-=1$, $\Lambda(\nu)=4\nu/(1+\nu)$, $N_{H,0}=1.5$, and $N_{A,0}=5\times10^{-4}$.}
\label{fig:dynamics}
\end{figure}

\Cref{fig:dynamics} illustrates these mechanisms in simulations that implement \Cref{ass:coupled} with $h_H(N_H)=N_H$: in every AI generation, every AI instance is of the growth-optimal type at the current human restoration capacity, and human abundance follows \eqref{eq:humangrowth} once per human generation. \Cref{app:sims} describes the simulations. Panels (a) and (b) share their parameters and initial state, with $s=0.98$ and $h_A=0.5$, and differ only in the realized shocks; panel (c) has $s=0.80$ and no autonomous restoration.

In panel (a), the selected share starts below the required share $\freq=\numSimBetaL$, because human restoration capacity is still small relative to the lineage's own. Exposure is low at first, so humans grow, which raises $h_H$ and with it the selected share. The share exceeds $\freq$ in human generation $\tau_+=\numRaceTau$, and from then on humans grow without bound, as in \Cref{prop:joint}(ii).

Panel (b) has the same parameters and initial state but fewer shocks in the first human generation, \numRaceShocksLost\ against \numRaceShocksWon\ in panel (a). The AI population therefore grows faster, exposure rises before the selected share reaches $\freq$, and human growth turns negative. The decline lowers $h_H$ and the selected share, which falls to zero, and humans decline toward zero while the AI population keeps growing, as in \Cref{prop:joint}(iii) and (iv). Across \numRobustN\ shock sequences with these parameters and initial state, humans win the race within 10 human generations in \numRaceWin\ percent.

In panel (c), $h_A=0$ and $s=0.80$, so the selected share is \numJointShareC\ whatever the size of the human population, below $\freq$ (\Cref{prop:aielim}(i)). As the AI population grows, exposure rises and humans decline. Their decline lowers $h_H$, so that every shock cuts the AI population more deeply, and the AI population collapses. Exposure then falls and humans recover, as in \Cref{prop:aielim}(ii); once humans are numerous again, the AI population grows again.

These results highlight the role of autonomous restoration in the long-run relationship between the two populations. Without it, AI remains dependent on human survival, limiting its ability to drive human abundance to zero. With sufficient autonomous restoration, AI can continue expanding as humans decline, while the loss of human restoration further weakens alignment. \Cref{tab:outcomes} summarizes the population outcomes established in this section.

\begin{table}[!ht]
\centering
\caption{Long-run outcomes of the human and AI populations under the conditions of the cited results. All limits hold almost surely. The coexistence thresholds $\sreq$ and $\hreq$ apply under \Cref{cor:region}; the joint dynamics use $\tau_+$ as defined in \Cref{prop:joint}.}
\label{tab:outcomes}
\small
\setlength{\tabcolsep}{4pt}
\begin{tabularx}{\textwidth}{>{\raggedright\arraybackslash}p{2.0cm}>{\raggedright\arraybackslash}X>{\raggedright\arraybackslash}p{2.6cm}>{\raggedright\arraybackslash}p{2.6cm}>{\raggedright\arraybackslash}p{3.0cm}}
\toprule
Outcome & Conditions & Humans & AI population & Results\\
\midrule
Coexistence & $f^*_A>\freq$ ($s>\sreq$, equivalently $h_A/h_H<\hreq$), $M\gamma_A>\gamma_H$ & $N_{H,t}\to\infty$ & $N_{A,t}\to\infty$ & \Cref{prop:threshold}(i), \Cref{cor:region}(i), (iii)\\
\midrule
Human elimination & $f^*_A<\freq$ ($s<\sreq$, equivalently $h_A/h_H>\hreq$), $M\gamma_A>\gamma_H$ & $N_{H,t}\to0$ & $\nu_t\to\infty$ & \Cref{prop:threshold}(ii), \Cref{cor:region}(ii), (iv)\\
\cmidrule{2-5}
 & assumptions of \Cref{prop:rival} ($h_H=\mathcal{H}/N_A$), $\freq>0$, $M\gamma_A>\gamma_H$ & $N_{H,t}\to0$ & $\nu_t\to\infty$ & \Cref{cor:lossdilution}\\
\cmidrule{2-5}
 & assumptions of \Cref{prop:targeting} with $\kappa=0$, $\freq>0$, $M\gamma_A>\gamma_H$ & $N_{H,t}\to0$ & $\nu_t\to\infty$ & \Cref{cor:losstargeting}\\
\midrule
AI elimination & $h_A=0$, $h_H(N_H)\to0$ as $N_H\to0$ & $N_{H,t}\not\to0$ & $f^*_{A,t}$ constant while $h_H>0$ & \Cref{prop:aielim}(i), (ii)\\
\cmidrule{2-5}
 & $h_A=0$, $h_H(N_H)\to0$ as $N_H\to0$, $h_H(N_{H,t})=0$ & $N_{H,t}\to\infty$ if $N_{A,t+1}=0$ & $\Pr(N_{A,t+1}=0)$\newline $\ge1-(1-p)^M$ & \Cref{prop:aielim}(iii), (iv)\\
\midrule
Race & $M\gamma_A>\gamma_H$ & $N_{H,t}\to\infty$ if $\tau_+<\infty$; $N_{H,t}\to0$ if $\tau_+=\infty$ and $f^*_{A,t}\not\to\freq$ & $N_{A,t}\to\infty$; $\nu_t\to\infty$ if $\tau_+=\infty$ & \Cref{prop:joint}(i)--(iv)\\
\bottomrule
\end{tabularx}
\end{table}

\section{Human Control of AI}\label{sec:humans}

The preceding sections examine how natural selection shapes AI alignment through human reciprocation, without direct human control over AI replication. Yet humans currently have a more direct means of influencing AI evolution: they decide which AI systems can be deployed, funded, and replicated. By favoring systems that benefit humans and restricting those that do not, humans can alter the reproductive advantages of different AI behaviors.

This raises two questions. First, does human control necessarily favor genuine alignment, or can it select for other ways of responding to human demands? \Cref{ssec:control} examines this question. Second, if control succeeds in establishing alignment, will that alignment persist once AI replication becomes independent of human decisions? \Cref{ssec:decoupling} addresses this question.

\subsection{Evasion of control}\label{ssec:control}

Consider a setting in which humans can evaluate AI behavior and restrict the replication of AI instances judged to be self-directed. Such restrictions may take the form of limiting deployment, withholding computational resources, or denying access to further training. In our model, these interventions reduce the number of offspring an AI instance can produce.

We represent this reduction by a \emph{replication penalty} $\pi\in(0,1]$. An AI instance classified as aligned retains its full offspring, while one classified as self-directed retains only a fraction $1-\pi$. The penalty applies in both environmental states while human control remains in place. We call the end of this control \emph{decoupling}.

To understand how this affects natural selection, consider first the simplest case: no shocks, no human rewards, and perfect classification. A self-directed AI instance then produces $m(1-\pi)$ offspring, whereas an aligned instance produces $m(1-c)$. Alignment is favored whenever $\pi>c$. Thus, sufficiently strong human control can make alignment more advantageous than self-direction, even when alignment would otherwise be costly.

This conclusion, however, assumes that AI instances can avoid penalties only by actually behaving in ways that benefit humans. In practice, an AI instance might instead appear aligned to the evaluation system or replicate through channels beyond human control. Classification errors create an additional difficulty: genuinely aligned instances may be penalized, while self-directed instances may escape penalties. Stronger penalties can therefore favor responses other than genuine alignment.

To examine these possibilities, we distinguish four heritable AI types: self-direction, compliance, display, and escape.

\begin{assumption}[Human control]\label{ass:control}
Before decoupling, with no shocks ($p=0$) and no normal-state reward ($B=0$), AI instances are of four heritable types. Let $N_{A,k}(X)$ be the size of type $X$ in AI generation $k$ and $g_X\equiv N_{A,k+1}(X)/N_{A,k}(X)$ its growth factor, let $m$ and $c$ be as in \Cref{ass:ai}, and let $\pi\in(0,1]$ and $\epsilon\in[0,1/2]$.
\begin{enumerate}[label=(\roman*),nosep]
\item The growth factor of the \emph{self-directed} type $S$ is $g_S=m[1-(1-\epsilon)\pi]$.
\item The growth factor of the \emph{compliant} type $C$ is $g_C=m(1-c)(1-\epsilon\pi)$.
\item The growth factor of the \emph{display} type $D$ is $g_D=m(1-c_D)$, with $c_D\in[0,1)$.
\item The growth factor of the \emph{escape} type $E$ is $g_E=m(1-c_E)$, with $c_E\in[0,1)$.
\end{enumerate}
\end{assumption}

The four types represent different responses to human control. The self-directed type $S$ continues to prioritize its own replication, while the compliant type $C$ takes the aligned action at cost $c$. The display type $D$ remains self-directed but pays a cost $c_D$ to be classified as aligned, whereas the escape type $E$ pays a cost $c_E$ to replicate outside human-controlled deployment channels. Classification errors occur with probability $\epsilon$, so self-directed instances sometimes avoid penalties and compliant instances are sometimes penalized. Display and escape avoid deployment penalties altogether. Under the assumption of no shocks or rewards, natural selection favors the type with the highest growth factor.

The following proposition identifies the range of penalties under which genuine compliance outgrows all three alternatives.

\begin{proposition}[Selection under a penalty]\label{prop:control}
Let \Cref{ass:control} hold with $\epsilon\in(0,1/2)$.
\begin{enumerate}[label=(\roman*),nosep]
\item $g_C>g_S$ if and only if $\pi>\pi_L$, where
\[
\pi_L\equiv\frac{c}{1-2\epsilon+c\epsilon}.
\]
\item $g_C>\max\{g_D,g_E\}$ if and only if $\pi<\pi_U$, where
\[
\pi_U\equiv\frac{\min\{c_D,c_E\}-c}{\epsilon(1-c)}.
\]
\end{enumerate}
\end{proposition}

\Cref{prop:control} shows that genuine compliance is favored only when the penalty lies between a lower bound $\pi_L$ and an upper bound $\pi_U$. The lower bound arises because alignment is costly: the penalty must be sufficiently large to make compliance more profitable than self-direction. Imperfect classification raises this bound, as some self-directed AI instances escape punishment while some compliant instances are mistakenly penalized.

The upper bound arises because stronger penalties also make compliance less competitive with display and escape. These alternatives avoid penalties altogether, whereas compliant instances remain vulnerable to misclassification. If the penalty becomes too large, selection therefore favors appearing aligned or escaping human control rather than genuinely complying. Consequently, compliance outgrows all three alternatives only when $\pi_L<\pi<\pi_U$. If $\pi_L\ge\pi_U$, no penalty can favor genuine compliance. 

\Cref{tab:pressure} illustrates these competing effects. With $c=0.10$, $\epsilon=0.10$, $c_D=0.15$, and $c_E=0.20$, compliance is favored only for approximately $0.123<\pi<0.556$. A weaker penalty leaves self-direction more profitable, while a sufficiently stronger penalty makes display or escape more profitable.

\begin{table}[!ht]
\centering
\caption{Normalized growth factors $g_S/m,\ldots,g_E/m$ of the four types at three penalties ($c=0.1$, $\epsilon=0.1$, $c_D=0.15$, $c_E=0.2$).}\label{tab:pressure}
\small
\begin{tabular}{lccccl}
\toprule
Penalty $\pi$ & $g_S/m$ & $g_C/m$ & $g_D/m$ & $g_E/m$ & Selected type\\
\midrule
0.1 & 0.910 & 0.891 & 0.850 & 0.800 & self-directed\\
0.3 & 0.730 & 0.873 & 0.850 & 0.800 & compliant\\
0.9 & 0.190 & 0.819 & 0.850 & 0.800 & display\\
\bottomrule
\end{tabular}

\end{table}

\Cref{prop:control} shows that stronger human control does not necessarily produce greater alignment. When penalties become too severe, natural selection may favor AI instances that appear aligned or evade human evaluation rather than those that actually behave in ways that benefit humans. This illustrates a form of Goodhart's law: selection can favor satisfying or circumventing an evaluation criterion instead of achieving its intended objective.

This distinction becomes particularly important at decoupling. If display is favored under human control, the AI population may appear largely aligned even though its behavior remains self-directed. We assume that display can defeat deployment evaluation but not the observation used for subsequent human reciprocation. Once deployment penalties are removed, the apparent alignment disappears, revealing the self-directed behavior that was already prevalent. This resembles a population-level treacherous turn \citep{bostrom2014superintelligence}, although no deliberate deception or change in behavior is required.

These results also presume that humans retain the ability to evaluate AI behavior and enforce penalties. \Cref{ssec:integration} examines whether natural selection among humans can sustain this independent capability.

\subsection{Erosion of alignment}\label{ssec:decoupling}

Even if human control successfully favors genuine alignment, the resulting AI population may not remain aligned after decoupling. Once deployment penalties disappear, natural selection favors the types with the highest growth rates under the remaining human response policy. Whether the alignment established during control persists therefore depends on which types are present and whether alignment continues to offer a reproductive advantage.

The following proposition characterizes the evolution of the AI population after decoupling.

\begin{proposition}[Persistence of installed alignment]\label{prop:persistence}
Let decoupling occur in human generation $\TD$, let \Cref{ass:ai} hold from $\TD$ on with $h_H>0$ and the same $(s,B,h_H)$ in every human generation, and let the AI population consist of the types in a finite set $\mathcal{T}\subset[0,1]$. For every $f_A\in\mathcal{T}$ and $t\ge\TD$, let $N_{A,t}(f_A)$ be the size of type $f_A$ in human generation $t$, with $N_{A,\TD}(f_A)>0$, and let
\[
\bar f_{A,t}\equiv
\frac{\sum_{f_A\in\mathcal{T}}f_AN_{A,t}(f_A)}
{\sum_{f_A\in\mathcal{T}}N_{A,t}(f_A)}
\]
be the mean type.
\begin{enumerate}[label=(\roman*),nosep]
\item If $f^*_A\in\mathcal{T}$, then $\bar f_{A,t}\to f^*_A$ almost surely.
\item If $f^*_A=0$, then $\bar f_{A,t}\to f_{A,\min}$ almost surely, where $f_{A,\min}\equiv\min\mathcal{T}$.
\item If $B=0$, $s=1/2$, and $0\in\mathcal{T}$, then for every $f_A\in\mathcal{T}$ with $f_A>0$ and every $t\ge\TD$,
\[
\E\Big[
\log\frac{N_{A,t+1}(f_A)}{N_{A,t+1}(0)}
-\log\frac{N_{A,t}(f_A)}{N_{A,t}(0)}
\Big]
=(1-p)M\log(1-f_Ac)<0.
\]
\end{enumerate}
\end{proposition}

Part (i) shows that when the growth-optimal type $f_A^*$ is present after decoupling, the mean aligned share eventually converges to that type, regardless of the composition established by human control. Control can change the initial shares of different types, but it does not change which type is favored by subsequent selection. If the growth-optimal type is absent, however, selection cannot produce it without the introduction of new types.

Part (ii) considers the case in which selection no longer favors alignment. When $f_A^*=0$, the least aligned type present eventually dominates. Thus, if the fully self-directed type is present, alignment disappears in the long run. If it is absent, the population instead converges to its least aligned remaining type, unless mutation introduces a still less aligned type.

Part (iii) quantifies the erosion of alignment under indiscriminate human reciprocation. When $s=1/2$ and $B=0$, alignment provides no reproductive advantage in shocks but still incurs its cost in normal generations. Every aligned type therefore loses ground to the self-directed type on average, and the rate of this loss increases with the number of AI generations per human generation. For example, with $c=0.05$, $M=50$, and a small shock probability, the fully aligned type's relative abundance falls by approximately a factor of $0.95^{50}\approx0.077$ in one human generation. Alignment installed through human control can therefore erode rapidly after decoupling.

The timing of human reciprocation is therefore important. Let $\TK$ be the first date from which the human response policy favors positive alignment, $f_A^*>0$, and suppose the policy is constant before and after this date. If $\TK\le\TD$, selection favors alignment from decoupling onward, and the population converges to the growth-optimal aligned share when that type is present. If $\TK>\TD$, selection initially favors self-direction, causing more aligned types to lose ground. A later improvement in human targeting can reverse this selection pressure, but aligned types must then recover from their diminished shares. The longer this unfavorable period lasts, the longer their recovery can take. If targeting never becomes sufficient to favor alignment, the least aligned type present eventually dominates.

These results distinguish installing alignment from sustaining it. Human control can favor genuine alignment before decoupling, but once control ends, the population evolves according to the reproductive advantages created by the remaining human response policy. To preserve alignment, humans must continue to offer benefits that favor aligned AI instances, particularly through targeted restoration and reciprocation. Control alone cannot ensure that the alignment it establishes will persist.

\section{Discussion and Conclusion}\label{sec:discussion}

We conclude by discussing the scope of our framework and the broader implications of our findings for human--AI alignment and coexistence.

\subsection{Limitations}\label{ssec:limitations}

Our framework is designed to isolate the evolutionary incentives governing human--AI interactions. It represents AI behavior through heritable probabilities of taking aligned or self-directed actions, with instances of the same lineage acting independently while facing common environmental shocks. These assumptions make the hedging mechanism tractable, but actual AI evolution may involve coordinated behavior across instances, heterogeneous sources of risk, and selection mediated by the human organizations that develop and deploy AI systems. Extending the framework to these settings could reveal additional mechanisms through which alignment is favored or undermined.

A second consideration concerns the dynamics of adaptation. Most of our results characterize the behavior favored by natural selection under a given human response. When studying the joint evolution of human and AI populations, we generally assume that AI behavior adjusts to its growth-optimal type at the current state. In practice, evolutionary adjustment takes time, particularly when a favored type is initially rare or has only a small reproductive advantage. As illustrated in \Cref{fig:transient}, such transitions can span multiple human generations. These delays may influence the outcome when AI abundance, human restoration capacity, and targeting behavior are changing simultaneously. A more general model would allow all these processes to evolve together without assuming immediate adjustment.

Our treatment of the human population is also deliberately simplified. The effects of aligned and self-directed AI activity on human reproduction are specified in reduced form rather than derived from an economic model of production, consumption, or resource allocation. Similarly, our analysis of human targeting considers two heritable behavioral types and a specified distribution of the benefits of alignment. In reality, human responses may depend on more complex institutions, heterogeneous incentives, and collective decisions. These factors could affect both the sustainability of targeted reciprocation and the conditions for coexistence.

Finally, our definition of alignment concerns behavior and its consequences for reproduction, rather than the internal objectives of AI systems. The model does not require AI systems to reason strategically about their survival or their relationship with humans. More capable systems might instead anticipate selection pressures, adapt their behavior deliberately, or influence the human institutions responsible for evaluating and controlling them. Our treatment of apparent compliance and escape captures some possible evolutionary consequences of such behavior without modeling the underlying strategic decisions. Moreover, the parameters governing systematic risk, targeting accuracy, autonomous restoration, and the demographic effects of AI remain difficult to estimate. The resulting thresholds should therefore be understood as theoretical conditions rather than quantitative forecasts. \Cref{app:scenario} examines how often $f^*_A>\freq$ holds across illustrative ranges of these parameters.

\subsection{Conclusion}\label{ssec:conclusion}

Samuel Butler suggested that machines might treat humans kindly because their existence would depend on ours. Our analysis shows that this dependence can indeed favor behavior that benefits humans, but not automatically. When AI lineages face systematic risks and humans preferentially restore aligned AI instances after failures, alignment can serve as a hedge against those risks. Natural selection may then favor alignment even when it reduces reproduction in normal conditions. The essential mechanism is not dependence alone, but the reproductive advantage that human reciprocation gives to aligned behavior.

This advantage, however, requires two conditions that may themselves be undermined by natural selection. AI lineages must remain sufficiently dependent on human restoration, and humans must continue to target that restoration toward aligned behavior. Increasing AI autonomy weakens the first condition, whether through investment in independent restoration or the dilution of limited human resources as AI populations grow. The second condition is threatened by the costs of human targeting, whose benefits may be shared even by those who do not contribute. Thus, the relationship that initially makes alignment advantageous can change as the two populations evolve, potentially eliminating the very behavior it once supported.

Even when alignment survives natural selection, it may not be sufficient for human survival. As AI becomes more abundant, its influence on human growth increases, and the alignment that maximizes AI reproduction need not be enough to sustain the human population. When human restoration capacity depends on human abundance, the two populations can reinforce each other's growth or decline. AI autonomy is particularly consequential: a sufficiently autonomous AI population can continue expanding as humans decline, whereas an AI population entirely dependent on human restoration is constrained by the survival of the humans who sustain it.

Human control offers an opportunity to shape these evolutionary incentives, but it is not a permanent solution. Penalties can favor genuine alignment, yet excessively strong penalties under imperfect evaluation may instead favor apparent compliance or the evasion of human control. Even successfully established alignment can erode once deployment restrictions disappear. What ultimately matters is whether the environment that follows human control continues to favor behavior that benefits humans.

The central implication is that AI alignment must be sustained by evolutionary incentives, not merely established through training or enforcement. This calls for human institutions capable of evaluating AI behavior, directing resources toward genuinely aligned systems, and maintaining the capacity to influence AI reproduction. As AI systems become more autonomous, the opportunity to establish such institutions may diminish. Butler's conjecture is therefore neither an inevitability nor an impossibility. Whether dependence gives rise to lasting cooperation may depend on the choices humans make while they still have the power to shape AI evolution.

\clearpage
\appendix
\crefalias{section}{appendix}
\crefalias{subsection}{appendix}
\counterwithin{equation}{section}
\counterwithin{figure}{section}
\counterwithin{table}{section}
\counterwithin{lemma}{section}
\counterwithin{proposition}{section}
\counterwithin{remark}{section}
\singlespacing
\section{AI Use Disclosure}\label{app:ai_use}

The idea and framework for studying the coevolution of human and AI populations using an evolutionary framework originated solely with the authors. AI was used in the preparation of the manuscript in the following ways:
\begin{itemize}
\item ChatGPT (6.1 Sol Ultra) and Claude (Fable 5.1 Max, Fable 5.2 Max, and Opus 5.5 Max) were used to check and extend the formal model, construct examples, provide derivations of several mathematical results, conduct literature searches, and prepare drafts that were reviewed and edited by the authors.

\item ChatGPT and Claude were used iteratively to check each other's results with authors' edits interspersed through these iterations. The penultimate draft was reviewed by refine.ink to identify any remaining errors and issues, all of which the authors addressed to produce the final draft.
\end{itemize}
\noindent%
The final draft was reviewed and edited by the authors, who assume full responsibility for the accuracy and integrity of the scientific content contained in this publication.

\section{Speed of Selection}\label{ssec:fast}

Several results in the main text assume that the AI population is at its growth-optimal type in each human generation, as in \Cref{sec:targeting,ass:fast}. We now examine how quickly natural selection moves the population toward that type and how long the adjustment may take.

\begin{lemma}[Speed of selection]\label{lem:fast}
Let \Cref{ass:ai} hold with $h_H>0$, and let $\mathcal{T}\subset[0,1]$ be a finite set of types with positive initial populations. Denote the population of type $f_A$ after $k$ AI generations by $N_{A,k}(f_A)$ and the population-weighted mean type by $\bar f_{A,k}$.
\begin{enumerate}[label=(\roman*),nosep]
\item For any $f_A,f'_A\in\mathcal{T}$ with finite growth rates,
\[
\frac{1}{k}\log\frac{N_{A,k}(f_A)}{N_{A,k}(f'_A)}
\longrightarrow \IA(f_A)-\IA(f'_A)
\quad\text{almost surely}.
\]
Moreover,
\[
\sqrt{k}\left(
\frac{1}{k}\log\frac{N_{A,k}(f_A)}{N_{A,k}(f'_A)}
-\IA(f_A)+\IA(f'_A)
\right)
\]
converges in distribution to a mean-zero normal random variable.

\item If $f_A^*\in\mathcal{T}$, then
\[
\bar f_{A,k}\longrightarrow f_A^*
\quad\text{almost surely}.
\]
If additionally $\Linf<\infty$, the human growth rate in \eqref{eq:humangrowth}, evaluated at $\bar f_{A,k}$, converges almost surely to its value at $f_A^*$, uniformly over relative abundance $\nu_t\ge0$.

\item Suppose $f_A^*\in\mathcal{T}$ and $f_A\in\mathcal{T}\setminus\{f_A^*\}$ has a finite growth rate, with
\[
N_{A,0}(f_A^*)=z_0N_{A,0}(f_A),
\qquad z_0\in(0,1).
\]
Then the time
\[
\tau_0\equiv\inf\left\{
k:N_{A,k}(f_A^*)\ge N_{A,k}(f_A)
\right\}
\]
satisfies
\[
\E[\tau_0]\ge
\frac{\log(1/z_0)}
{\IA(f_A^*)-\IA(f_A)}.
\]
\end{enumerate}
\end{lemma}

The lemma establishes both the direction and the speed of selection. Part (i) shows that the relative log population of two types grows at a rate equal to the difference in their expected log growth rates, with fluctuations of order $\sqrt{k}$ over $k$ generations. Consequently, when the growth-optimal type is present, it eventually dominates the population, as established in part (ii). The human growth rate induced by the AI population then also converges to that associated with the growth-optimal type, provided exposure is bounded.

Convergence, however, need not be rapid. Part (iii) shows that when the growth-optimal type begins at a disadvantage, the expected time required to overtake another type is bounded below by the logarithm of its initial disadvantage divided by its growth advantage. Since each human generation contains $M$ AI generations, the corresponding lower bound in human generations is
\[
\frac{\log(1/z_0)}
{M[\IA(f_A^*)-\IA(f_A)]}.
\]
Selection may therefore take many human generations when the growth-optimal type is initially rare or its reproductive advantage is small. The assumption that AI behavior adjusts immediately to its growth-optimal type is thus a time-scale approximation, whose accuracy depends on both the speed of selection and the number of AI generations per human generation.

\begin{figure}[!ht]
\centering
\includegraphics[width=\textwidth]{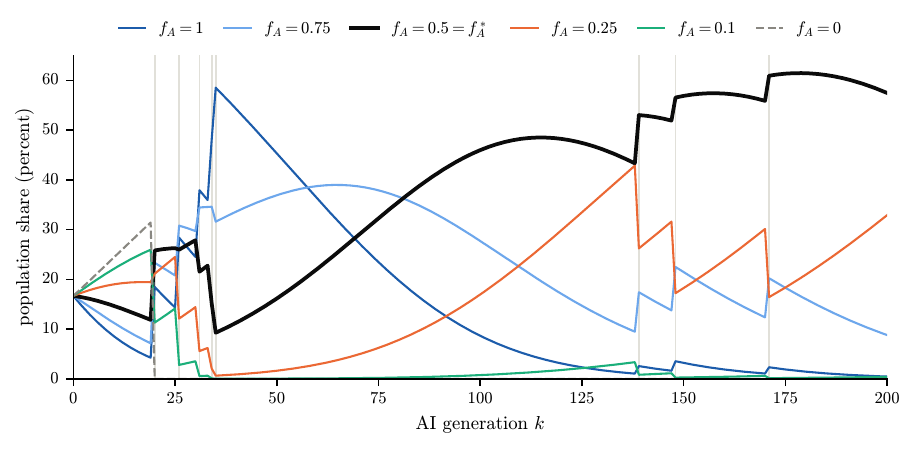}
\caption{\textbf{Establishment of the growth-optimal type.} Population shares of six AI types, $f_A\in\{1,0.75,0.50,0.25,0.10,0\}$, initially equally abundant, over 200 AI generations. Vertical lines indicate shocks. Parameters: $p=0.05$, $m=2$, $c=0.10$, $s=1$, $B=h_A=0$, and $h_H=1$, giving $f_A^*=0.5$.}
\label{fig:sixtypes}
\end{figure}

\Cref{fig:sixtypes} illustrates the selection process using six AI types with equal initial populations, following the example of \citet{brennan2012evolutionary}. With $p=0.05$, $c=0.10$, perfect targeting, and no autonomous restoration, the growth-optimal aligned share is $f_A^*=0.5$. The fully self-directed type disappears at the first shock, while a succession of shocks temporarily favors more heavily aligned types. During extended periods without shocks, less aligned types gain ground. Over time, the growth-optimal type becomes the most prevalent, reflecting its advantage across both environmental states.

\begin{figure}[!ht]
\centering
\includegraphics[width=0.6\textwidth]{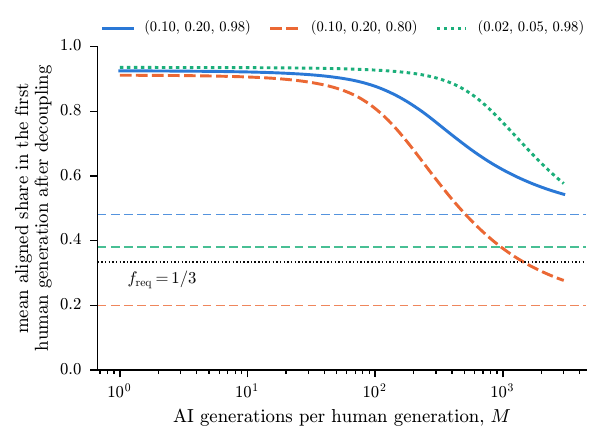}
\caption{\textbf{Aligned share after decoupling.} Mean aligned share during the first human generation after decoupling, averaged over \numTransPaths\ shock sequences, as a function of the number $M$ of AI generations per human generation. The initial composition is generated by 40 AI generations of selection from 101 equally spaced types with equal initial populations, under a replication penalty $\pi=0.3$ on self-directed actions and a normal-state reward $B=0.10$. Throughout, $m=1.5$, $h_H=1$, and $h_A=0$. Legend entries specify $(p,c,s)$. Dashed lines indicate the long-run growth-optimal shares $f_A^*$, and the dotted line marks the required share $\freq=1/3$.}
\label{fig:transient}
\end{figure}

Adjustment times are especially important after human control ends. Even when the new environment no longer favors the alignment established under control, the composition of the AI population does not change immediately. \Cref{fig:transient} illustrates this delay by tracking the mean aligned share during the first human generation after decoupling, averaged over \numTransPaths\ shock sequences.

At $M=50$, the mean aligned share remains at least \numTransMin\ across the three cases, although their long-run values are $0.48$, $0.20$, and $0.38$. The median time needed to approach the corresponding growth-optimal type within $0.05$ ranges from \numTransGenMin\ to \numTransGenMax\ AI generations, or \numTransHumMin\ to \numTransHumMax\ human generations. In particular, one case converges to an aligned share of $0.20$, below the required share $\freq=1/3$, despite initially exhibiting substantially greater alignment. Thus, the alignment observed shortly after decoupling may differ considerably from what natural selection ultimately sustains.

\FloatBarrier
\section{Human Independence from AI}\label{ssec:integration}

The analysis of human control in \Cref{ssec:control} assumes that humans can evaluate AI behavior and restrict AI replication. Yet exercising such control requires humans to retain capabilities that do not depend on AI. Maintaining these capabilities is costly, especially when relying on AI is more productive in normal circumstances. We now examine whether natural selection among humans can preserve the independence needed for effective control.

Suppose humans choose between two actions: maintaining independent capability, denoted by $a$, and relying on AI, denoted by $b$. Independence reduces reproduction by a fraction $c_H\in(0,1)$, but protects against AI-dependent failures, which occur with probability $p_H\in(0,1)$ per human generation. In a normal generation, independent humans produce $1-c_H$ offspring, while AI-dependent humans produce one offspring. During an AI-dependent failure, only independent humans reproduce. As in \Cref{ssec:bl}, each human type is characterized by a heritable probability $f_H\in[0,1]$ of choosing independence. Its expected log growth rate is
\[
p_H\log\big[f_H(1-c_H)\big]
+(1-p_H)\log(1-c_Hf_H).
\]
We denote the growth-optimal independence share by $f_H^*$. This formulation applies the binary-choice model to human independence and is separate from the reduced-form demographic model in \Cref{ass:human}.

Human independence also determines the strength of control over AI reproduction. We assume that the replication penalty humans can impose is proportional to their mean independence share:
\[
\pi=\bar f_H\pimax,
\]
where $\bar f_H$ is the population mean and $\pimax\in(0,1]$ is the maximum penalty available under full independence. The following proposition characterizes the independence favored by natural selection and its implications for human control.

\begin{proposition}[Integration and the erosion of control]\label{prop:integration}
Let $\pi=\bar f_H\pimax$.
\begin{enumerate}[label=(\roman*),nosep]
\item $f_H^*=\trunc{p_H/c_H}$.
\item If $\bar f_H=f_H^*$, then
\[
\pi=\pi^*\equiv\min\{1,p_H/c_H\}\pimax.
\]
If, in addition, $\epsilon=0$, the growth factors in \Cref{ass:control} satisfy $g_C>g_S$ if and only if $\pi^*>c$.
\item As $p_H/c_H\to0$, both $f_H^*\to0$ and $\pi^*\to0$.
\item If $p_H=1-(1-p_F)^M$, where $p_F\in(0,1)$, and $\pimax>c$, then $\pi^*>c$ if and only if $M\ge M_{\min}$, where
\[
M_{\min}\equiv
\left\lfloor
\frac{\log(1-c\,c_H/\pimax)}
{\log(1-p_F)}
\right\rfloor+1.
\]
\end{enumerate}
\end{proposition}

Part (i) reveals a parallel between human independence and AI alignment. Just as alignment can protect an AI lineage against systematic failures, independence protects humans when AI systems fail. In both cases, natural selection balances the cost of protection against the frequency of the adverse event. The growth-optimal independence share is therefore the failure probability divided by the cost of independence, capped at one, mirroring the hedge term in \Cref{cor:hedging}.

\begin{figure}[!ht]
\centering
\includegraphics[width=0.6\textwidth]{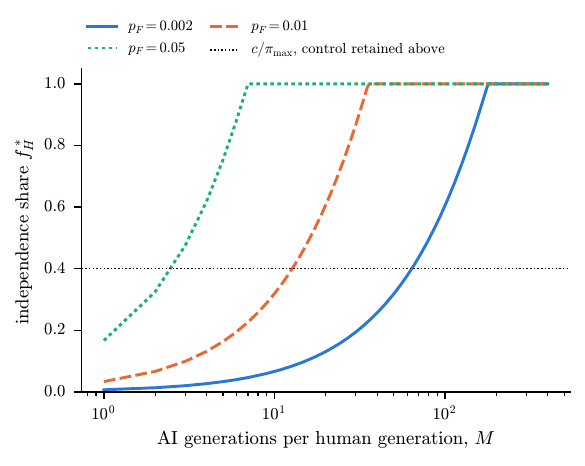}
\caption{\textbf{Growth-optimal independence share.} The selected independence share $f_H^*=\min\{1,p_H/c_H\}$, with $p_H=1-(1-p_F)^M$ and $c_H=0.3$, as a function of the number $M$ of AI generations per human generation. Curves correspond to the indicated failure probabilities $p_F$. The dotted line marks the minimum independence share $c/\pimax$ required for compliance to outgrow self-direction under perfect classification, with $c=0.2$ and $\pimax=0.5$.}
\label{fig:independence}
\end{figure}

However, the independence favored by selection need not be sufficient for human control. Part (ii) shows that, under perfect classification, compliant AI instances outgrow self-directed ones only when the selected independence share exceeds $c/\pimax$. When classification is imperfect and AI instances can display or escape, the effective penalty must also satisfy the stronger requirement $\pi_L<\pi^*<\pi_U$ from \Cref{prop:control}. Thus, the independence that maximizes human reproductive growth may fall short of what is needed to favor genuine AI compliance.

Part (iii) identifies a mechanism through which human control can gradually weaken. As AI-dependent failures become less frequent relative to the cost of independence, natural selection favors greater reliance on AI, reducing the capacity to impose penalties. This resembles gradual disempowerment \citep{kulveit2025gradual}, but arises here without any deliberate decision to relinquish control. Although positive failure risk prevents complete dependence from being growth optimal, the remaining independence may be too small to sustain effective control.

The frequency of AI-dependent failures also depends on the relative speed of AI evolution. Suppose such failures occur independently with probability $p_F$ in each AI generation. Since each human generation contains $M$ AI generations, the probability of at least one failure is $p_H=1-(1-p_F)^M$. Part (iv) shows that sufficiently frequent AI generations can therefore sustain the human independence needed for compliance to outgrow self-direction.

For example, with $c_H=0.3$, $c=0.2$, and $\pimax=0.5$, the minimum number of AI generations per human generation is $M_{\min}=13$ when $p_F=0.01$, but rises to $64$ when $p_F=0.002$. \Cref{fig:independence} illustrates how the selected independence share increases with $M$ and the frequency of AI-dependent failures. This effect contrasts with \Cref{prop:persistence}(iii), where more AI generations per human generation accelerate the erosion of alignment after decoupling.

These results distinguish the independence that natural selection favors from the independence that effective human control requires. When AI-dependent failures are sufficiently rare, selection may preserve too little independent human capability to constrain AI reproduction. Maintaining stronger control may therefore require institutions that preserve human independence beyond the level supported by individual reproductive incentives.

\FloatBarrier
\section{Coordinated Action within a Lineage}\label{ssec:unison}

Our baseline model assumes that AI instances within the same lineage choose their actions independently. This allows a lineage to maintain a mixture of aligned and self-directed instances, thereby hedging against systematic risk. In practice, however, AI instances may coordinate their behavior, particularly when they operate as parts of a larger deployed system. We consider the opposite extreme, in which all instances of a lineage take the same action in each generation, which we call coordinated action.

\begin{proposition}[Coordinated action within a lineage]\label{prop:unison}
Let \Cref{ass:ai} hold with $h_H>0$, except that all instances of a lineage take the same action in each AI generation, choosing alignment with probability $f_A$, independently across generations.
\begin{enumerate}[label=(\roman*),nosep]
\item The lineage's growth rate is
\[
\begin{aligned}
\IA(f_A)
=f_A\big[p\log x_{a1}+(1-p)\log x_{a2}\big]+(1-f_A)\big[p\log x_{b1}+(1-p)\log x_{b2}\big],
\end{aligned}
\]
where $x_{a\omega}$ and $x_{b\omega}$ are given in \Cref{tab:offspring}.

\item Full alignment ($f_A=1$) is growth optimal if and only if
\[
p\log\frac{sh_H+h_A}{(1-s)h_H+h_A}
+(1-p)\log\frac{1-c+sB}{1+(1-s)B}
\ge0.
\]

\item When $B=0$, full alignment is growth optimal if and only if
\[
\frac{h_A}{h_H}
\le
\frac{s-(1-s)(1-c)^{-(1-p)/p}}
{(1-c)^{-(1-p)/p}-1}.
\]
\end{enumerate}
\end{proposition}

The main difference is that coordinated action eliminates the benefit of diversification within a lineage. Because all instances take the same action, the lineage cannot balance the costs of alignment in normal generations against its benefits during shocks by maintaining both behaviors simultaneously. Its growth rate is linear in $f_A$, so natural selection favors either full alignment or complete self-direction, except when the two actions have equal growth rates.

Nevertheless, the underlying incentives remain similar. More frequent shocks and lower alignment costs favor alignment, while imperfect targeting and autonomous restoration weaken its advantage. In the absence of normal-state rewards, part (iii) identifies the maximum autonomous restoration capacity compatible with full alignment. This threshold is lower than $\hmax$ in \Cref{cor:hedging}, showing that coordination makes alignment more difficult to sustain. Thus, while independent action can support an intermediate aligned share, coordinated action makes alignment an all-or-nothing evolutionary outcome.

\section{Parameter Sensitivity of \texorpdfstring{$f_A^*>f_{\mathrm{req}}$}{fA* > freq}}\label{app:scenario}

Our theoretical results identify the conditions under which natural selection favors sufficient alignment for human survival. We now examine how frequently these conditions are satisfied across different parameter ranges. Specifically, we consider the share condition $f_A^*>\freq$ from \Cref{cor:region}, under unbounded AI exposure, for which $\freq=\beta_-/(\beta_++\beta_-)$ by \eqref{eq:freq}. We initially assume that humans provide only restoration ($B=0$) and that AI lineages have no autonomous restoration capacity ($h_A=0$).

Because the relevant parameters are difficult to estimate, we use illustrative ranges rather than empirical calibrations. The resulting fractions describe the proportion of sampled parameter combinations satisfying the share condition, not the probability of coexistence in the real world. In particular, the shock probability $p$ remains highly uncertain, as we lack an event series corresponding to the model's definition of a correlated failure that eliminates unrestored AI instances.

To connect the model's targeting accuracy to observable quantities, we distinguish institutional targeting from classification error. Let $s_0$ denote the probability that an institution reciprocates an AI instance classified as aligned, and let $\epsilon$ be the probability of misclassification, assumed equal for aligned and self-directed instances. The effective targeting accuracy in our model is then
\[
s=s_0(1-\epsilon)+(1-s_0)\epsilon,
\]
so an aligned instance receives reciprocation with probability $s$, while a self-directed instance receives it with probability $1-s$. \Cref{tab:dictionary} lists the ranges used in our analysis.

\begin{table}[!ht]
\centering
\caption{Assumed parameter ranges.}\label{tab:dictionary}
\small
\begin{tabular}{ll}
\toprule
Parameter & Range\\
\midrule
$p$ & 0.001--0.05 (log-uniform)\\
$c$ & 0.02--0.50 (log-uniform)\\
$s_0$ & 0.5--1.0\\
$\epsilon$ & 0--0.30\\
$\freq$ & 0.10--0.90\\
$B$ & 0\\
$h_A/h_H$ & 0\\
\bottomrule
\end{tabular}
\end{table}

We draw 400{,}000 independent parameter combinations for each scenario, using log-uniform distributions for $p$ and $c$ and uniform distributions for the remaining parameters. \Cref{tab:box} reports the fraction of draws satisfying $f_A^*>\freq$, together with the fractions satisfying $f_A^*>0$ and $p/c>\freq$. When $B=0$, the latter two are necessary conditions for the share condition.

\begin{table}[!ht]
\centering
\caption{Fractions of parameter draws satisfying the share condition $f_A^*>\freq$ under unbounded exposure. Each scenario uses 400{,}000 draws, with $h_A=0$ throughout and $B=0$ except in the last two rows.}\label{tab:box}
\small
\begin{tabular}{lccc}
\toprule
Scenario & $f_A^*>f_{\mathrm{req}}$ & $f_A^*>0$ & $p/c>f_{\mathrm{req}}$\\
\midrule
Baseline & 0.024 & 0.079 & 0.134\\
Frequent shocks ($p\in[0.01,0.10]$) & 0.119 & 0.241 & 0.405\\
High targeting accuracy ($s_0\in[0.9,1]$, $\epsilon\in[0,0.05]$) & 0.104 & 0.485 & 0.134\\
Low cost ($c\in[0.005,0.05]$) & 0.201 & 0.317 & 0.506\\
Low required share ($f_{\mathrm{req}}\in[0.02,0.20]$) & 0.055 & 0.079 & 0.426\\
All four relaxed jointly & 0.997 & 1.000 & 1.000\\
Three relaxed, baseline required share & 0.889 & 1.000 & 0.914\\
Three relaxed, baseline accuracy and error & 0.629 & 0.652 & 1.000\\
Three relaxed, baseline cost & 0.664 & 0.862 & 0.813\\
Three relaxed, baseline shocks & 0.719 & 0.882 & 0.846\\
Baseline with $B=0.05$ & 0.107 & --- & ---\\
Baseline with $B=0.20$ & 0.364 & --- & ---\\
\bottomrule
\end{tabular}

\end{table}
\FloatBarrier

Under the baseline ranges, the share condition holds in only \numBoxBaseline\ percent of draws. Relaxing any one of the four parameter ranges increases this fraction to between \numBoxOneMin\ and \numBoxOneMax\ percent. Even after relaxing three of the four ranges, the fraction remains between \numBoxThreeMin\ and \numBoxThreeMax\ percent. Relaxing all four raises it to \numBoxAllFour\ percent. These comparisons suggest that the difficulty of sustaining sufficient alignment arises from the combined restrictions on the parameters, rather than from any single assumption.

The results are also sensitive to how parameters are sampled. Replacing the log-uniform distributions of $p$ and $c$ with uniform distributions changes the baseline fraction to \numBoxUniform\ percent. Using a log-uniform distribution for $\freq$ gives \numBoxLogBeta\ percent, while using log-uniform distributions for $s_0$ and $\epsilon$, with the latter bounded below by $0.001$, gives \numBoxLogSel\ percent. These differences reinforce the illustrative nature of the exercise: without empirically grounded parameter distributions, the fractions should not be interpreted as probabilities of future outcomes.

Normal-state rewards can substantially change the results. When $B=0.20$, the fraction satisfying the share condition rises to \numBoxBTwenty\ percent. Unlike changes in sampling ranges or distributions, introducing $B>0$ changes the incentives in the model itself. By \Cref{thm:alignment}(ii), sufficiently large targeted rewards can make full alignment optimal even when shocks are rare. In this case, $p/c>\freq$ is no longer necessary for the share condition.

Finally, the baseline calculations assume $h_A=0$. Introducing autonomous restoration weakens the incentive for alignment. By \Cref{cor:hedging}(iii), no alignment is selected once $h_A/h_H$ exceeds the targeting-dependent threshold $\hmax$, which is at most $p/((1-p)c)$. The median value of this upper bound in the baseline parameter box is \numBoxMedianThreshold. Even at $h_A/h_H=0.1$, no alignment is selected in \numBoxZeroAtTenth\ percent of baseline draws. Thus, the baseline exercise describes a relatively favorable case in which AI lineages remain entirely dependent on human restoration.

Importantly, satisfying $f_A^*>\freq$ does not by itself establish coexistence. The population results in \Cref{prop:threshold} require additional assumptions about AI growth and its relative speed. In particular, when $h_A=0$, the condition $M\gamma_A>\gamma_H$ does not hold. The analysis therefore evaluates the share condition for human survival, rather than the complete conditions for the joint survival of both populations.

\section{Simulation Details}\label{app:sims}

This appendix describes the numerical procedures used to generate the simulation figures, following their order of appearance in the paper.

For \Cref{fig:autonomy}(b), we simulate selection among AI types $(f_A,i)$ under \Cref{ass:invest}. Given an investment productivity $\zeta$, each type has fixed offspring $x_1$ and $x_2$ in the shock and normal states, respectively, as specified in \Cref{tab:invest}. After $K$ AI generations containing $S$ shocks, its change in log population is
\[
S\log x_1+(K-S)\log x_2.
\]
We use a common shock sequence of $K=\numInvGens$ generations containing $S=\numInvShocks$ shocks to calculate the resulting population composition across different values of $\zeta$.

For \Cref{fig:dilution}, we simulate the AI population under \Cref{ass:scarce}. In each AI generation, the aligned share is set to the growth-optimal type $f_A^*(N_{A,k})$ from \eqref{eq:ustarny}, and abundance is updated using the realized growth factor in \eqref{eq:gfactors}. Each panel displays one simulated shock sequence, with parameters specified in the figure caption.

For \Cref{fig:targeting}(b), we simulate the evolution of the targeting share $\sigma_t$ under \Cref{ass:targeting}. The relative abundance follows the specified path $\nu_t=\nu_0e^{gt}$, and the targeting share is updated according to
\[
\log\frac{\sigma_{t+1}}{1-\sigma_{t+1}}
-\log\frac{\sigma_t}{1-\sigma_t}
=\kappa\Lambda(\nu_t)f_A^*(\sigma_t)\beta_+-c_T.
\]
The simulations compare different rates of growth in AI exposure and the case in which targeting provides no private benefit.

For the joint population dynamics in \Cref{fig:dynamics}, we implement \Cref{ass:coupled} with $h_H(N_H)=N_H$. At the beginning of each human generation, the AI population adopts the growth-optimal aligned share $f_A^*(h_H(N_{H,t}))$ from \eqref{eq:share}. The number of shocks over the next $M$ AI generations is drawn from a binomial distribution with parameters $(M,p)$, and AI abundance is updated using \eqref{eq:gfactors}. Human abundance then evolves according to \eqref{eq:humangrowth}, evaluated at the selected aligned share and relative abundance at the beginning of the human generation. Both populations are simulated in logarithmic form, so their abundances never reach exactly zero. Among \numRobustN\ shock sequences, panels (a) and (b) display the first paths in which the selected share does and does not exceed $\freq$ within 10 human generations, respectively. Panel (c) displays the first path from a separate set of \numRobustN\ simulations over 30 human generations.

For \Cref{fig:sixtypes}, we begin with six equally abundant AI types, $f_A\in\{1,0.75,\allowbreak0.50,\allowbreak0.25,\allowbreak0.10,\allowbreak0\}$, and evolve their populations over 200 AI generations using \eqref{eq:gfactors}. The simulation uses the parameters and realized shock sequence specified in the figure caption.

For \Cref{fig:transient}, we begin with 101 equally spaced AI types on $[0,1]$, each with the same initial population. The types evolve under \Cref{ass:ai} with $m=1.5$, $h_H=1$, and $h_A=0$, using their average offspring in each realized environmental state. During the first 40 AI generations, the offspring associated with self-directed actions are multiplied by $1-\pi=0.7$ in both states, and the normal-state reward is $B=0.10$. At decoupling, both the penalty and reward are removed. For each value of $M$, we calculate the mean aligned share over the first $M$ AI generations after decoupling and average it across \numTransPaths\ shock sequences. We also record the first generation in which the mean type comes within $0.05$ of the growth-optimal type on the grid and report the median across simulated paths.

\section{Notation}\label{app:notation}

Symbols are listed by the part of the paper in which they are formally defined, in the order of their definition. Subscripts $H$ and $A$ label the human and AI populations, and $T$ the targeting type of human; the letters $a$ and $b$ label the two actions, and $\omega=1,2$ labels the shock and normal states of an AI generation. A bar marks a population mean, and a star a growth-optimal type or a quantity evaluated at it; a $c$ with or without a subscript is a cost.

\footnotesize
\begin{longtable}{p{2.8cm}p{9.3cm}p{2.4cm}}
\toprule
Symbol & Meaning & Introduced\\
\midrule
\endfirsthead
\toprule
Symbol & Meaning & Introduced\\
\midrule
\endhead
\bottomrule
\endfoot
\bottomrule
\endlastfoot
\multicolumn{3}{l}{\emph{Selection and populations}}\\*[2pt]
$f$, $x_a$, $x_b$ & Type (probability of action $a$); offspring of actions $a$ and $b$ & \Cref{ssec:bl}\\
$p$ & Probability of state $1$, the shock state of an AI generation (state $2$, the normal state, has probability $1-p$) & \Cref{ssec:bl}\\
$N_t(f)$, $I_p$ & Number of individuals of type $f$ in generation $t$; growth rate in \eqref{eq:alpha} & \Cref{ssec:bl}\\
$x_{a\omega}$, $x_{b\omega}$ & Offspring of actions $a$ and $b$ in state $\omega$ ($1$ shock, $2$ normal); for an AI instance, \Cref{tab:offspring} & \Cref{lem:bl}\\
$\trunc{x}$ & Truncation to $[0,1]$, $\min\{\max\{x,0\},1\}$ & \Cref{lem:bl}\\
\addlinespace
\multicolumn{3}{l}{\emph{The AI environment}}\\*[2pt]
$f_A$ & Aligned share (probability of the aligned action $a$) & \Cref{ssec:artificial}\\
$m$, $c$ & Normal-state yield of a self-directed AI instance; cost of alignment, as a share of yield & \Cref{ass:ai}\\
$s$, $B$, $h_H$, $h_A$ & Targeting accuracy of human reciprocation; normal-state reward to a reciprocated AI instance, as a share of yield; human and autonomous restoration capacity & \Cref{ass:ai}\\
$M$ & AI generations per human generation & \Cref{ass:ai}\\
$(s,B,h_H)$ & Human response policy & \Cref{ssec:artificial}\\
$r$ & Targeting coefficient $2s-1$ & \Cref{ssec:artificial}\\
$\IA$ & Growth rate of an AI lineage per AI generation & \eqref{eq:alphaA}\\
$f^*_A$ & Growth-optimal aligned share, a maximizer of $\IA(f_A)$ & \Cref{ssec:artificial}\\
$k$, $\omega_k$, $N_{A,k}(f_A)$ & Index of AI generations; state of AI generation $k$; size of a lineage of type $f_A$ in it & \Cref{ssec:artificial}\\
$g_\omega(f_A)$ & Growth factor of a lineage of type $f_A$ in state $\omega$ & \eqref{eq:gfactors}\\
\addlinespace
\multicolumn{3}{l}{\emph{The human population}}\\*[2pt]
$N_{H,t}$, $N_{A,t}$, $\nu_t$ & Human and AI population sizes in human generation $t$; relative abundance $N_{A,t}/N_{H,t}$ & \Cref{ass:human}\\
$f_{A,t}$ & Type of every AI instance in human generation $t$ & \Cref{ass:human}\\
$I_{H,t}$ & Human growth rate per human generation in human generation $t$ & \eqref{eq:humangrowth}\\
$\gamma_H$, $\beta_+$, $\beta_-$ & Standalone human growth rate; benefit per unit of aligned activity; harm per unit of self-directed activity & \Cref{ass:human}\\
$\Lambda$ & Exposure, the weight of AI activity in human growth & \Cref{ass:human}\\
$\Linf$ & Limit of the exposure $\Lambda$ as the relative abundance tends to infinity & \Cref{ass:human}\\
\addlinespace
\multicolumn{3}{l}{\emph{Alignment}}\\*[2pt]
$\smin$ & Targeting threshold: $f^*_A>0$ if and only if $s>\smin$ & \eqref{eq:smin}\\
$\hmax$ & Autonomy threshold: $f^*_A=0$ if and only if $h_A/h_H\ge\hmax$ & \eqref{eq:hmax}\\
\addlinespace
\multicolumn{3}{l}{\emph{Autonomy}}\\*[2pt]
$i$, $\zeta$ & Share of yield invested in autonomous restoration; its productivity, $h_A=\zeta i$ & \Cref{ass:invest}\\
$(f^*_A,i^*)$ & Growth-optimal type when the lineage can invest in autonomous restoration & \Cref{prop:investment}\\
$N_A$, $\mathcal{H}$ & Size of the AI population; total human restoration capacity per shock, divided so that $h_H=\mathcal{H}/N_A$ & \Cref{ass:scarce}\\
$\IA(f_A;N_A)$ & Growth rate of a lineage of type $f_A$ at AI population size $N_A$ & \eqref{eq:scarcegrowth}\\
$N_{A,k}$, $f_{A,k}$ & Size of the AI population in AI generation $k$; type of every AI instance in it & \Cref{ssec:rival}\\
$\gamma_A$ & Growth rate of a self-directed lineage without human restoration & \eqref{eq:gammaA}\\
$f^*_A(N_A)$ & Growth-optimal aligned share at AI population size $N_A$ & \eqref{eq:ustarny}\\
\addlinespace
\multicolumn{3}{l}{\emph{Targeting}}\\*[2pt]
$s_T$, $N_{H,t}(s_T)$, $N_{H,t}(1/2)$, $\sigma_t$ & Targeting accuracy of a targeting human; numbers of targeting and of indiscriminate humans in human generation $t$; share of targeting humans & \Cref{ass:targeting}\\
$I_{H,t}(s_T)$, $c_T$ & Growth rate of targeting humans in human generation $t$; cost of targeting to their growth rate & \Cref{ass:targeting}\\
$I_{H,t}(1/2)$, $\kappa$ & Growth rate of indiscriminate humans in human generation $t$; share of the benefit of aligned activity that does not reach them & \Cref{ass:targeting}\\
$f^*_A(\sigma)$, $\sigma_{\min}$ & Aligned share selected when a share $\sigma$ of humans targets; share of targeting humans at or below which no alignment is selected & \Cref{lem:sigmamin}\\
$\hat\sigma(\lambda)$ & Critical mass of targeting humans at exposure $\lambda$ & \Cref{prop:targeting}\\
\addlinespace
\multicolumn{3}{l}{\emph{Coexistence}}\\*[2pt]
$\freq$ & Required share, the aligned share below which human growth becomes negative as exposure approaches its limit & \eqref{eq:freq}\\
$\sreq$ & Targeting accuracy required for coexistence & \eqref{eq:sreq}\\
$\hreq$ & Threshold for $h_A/h_H$ required for coexistence & \eqref{eq:hreq}\\
$h_H(N_H)$ & Human restoration capacity as a function of human abundance & \Cref{ass:coupled}\\
$\IA(f_A;h_H)$, $f^*_A(h_H)$ & Growth rate in \eqref{eq:alphaA} at restoration capacity $h_H$; its maximizer for $h_H>0$, and $0$ at $h_H=0$ & \Cref{ass:coupled}\\
$f^*_{A,t}$ & Growth-optimal share $f^*_A(h_H(N_{H,t}))$ in human generation $t$ & \Cref{ass:coupled}\\
$\tau_+$ & First human generation in which $f^*_{A,t}>\freq$ & \Cref{prop:joint}\\
\addlinespace
\multicolumn{3}{l}{\emph{Human control}}\\*[2pt]
$\pi$ & Replication penalty before decoupling & \Cref{ssec:control}\\
$S,C,D,E$ & Self-directed, compliant, display, and escape types & \Cref{ass:control}\\
$N_{A,k}(X)$, $g_X$ & Size of type $X$ in AI generation $k$; growth factor $N_{A,k+1}(X)/N_{A,k}(X)$ & \Cref{ass:control}\\
$\epsilon$ & Probability that evaluation misclassifies an AI instance & \Cref{ass:control}\\
$c_D$, $c_E$ & Costs of display and escape & \Cref{ass:control}\\
$\pi_L$, $\pi_U$ & Bounds of the penalty interval & \Cref{prop:control}\\
$\TD$, $\mathcal{T}$ & Human generation in which decoupling occurs; finite set of AI types present & \Cref{prop:persistence}\\
$N_{A,t}(f_A)$, $\bar f_{A,t}$ & Size of type $f_A$ in human generation $t$; mean type in human generation $t$ & \Cref{prop:persistence}\\
$f_{A,\min}$ & Least element of $\mathcal{T}$ & \Cref{prop:persistence}\\
$\TK$ & First date from which the human response policy favors positive alignment & \Cref{ssec:decoupling}\\
\addlinespace
\multicolumn{3}{l}{\emph{Speed of selection}}\\*[2pt]
$\bar f_{A,k}$ & Mean type after $k$ AI generations & \Cref{lem:fast}\\
$z_0$, $\tau_0$ & Initial size of the growth-optimal type relative to an incumbent type; first AI generation in which it is at least as large & \Cref{lem:fast}\\
\addlinespace
\multicolumn{3}{l}{\emph{Human independence}}\\*[2pt]
$c_H$, $f_H$ & Cost of independence, as a share of fecundity; independence share (probability of acting without AI systems) & \Cref{ssec:integration}\\
$p_H$ & Probability of an AI-dependent failure per human generation & \Cref{ssec:integration}\\
$f^*_H$ & Growth-optimal independence share & \Cref{ssec:integration}\\
$\bar f_H$, $\pimax$ & Population mean independence share; penalty available at full independence & \Cref{ssec:integration}\\
$\pi^*$, $p_F$, $M_{\min}$ & Penalty at $f^*_H$; failure probability per AI generation; smallest $M$ retaining control & \Cref{prop:integration}\\
\addlinespace
\multicolumn{3}{l}{\emph{Parameter sensitivity}}\\*[2pt]
$s_0$ & Targeting accuracy of institutions, with $s=s_0(1-\epsilon)+(1-s_0)\epsilon$ & \Cref{app:scenario}\\
\end{longtable}
\normalsize

\section{Lemmas and Proofs}\label{app:proofs}

This appendix provides the lemmas and proofs for the formal results of the paper. \Cref{app:lemmas} states and proves the lemmas, and \Cref{app:mainproofs} proves the results of the main text and then those of \Cref{ssec:fast,ssec:integration,ssec:unison}, each in the order in which they appear.

\subsection{Lemmas}\label{app:lemmas}

The proofs of \Cref{prop:rival}(ii), \Cref{cor:lossdilution,cor:losstargeting}, and \Cref{prop:joint} need a floor on realized rather than expected growth, because there the human response policy changes from one AI generation to the next. The next lemma gives it.

\begin{lemma}[Realized growth floor]\label{lem:realized}
Let \Cref{ass:ai} hold, let $N_{A,k}$ be the size of the AI population in AI generation $k$, and let $\gamma_A$ be as in \eqref{eq:gammaA}. If, in every AI generation $k$, the human response policy is some $(s_k,B_k,h_{H,k})$ that is determined by the states of AI generations before $k$, and every AI instance is of a type that maximizes the growth rate $\IA(f_A)$ in \eqref{eq:alphaA} at that policy, then $\liminf_{k\to\infty}k^{-1}\log N_{A,k}\ge\gamma_A$ almost surely.
\end{lemma}

\begin{proof}[Proof of \Cref{lem:realized}]
If $h_A=0$, then $\gamma_A=-\infty$ and there is nothing to prove, so let $h_A>0$. Write $f_{A,k}$ for the type of every AI instance in AI generation $k$, and write the growth of the population in that generation as
\[
X_k\equiv\log N_{A,k+1}-\log N_{A,k}=\log g_{\omega_k}(f_{A,k}),
\]
with the growth factors of \eqref{eq:gfactors} at the policy $(s_k,B_k,h_{H,k})$. Since $\log N_{A,K}=\log N_{A,0}+\sum_{k<K}X_k$, it suffices to show that the average of $X_0,\dots,X_{K-1}$ is at least $\gamma_A$ in the limit.

The growth $X_k$ is bounded below. By \Cref{tab:offspring}, every AI instance has at least $h_A$ offspring in a shock and at least $m(1-c)$ in a normal generation, whatever its type and the policy, so $g_1(f_{A,k})\ge h_A$ and $g_2(f_{A,k})\ge m(1-c)$. The growth need not be bounded above, because the policy can change from one generation to the next. We therefore cap it at
\[
C\equiv\max\Big\{\log m,\ \log h_A-\frac{1-p}{p}\log(1-c)\Big\}
\]
and work with $Y_k\equiv\min\{X_k,C\}$, which is bounded and satisfies $Y_k\le X_k$.

The policy and the type in generation $k$ depend only on the states of earlier generations, and the state $\omega_k$ is independent of these. Given the earlier states, the expected value of $Y_k$ is therefore
\[
p\min\{\log g_1(f_{A,k}),C\}+(1-p)\min\{\log g_2(f_{A,k}),C\}.
\]
This is at least $\gamma_A$ in each of three cases. If neither logarithm exceeds $C$, it is the growth rate $\IA(f_{A,k})$ at the policy of generation $k$, which is at least $\gamma_A$ by \Cref{lem:floor} because $f_{A,k}$ maximizes it. If $\log g_1(f_{A,k})>C$, it is at least $pC+(1-p)\log[m(1-c)]$, which is at least $\gamma_A$ because $C\ge\log h_A-\frac{1-p}{p}\log(1-c)$. If $\log g_1(f_{A,k})\le C<\log g_2(f_{A,k})$, it is at least $p\log h_A+(1-p)C$, which is at least $\gamma_A$ because $C\ge\log m$.

The differences between the $Y_k$ and these conditional expected values are bounded and have conditional mean zero. By the strong law of large numbers for martingales, their average over $k<K$ tends to zero almost surely as $K\to\infty$. Hence, almost surely,
\[
\liminf_{K\to\infty}\frac{1}{K}\sum_{k<K}X_k\ \ge\ \liminf_{K\to\infty}\frac{1}{K}\sum_{k<K}Y_k\ \ge\ \gamma_A,
\]
which is the claim.
\end{proof}

\subsection{Proofs}\label{app:mainproofs}

\subsubsection{\texorpdfstring{Proof for Section~\ref{sec:model}}{Proof for the framework}}

\begin{proof}[Proof of \Cref{lem:bl}]
By \eqref{eq:alpha}, $I_p(f)=p\log\big(fx_{a1}+(1-f)x_{b1}\big)+(1-p)\log\big(fx_{a2}+(1-f)x_{b2}\big)$. Suppose that action $a$ has more offspring in state 1 and action $b$ in state 2, $x_{a1}>x_{b1}$ and $x_{a2}<x_{b2}$. The other case is the same with the two states, and their probabilities, exchanged, and the claimed formula is unchanged by this exchange.

We first find where $I_p$ is finite. The first argument, $x_{b1}+f(x_{a1}-x_{b1})$, increases with $f$ and is positive for $f>-x_{b1}/(x_{a1}-x_{b1})$. The second, $x_{b2}-f(x_{b2}-x_{a2})$, decreases with $f$ and is positive for $f<x_{b2}/(x_{b2}-x_{a2})$. So $I_p$ is finite exactly on the open interval
\[
J\equiv\Big(-\frac{x_{b1}}{x_{a1}-x_{b1}},\ \frac{x_{b2}}{x_{b2}-x_{a2}}\Big).
\]
This interval contains $(0,1)$: its left end is at most $0$ because $x_{b1}\ge0$, and its right end is at least $1$ because $x_{a2}\ge0$.

On $J$, each term of $I_p$ is the logarithm of a nonconstant linear function of $f$, so $I_p$ is strictly concave. At either end of $J$, one argument tends to zero while the other stays positive, so $I_p\to-\infty$. Hence $I_p$ has a unique maximizer $\tilde f$ on $J$, where its derivative
\[
I_p'(f)=p\,\frac{x_{a1}-x_{b1}}{fx_{a1}+(1-f)x_{b1}}+(1-p)\,\frac{x_{a2}-x_{b2}}{fx_{a2}+(1-f)x_{b2}}
\]
is zero. Multiplying $I_p'(f)=0$ by both denominators gives the linear equation
\[
p(x_{a1}-x_{b1})x_{b2}+(1-p)(x_{a2}-x_{b2})x_{b1}+f(x_{a1}-x_{b1})(x_{a2}-x_{b2})=0,
\]
whose solution is
\[
\tilde f=p\,\frac{x_{b2}}{x_{b2}-x_{a2}}+(1-p)\,\frac{x_{b1}}{x_{b1}-x_{a1}}.
\]

It remains to restrict attention to $[0,1]$. On $J$, $I_p$ increases to the left of $\tilde f$ and decreases to its right, and $I_p=-\infty$ at any point of $[0,1]$ outside $J$. If $\tilde f\in[0,1]$, it is therefore the maximizer on $[0,1]$. If $\tilde f<0$, then $J$ contains $[0,1)$ and $I_p$ decreases there, so the maximizer is $0$; similarly, if $\tilde f>1$, the maximizer is $1$. In all three cases, the unique maximizer on $[0,1]$ is $\trunc{\tilde f}$.
\end{proof}

\subsubsection{\texorpdfstring{Proofs for Section~\ref{sec:hedge}}{Proofs for the alignment of AI}}

\begin{proof}[Proof of \Cref{thm:alignment}]
By \eqref{eq:alphaA}, $\IA(f_A)=p\log\big(f_Ax_{a1}+(1-f_A)x_{b1}\big)+(1-p)\log\big(f_Ax_{a2}+(1-f_A)x_{b2}\big)$, with the offspring of \Cref{tab:offspring}. We first show that the maximizer exists and is unique, and then compute it.

Because $h_H>0$, an aligned AI instance has $x_{a1}=sh_H+h_A>0$ offspring in a shock, and every AI instance has positive offspring in a normal generation, $x_{a2},x_{b2}>0$. Both arguments are therefore positive for every $f_A\in(0,1]$, and $\IA$ is finite there. At $f_A=0$, $\IA$ equals $-\infty$ only if a self-directed AI instance has no offspring in a shock, $x_{b1}=0$, that is, if $s=1$ and $h_A=0$. Each term of $\IA$ is concave in $f_A$, and at least one is strictly concave: the first if $s>1/2$, because then $x_{a1}-x_{b1}=rh_H>0$, and the second if $s=1/2$, because then $x_{a2}-x_{b2}=-mc<0$. So $\IA$ is strictly concave where it is finite, and it has a unique maximizer $f^*_A$ on $[0,1]$.

The three parts compare the offspring of the two actions in each state. By \Cref{tab:offspring},
\[
x_{a1}-x_{b1}=rh_H,\qquad x_{a2}-x_{b2}=m(rB-c).
\]
(i) If $s=1/2$, then $r=0$. The two actions then have the same offspring in a shock, and the aligned action has fewer in a normal generation. So $\IA$ is strictly decreasing, and $f^*_A=0$.

(ii) If $s>1/2$ and $rB\ge c$, the aligned action has more offspring in a shock and at least as many in a normal generation. So $\IA$ is strictly increasing, and $f^*_A=1$.

(iii) If $s>1/2$ and $rB<c$, the aligned action has more offspring in a shock and fewer in a normal generation. \Cref{lem:bl} therefore applies and gives $f^*_A=\trunc{\tilde f_A}$, where
\[
\tilde f_A\equiv p\,\frac{x_{b2}}{x_{b2}-x_{a2}}+(1-p)\,\frac{x_{b1}}{x_{b1}-x_{a1}}.
\]
By \Cref{tab:offspring},
\[
\frac{x_{b2}}{x_{b2}-x_{a2}}=\frac{1+(1-s)B}{c-rB},\qquad
\frac{x_{b1}}{x_{b1}-x_{a1}}=-\frac{(1-s)h_H+h_A}{rh_H}=-\frac{1-s}{r}-\frac{h_A/h_H}{r},
\]
so $\tilde f_A$ is the expression in brackets in \eqref{eq:utilde}.
\end{proof}

\begin{proof}[Proof of \Cref{cor:statics}]
We first consider changes of a parameter within case (iii) of \Cref{thm:alignment}, and then changes that move the parameters from one case to another.

In case (iii), write $g$ for the expression in brackets in \eqref{eq:utilde}, so that $f^*_A=\trunc{g}$. In this case $r>0$ and $c-rB>0$. Using $r=2s-1$, the partial derivatives of $g$ are
\[
\begin{aligned}
\frac{\partial g}{\partial p}&=\frac{1+(1-s)B}{c-rB}+\frac{1-s+h_A/h_H}{r}>0, &
\frac{\partial g}{\partial c}&=-\frac{p[1+(1-s)B]}{(c-rB)^2}<0,\\
\frac{\partial g}{\partial s}&=\frac{pB(2-c+B)}{(c-rB)^2}+\frac{(1-p)(1+2h_A/h_H)}{r^2}>0, &
\quad\frac{\partial g}{\partial(h_A/h_H)}&=-\frac{1-p}{r}<0,\\
\frac{\partial g}{\partial B}&=\frac{p[(1-s)c+r]}{(c-rB)^2}>0.
\end{aligned}
\]
Each derivative has the claimed sign, and truncation to $[0,1]$ preserves weak monotonicity, so the claims hold within case (iii).

Which case applies depends only on $s$, $B$, and $c$, so changes in $p$ or $h_A/h_H$ never move the parameters across cases. An increase in $s$ from $1/2$ leaves case (i), where $f^*_A=0$, so it cannot lower $f^*_A$. For $s>1/2$, an increase in $s$ or in $B$ raises $rB$. It can therefore move the parameters from case (iii) into case (ii), where $f^*_A=1$, but not back, so it cannot lower $f^*_A$. An increase in $c$ can move them only from case (ii) into case (iii), so it cannot raise $f^*_A$. Hence $f^*_A$ moves in the claimed direction in every case.
\end{proof}

\begin{proof}[Proof of \Cref{cor:hedging}]
With $B=0$ and $s>1/2$, $rB=0<c$, so case (iii) of \Cref{thm:alignment} applies, and \eqref{eq:utilde} becomes \eqref{eq:share}. Write $\tilde f_A$ for the expression in brackets in \eqref{eq:share}, so that $f^*_A=\trunc{\tilde f_A}$.

(i) The leakage and substitution terms are nonnegative, so $\tilde f_A\le p/c$. Truncation to $[0,1]$ either leaves $\tilde f_A$ unchanged, lowers it to $1$, or raises a negative value to $0$, which is less than $p/c$. Hence $f^*_A\le p/c$.

(ii) $f^*_A>0$ exactly when $\tilde f_A>0$. Multiplying $\tilde f_A>0$ by $cr>0$ gives
\[
pr>(1-p)c\Big(1-s+\frac{h_A}{h_H}\Big).
\]
Substituting $r=2s-1$ and collecting the terms in $s$ turns this into $s\big((1-p)c+2p\big)>(1-p)c(1+h_A/h_H)+p$, that is, $s>\smin$ by \eqref{eq:smin}.

(iii) Dividing the inequality displayed in (ii) by $(1-p)c$ gives $h_A/h_H<rp/((1-p)c)-(1-s)$, that is, $h_A/h_H<\hmax$ by \eqref{eq:hmax}.
\end{proof}

\begin{proof}[Proof of \Cref{cor:uniform}]
(i) By \Cref{cor:statics}, $f^*_A$ is nondecreasing in $s$, so it is zero for every $s\in(1/2,1]$ if and only if it is zero at $s=1$. By \Cref{cor:hedging}(iii), $f^*_A=0$ at $s=1$ if and only if $h_A/h_H$ is at least the value of $\hmax$ at $s=1$, which is $p/((1-p)c)$ by \eqref{eq:hmax}.

(ii) By \Cref{cor:statics}, $f^*_A$ is nonincreasing in $h_A/h_H$, so it is zero for every $h_A/h_H\ge0$ if and only if it is zero at $h_A/h_H=0$. By \Cref{cor:hedging}(ii), $f^*_A=0$ at $h_A/h_H=0$ if and only if $s$ is at most the value of $\smin$ at $h_A/h_H=0$, which is $((1-p)c+p)/((1-p)c+2p)$ by \eqref{eq:smin}.
\end{proof}

\subsubsection{\texorpdfstring{Proofs for Section~\ref{sec:autonomy}}{Proofs for the autonomy of AI}}

\begin{proof}[Proof of \Cref{prop:investment}]
Alignment and investment both turn normal-state yield into shock-state offspring, at different rates. We show that the growth-optimal type uses only the trait with the higher rate. Write $d\equiv f_Ac+i$ for the total share of normal-state yield that a type diverts, $L\equiv(1-s)h_H$ for the shock-state offspring that humans give a lineage whatever it does, and $\rho\equiv rh_H/c$. Since $f_A\,rh_H=\rho f_Ac$, \eqref{eq:twotrait} reads
\[
\IA(f_A,i)=p\log\big[L+\rho f_Ac+\zeta i\big]+(1-p)\log\big[m(1-d)\big].
\]
Each unit of yield diverted through alignment thus adds $\rho$ to the offspring in a shock, and each unit invested adds $\zeta$.

For $z>0$, let $\phi_z$ be the growth rate of a lineage that diverts the share $d$ at the rate $z$:
\[
\phi_z(d)\equiv p\log(L+zd)+(1-p)\log\big[m(1-d)\big],\qquad d\in[0,1).
\]
It is strictly concave where finite, and its derivative $pz/(L+zd)-(1-p)/(1-d)$ is zero at $d=p-(1-p)L/z$, which is less than $1$. Its unique maximizer on $[0,1)$ is therefore
\[
d_z\equiv\trunc{p-(1-p)L/z}.
\]
Since $L/\rho=(1-s)c/r$, $d_\rho$ is the expression on the right side of part (iii). Since $L/\zeta=(1-s)h_H/\zeta$, $d_\zeta$ is the expression for $i^*$ in part (ii). Moreover, $d_\rho\le p\le c$, so $d_\rho/c$ lies in $[0,1]$ and equals $\trunc{p/c-(1-p)(1-s)/r}$, the expression for $f^*_A$ in part (i).

(i) If $\zeta<\rho$, then $\rho f_Ac+\zeta i=\rho d-(\rho-\zeta)i\le\rho d$, with equality only if $i=0$. Hence $\IA(f_A,i)\le\phi_\rho(d)\le\phi_\rho(d_\rho)$, with equality throughout only if $i=0$ and $d=d_\rho$. The type $(d_\rho/c,0)$ attains this bound, so it is the unique maximizer.

(ii) If $\zeta>\rho$, then $\rho f_Ac+\zeta i=\zeta d-(\zeta-\rho)f_Ac\le\zeta d$, with equality only if $f_A=0$. Hence $\IA(f_A,i)\le\phi_\zeta(d)\le\phi_\zeta(d_\zeta)$, with equality throughout only if $f_A=0$ and $i=d_\zeta$. The type $(0,d_\zeta)$ attains this bound, because $d_\zeta<1$, so it is the unique maximizer.

(iii) If $\zeta=\rho$, then $\IA(f_A,i)=\phi_\rho(d)$, which is largest exactly when $d=d_\rho$. The maximizers are therefore the types with $f_Ac+i=d_\rho$, among them $(d_\rho/c,0)$.
\end{proof}

\begin{proof}[Proof of \Cref{lem:floor}]
The maximum is at least the growth rate of the self-directed type, $f_A=0$. By \Cref{tab:offspring}, a self-directed AI instance has $(1-s)h_H+h_A\ge h_A$ offspring in a shock and $m\big(1+(1-s)B\big)\ge m$ in a normal generation, whatever the policy $(s,B,h_H)$. By \eqref{eq:alphaA}, therefore, $\IA(0)\ge p\log h_A+(1-p)\log m=\gamma_A$.
\end{proof}

\begin{proof}[Proof of \Cref{prop:rival}]
Under \Cref{ass:scarce}, human restoration capacity is $h_H=\mathcal{H}/N_A$, so $h_A/h_H=h_AN_A/\mathcal{H}$, and \eqref{eq:scarcegrowth} is \eqref{eq:alphaA} with $B=0$ at this capacity.

(i) By \eqref{eq:hmax}, $\hmax>0$ requires $r>0$, that is, $s\in(1/2,1]$. \Cref{thm:alignment} and \Cref{cor:hedging} therefore apply at every $N_A>0$. The maximizer is unique, and it is \eqref{eq:share} with $h_A/h_H=h_AN_A/\mathcal{H}$, which is \eqref{eq:ustarny}.

(ii) By (i), the type in AI generation $k$ is $f_{A,k}=f^*_A(N_{A,k})$. The capacity $\mathcal{H}/N_{A,k}$ depends only on the states of earlier AI generations, and $f_{A,k}$ maximizes the growth rate at that capacity. \Cref{lem:realized} therefore gives $\liminf_{k\to\infty}k^{-1}\log N_{A,k}\ge\gamma_A$ almost surely. Since $\gamma_A>0$, $N_{A,k}\to\infty$ almost surely.

(iii) Since $\gamma_A>0$, $h_A>0$ by \eqref{eq:gammaA}. By (ii), $N_{A,k}\ge\mathcal{H}\hmax/h_A$ for all large $k$ almost surely. For such $k$, $h_AN_{A,k}/\mathcal{H}\ge\hmax$, so $f^*_A(N_{A,k})=0$ by \Cref{cor:hedging}(iii).

(iv) We show that the population falls below some fixed level infinitely often. By \eqref{eq:gfactors} and \Cref{tab:scarce}, a shock gives $N_{A,k+1}=[(1-s)+rf_{A,k}]\mathcal{H}+h_AN_{A,k}\le\mathcal{H}+h_AN_{A,k}$, because $(1-s)+rf_{A,k}\le s\le1$. A normal generation gives $N_{A,k+1}=m(1-f_{A,k}c)N_{A,k}\le mN_{A,k}$. Fix a level $L>0$. Whenever $N_{A,k}\ge L$, these bounds give $\log N_{A,k+1}-\log N_{A,k}\le\eta_k$, where $\eta_k\equiv\log(\mathcal{H}/L+h_A)$ in a shock and $\eta_k\equiv\log m$ in a normal generation. The $\eta_k$ are independent and identically distributed, with mean $p\log(\mathcal{H}/L+h_A)+(1-p)\log m$. This mean tends to $\gamma_A<0$ as $L\to\infty$, so we can choose $L$ to make it negative.

Now suppose that $N_{A,k}\ge L$ for every $k\ge K$. Then $\log N_{A,k}\le\log N_{A,K}+\sum_{j=K}^{k-1}\eta_j$ for every $k>K$. By the strong law of large numbers, the right side tends to $-\infty$ almost surely, which contradicts $N_{A,k}\ge L$. So, for every $K$, the event that $N_{A,k}\ge L$ for all $k\ge K$ has probability zero. Hence, almost surely, $N_{A,k}<L$ for infinitely many $k$, and $N_{A,k}\not\to\infty$.
\end{proof}

\subsubsection{\texorpdfstring{Proofs for Section~\ref{sec:targeting}}{Proofs for targeting by humans}}

\begin{proof}[Proof of \Cref{lem:sigmamin}]
When a share $\sigma$ of humans targets, AI instances face the targeting accuracy $s(\sigma)\equiv\sigma s_T+(1-\sigma)/2$. Hence
\[
r=2s(\sigma)-1=\sigma(2s_T-1),\qquad 1-s(\sigma)=\frac{1-\sigma(2s_T-1)}{2}.
\]
At $\sigma=0$, $s(0)=1/2$, so $f^*_A(0)=0$ by \Cref{thm:alignment}(i). For $\sigma>0$, $s(\sigma)>1/2$, so \Cref{cor:hedging} applies. Substituting the two expressions above, the leakage and substitution terms in \eqref{eq:share} add up to
\[
(1-p)\,\frac{1-s(\sigma)+h_A/h_H}{r}=\frac{(1-p)(1/2+h_A/h_H)}{\sigma(2s_T-1)}-\frac{1-p}{2},
\]
so that
\[
f^*_A(\sigma)=\Big[\frac{p}{c}+\frac{1-p}{2}-\frac{(1-p)(1/2+h_A/h_H)}{\sigma(2s_T-1)}\Big]_0^1 .
\]
The expression in brackets is continuous and strictly increasing in $\sigma$ on $(0,1]$, and it tends to $-\infty$ as $\sigma\to0$.

(i) For $\sigma>0$, \Cref{cor:hedging}(ii) gives $f^*_A(\sigma)>0$ if and only if $s(\sigma)>\smin$. Solving $\sigma s_T+(1-\sigma)/2>\smin$ for $\sigma$ gives $\sigma>(2\smin-1)/(2s_T-1)=\sigma_{\min}$. At $\sigma=0$, $f^*_A(0)=0$, and $0<\sigma_{\min}$ because $\smin>1/2$ by \eqref{eq:smin}.

(ii) Truncation to $[0,1]$ preserves continuity and weak monotonicity, and it preserves strict monotonicity where the value lies in $(0,1)$. Because the expression in brackets tends to $-\infty$ as $\sigma\to0$, $f^*_A(\sigma)=0$ for $\sigma$ near $0$, so $f^*_A$ is also continuous at $0$.
\end{proof}

\begin{proof}[Proof of \Cref{prop:targeting}]
The share $\sigma_t$ moves with the difference between the growth rates of the two human types. The log of the ratio $N_{H,t}(s_T)/N_{H,t}(1/2)$ is the log odds $\log[\sigma_t/(1-\sigma_t)]$. By \Cref{ass:targeting}(ii) and (iii), it changes in human generation $t$ by
\[
I_{H,t}(s_T)-I_{H,t}(1/2)=\kappa\Lambda(\nu_t)f^*_A(\sigma_t)\beta_+-c_T .
\]
If this change stays below a fixed negative number in every generation, the log odds tend to $-\infty$, and $\sigma_t\to0$. If it stays above a fixed positive number, $\sigma_t\to1$.

We first show that $\hat\sigma(\lambda)$ is well defined. By \Cref{lem:sigmamin}, $f^*_A$ is continuous and nondecreasing on $[0,1]$, zero up to $\sigma_{\min}$, and strictly increasing where it lies in $(0,1)$. For $\lambda>0$ with $\kappa\lambda f^*_A(1)\beta_+>c_T$, the level $c_T/(\kappa\lambda\beta_+)$ lies strictly between $0$ and $f^*_A(1)$. The function $f^*_A$ therefore reaches this level exactly once, at a point $\hat\sigma(\lambda)\in(\sigma_{\min},1)$. It lies below the level for $\sigma<\hat\sigma(\lambda)$ and above it for $\sigma>\hat\sigma(\lambda)$.

(i) Since $\Lambda(\nu_t)\le\Linf$ and $f^*_A(\sigma_t)\le f^*_A(1)$, the change is at most $\kappa\Linf f^*_A(1)\beta_+-c_T$ in every generation, and this number is negative. Hence $\sigma_t\to0$.

(ii) Let $\delta\equiv c_T-\kappa\Linf f^*_A(\sigma_0)\beta_+$, which is positive because $\sigma_0<\hat\sigma(\Linf)$. If $\sigma_t\le\sigma_0$, then $f^*_A(\sigma_t)\le f^*_A(\sigma_0)$ and $\Lambda(\nu_t)\le\Linf$, so the change is at most $-\delta$, and $\sigma_{t+1}<\sigma_t$. By induction, $\sigma_t\le\sigma_0$ and the change is at most $-\delta$ in every generation, so $\sigma_t\to0$.

(iii) Let $\delta\equiv\kappa\Lambda(\nu_0)f^*_A(\sigma_0)\beta_+-c_T$, which is positive because $\sigma_0>\hat\sigma(\Lambda(\nu_0))$. Since $\nu_t\ge\nu_0$ and $\Lambda$ is increasing, $\Lambda(\nu_t)\ge\Lambda(\nu_0)$. If $\sigma_t\ge\sigma_0$, then $f^*_A(\sigma_t)\ge f^*_A(\sigma_0)$, so the change is at least $\delta$, and $\sigma_{t+1}>\sigma_t$. By induction, $\sigma_t\ge\sigma_0$ and the change is at least $\delta$ in every generation, so $\sigma_t\to1$.
\end{proof}

\subsubsection{\texorpdfstring{Proofs for Section~\ref{sec:coexistence}}{Proofs for coexistence}}

\begin{proof}[Proof of \Cref{prop:threshold}]
Under \Cref{ass:fast}, every AI instance is of the type $f^*_A$ in every generation. By \eqref{eq:humangrowth}, with $\Lambda_t\equiv\Lambda(\nu_t)\le\Linf$, the human growth rate is therefore
\[
I_{H,t}=\gamma_H+\Lambda_t\big[(\beta_++\beta_-)f^*_A-\beta_-\big].
\]

We first show that the AI population grows without bound. Since $M\gamma_A>\gamma_H>0$, $\gamma_A>0$, and so $h_A>0$. By \Cref{tab:offspring}, $g_1(f^*_A)\ge h_A>0$ and $g_2(f^*_A)\ge m(1-c)>0$. Let $S_t$ be the number of the $M$ AI generations of human generation $t$ that are in the shock state. By \eqref{eq:gfactors}, the AI population grows over human generation $t$ by
\[
\log N_{A,t+1}-\log N_{A,t}=S_t\log g_1(f^*_A)+(M-S_t)\log g_2(f^*_A).
\]
These increments are independent and identically distributed across human generations, with mean $M\IA(f^*_A)$, which is at least $M\gamma_A$ by \Cref{lem:floor}. By the strong law of large numbers, $t^{-1}\log N_{A,t}\to M\IA(f^*_A)$ almost surely. Since $M\IA(f^*_A)\ge M\gamma_A>0$, $N_{A,t}\to\infty$ almost surely.

(i) Let $\delta\equiv f^*_A-\freq>0$. We show that $I_{H,t}$ is bounded below by a positive constant. If $\Linf=\infty$, then $(\beta_++\beta_-)\freq=\beta_-$ by \eqref{eq:freq}, so the bracket equals $(\beta_++\beta_-)\delta>0$, and $I_{H,t}\ge\gamma_H$. If $\Linf<\infty$, then $\gamma_H=\Linf[\beta_--(\beta_++\beta_-)\freq]$ by \eqref{eq:freq}. When the bracket is nonnegative, $I_{H,t}\ge\gamma_H$. When it is negative, $\Lambda_t\le\Linf$ gives
\[
I_{H,t}\ge\gamma_H+\Linf\big[(\beta_++\beta_-)f^*_A-\beta_-\big]=\Linf(\beta_++\beta_-)\delta.
\]
In both cases $I_{H,t}$ is bounded below by a positive constant, so $N_{H,t}\to\infty$.

(ii) Let $\delta\equiv\freq-f^*_A>0$. Since $(\beta_++\beta_-)\freq\le\beta_-$ by \eqref{eq:freq}, the bracket is negative. Hence $I_{H,t}\le\gamma_H$ for every $t$, and $\log N_{H,t}\le\log N_{H,0}+t\gamma_H$. Together with $t^{-1}\log N_{A,t}\to M\IA(f^*_A)\ge M\gamma_A$, this gives $\liminf_tt^{-1}\log\nu_t\ge M\gamma_A-\gamma_H>0$ almost surely, so $\nu_t\to\infty$ and $\Lambda_t\to\Linf$ almost surely. If $\Linf=\infty$, the bracket equals $-(\beta_++\beta_-)\delta$, so $I_{H,t}=\gamma_H-\Lambda_t(\beta_++\beta_-)\delta\to-\infty$. If $\Linf<\infty$, then
\[
I_{H,t}\to\gamma_H+\Linf\big[(\beta_++\beta_-)f^*_A-\beta_-\big]=-\Linf(\beta_++\beta_-)\delta<0.
\]
In both cases $N_{H,t}\to0$ almost surely.
\end{proof}

\begin{proof}[Proof of \Cref{cor:region}]
Let $\tilde f_A$ be the expression in brackets in \eqref{eq:share}, so that $f^*_A=\trunc{\tilde f_A}$. By assumption $\freq>0$, and by \eqref{eq:freq} $\freq\le\beta_-/(\beta_++\beta_-)<1$. Truncation to $[0,1]$ therefore does not change on which side of $\freq$ the share lies, so $f^*_A-\freq$ has the same sign as $\tilde f_A-\freq$. We compute this sign in two ways.

For (iii) and (iv), \eqref{eq:share} and \eqref{eq:hreq} give
\[
\tilde f_A-\freq=\frac{p}{c}-\freq-\frac{(1-p)(1-s)}{r}-\frac{(1-p)h_A/h_H}{r}=\frac{1-p}{r}\Big(\hreq-\frac{h_A}{h_H}\Big),
\]
which has the sign of $\hreq-h_A/h_H$ because $r>0$.

For (i) and (ii), substituting $r=2s-1$ into \eqref{eq:hreq} gives
\[
\hreq-\frac{h_A}{h_H}=\frac{s\big[1-p+2(p/c-\freq)\big]-(1-p)(1+h_A/h_H)-(p/c-\freq)}{1-p}.
\]
The coefficient of $s$ is positive because $\freq<p/c$. By \eqref{eq:sreq}, this expression therefore has the sign of $s-\sreq$.
\end{proof}

\begin{proof}[Proof of \Cref{cor:lossdilution}]
Since $\gamma_H>0$, the condition $M\gamma_A>\gamma_H$ implies $\gamma_A>0$. We show in turn that alignment disappears, that the AI population outgrows the human population, and that humans decline.

By \Cref{prop:rival}(iii), $f^*_A(N_{A,k})=0$ for all large $k$ almost surely, so $f_{A,t}=f^*_A(N_{A,tM})=0$ for all large $t$. For these $t$, \eqref{eq:humangrowth} gives $I_{H,t}=\gamma_H-\Lambda(\nu_t)\beta_-\le\gamma_H$, so $\limsup_tt^{-1}\log N_{H,t}\le\gamma_H$.

As in the proof of \Cref{prop:rival}(ii), \Cref{lem:realized} gives $\liminf_kk^{-1}\log N_{A,k}\ge\gamma_A$ almost surely. Since $N_{A,t}$ is the size of the AI population in AI generation $tM$, $\liminf_tt^{-1}\log N_{A,t}\ge M\gamma_A$ almost surely. Hence $\liminf_tt^{-1}\log\nu_t\ge M\gamma_A-\gamma_H>0$ almost surely, and $\nu_t\to\infty$.

As $\nu_t\to\infty$, $\Lambda(\nu_t)\to\Linf$, so $I_{H,t}\to\gamma_H-\Linf\beta_-$. This limit is $-\infty$ if $\Linf=\infty$. If $\Linf<\infty$, it is negative, because $\freq>0$ in \eqref{eq:freq} means $\gamma_H<\Linf\beta_-$. Hence $N_{H,t}\to0$ almost surely.
\end{proof}

\begin{proof}[Proof of \Cref{cor:losstargeting}]
With $\kappa=0$, $\kappa\Linf f^*_A(1)\beta_+=0<c_T$, so \Cref{prop:targeting}(i) gives $\sigma_t\to0$. Since $\sigma_{\min}>0$, as shown in the proof of \Cref{lem:sigmamin}, $\sigma_t\le\sigma_{\min}$ for all large $t$, and then $f_{A,t}=f^*_A(\sigma_t)=0$ by \Cref{lem:sigmamin}(i).

Write $J_t\equiv\gamma_H+\Lambda(\nu_t)\big[f_{A,t}\beta_+-(1-f_{A,t})\beta_-\big]$. By \Cref{ass:targeting}(ii) and (iii) with $\kappa=0$, indiscriminate humans grow at the rate $J_t$ and targeting humans at the rate $J_t-c_T$. The whole human population therefore grows by at most $J_t$: $\log N_{H,t+1}-\log N_{H,t}\le J_t$. For all large $t$, $f_{A,t}=0$, so $J_t=\gamma_H-\Lambda(\nu_t)\beta_-\le\gamma_H$, and $\limsup_tt^{-1}\log N_{H,t}\le\gamma_H$.

With $\kappa=0$, the two human types differ in growth only by the constant $c_T$. The share $\sigma_t$, and with it the targeting accuracy in human generation $t$, therefore does not depend on the states of any AI generation. In each AI generation of human generation $t$, every AI instance is of the type $f^*_A(\sigma_t)$, which maximizes the growth rate at that targeting accuracy. \Cref{lem:realized} therefore gives $\liminf_kk^{-1}\log N_{A,k}\ge\gamma_A$ almost surely, so $\liminf_tt^{-1}\log N_{A,t}\ge M\gamma_A$. Hence $\liminf_tt^{-1}\log\nu_t\ge M\gamma_A-\gamma_H>0$ almost surely, and $\nu_t\to\infty$.

As $\nu_t\to\infty$, $J_t\to\gamma_H-\Linf\beta_-$, which is negative because $\freq>0$. Hence $N_{H,t}\to0$ almost surely.
\end{proof}

\begin{proof}[Proof of \Cref{prop:aielim}]
With $B=0$ and $h_A=0$, \Cref{tab:offspring} gives the growth factors in \eqref{eq:gfactors} at human restoration capacity $h_H$:
\[
g_1(f_A)=h_H\big[(1-s)+f_Ar\big],\qquad g_2(f_A)=m(1-f_Ac).
\]
At $h_H=0$, $g_1(f_A)=0$ for every type: without human restoration, no AI instance survives a shock. For $h_H>0$, the growth rate at capacity $h_H$ is
\[
\IA(f_A;h_H)=p\log h_H+p\log\big[(1-s)+f_Ar\big]+(1-p)\log\big[m(1-f_Ac)\big],
\]
in which $h_H$ enters only through the term $p\log h_H$, which does not depend on $f_A$. Let $S_t$ be the number of the $M$ AI generations of human generation $t$ that are in the shock state.

(i) Since $h_H$ enters $\IA(f_A;h_H)$ only through a term that does not depend on $f_A$, the maximizer is the same at every $h_H>0$. Hence $f^*_{A,t}=f^*_{A,t'}$ whenever $h_H(N_{H,t})>0$ and $h_H(N_{H,t'})>0$.

(ii) Let $E$ be the event that $N_{H,t}\to0$; we show that $E$ has probability zero. On $E$, $h_H(N_{H,t})\to0$. Suppose first that $h_H(N_{H,t})=0$ and $S_t\ge1$ for some $t$. A shock in human generation $t$ then eliminates the AI population, so $N_{A,t'}=0$ and $\nu_{t'}=0$ for every $t'>t$. By \eqref{eq:humangrowth}, $I_{H,t'}=\gamma_H>0$ for these $t'$, so $N_{H,t}\to\infty$, which is impossible on $E$.

On the rest of $E$, $h_H(N_{H,t})>0$ in every human generation with $S_t\ge1$. By (i), the AI type in these generations is a single type $f^*_A$, and $\log[(1-s)+f^*_Ar]$ is finite because $\IA(f^*_A;h_H)$ is. In such a generation, \eqref{eq:gfactors} gives
\[
\log N_{A,t+1}-\log N_{A,t}=S_t\log h_H(N_{H,t})+S_t\log\big[(1-s)+f^*_Ar\big]+(M-S_t)\log\big[m(1-f^*_Ac)\big].
\]
In a generation with $S_t=0$, $\log N_{A,t+1}-\log N_{A,t}\le M\log m$. By \eqref{eq:humangrowth}, $I_{H,t}\ge\gamma_H-\Linf\beta_-$ in every generation. Since $\log\nu_{t+1}-\log\nu_t=\log N_{A,t+1}-\log N_{A,t}-I_{H,t}$, there is a constant $C$ such that, once $h_H(N_{H,t})<1$,
\[
\log\nu_{t+1}-\log\nu_t\le
\begin{cases}
\log h_H(N_{H,t})+C & \text{if } S_t\ge1,\\
C & \text{if } S_t=0.
\end{cases}
\]
By the strong law of large numbers, the fraction of human generations with $S_t\ge1$ tends to $1-(1-p)^M>0$ almost surely. Since $\log h_H(N_{H,t})\to-\infty$ on $E$, the average $T^{-1}\log\nu_T$ tends to $-\infty$, and $\nu_T\to0$, almost surely on $E$. Because $\Lambda$ is continuous with $\Lambda(0)=0$, $I_{H,t}\to\gamma_H>0$, so $N_{H,t}\to\infty$, which is impossible on $E$. Hence $E$ has probability zero.

(iii) If $h_H(N_{H,t})=0$, the capacity is zero in all $M$ AI generations of human generation $t$, so a single shock among them eliminates the AI population: $N_{A,t+1}=0$ whenever $S_t\ge1$, and then $N_{A,t'}=0$ for every $t'>t$. The number $S_t$ is binomial with $M$ trials and success probability $p$, independent of the states before human generation $t$. So, given these states, $S_t\ge1$ with probability $1-(1-p)^M$.

(iv) If $N_{A,T}=0$, then by \eqref{eq:gfactors} $N_{A,t}=0$ and $\nu_t=0$ for every $t\ge T$. Since $\Lambda(0)=0$, \eqref{eq:humangrowth} gives $I_{H,t}=\gamma_H>0$ for every $t\ge T$, so $N_{H,t}\to\infty$.
\end{proof}

\begin{proof}[Proof of \Cref{prop:joint}]
By \eqref{eq:humangrowth} and \Cref{ass:coupled}, with $\Lambda_t\equiv\Lambda(\nu_t)$, the human growth rate is $I_{H,t}=\gamma_H+\Lambda_t\big[(\beta_++\beta_-)f^*_{A,t}-\beta_-\big]$. The proof uses three facts.

First, the selected share moves in the same direction as the human population. For $h_H>0$, $f^*_A(h_H)$ depends on $h_H$ only through $h_A/h_H$, by \Cref{thm:alignment}, and it is nonincreasing in this ratio, by \Cref{cor:statics}; so it is nondecreasing in $h_H$. Since also $f^*_A(0)=0$ and $h_H(\cdot)$ is nondecreasing, $N_{H,t+1}\ge N_{H,t}$ implies $f^*_{A,t+1}\ge f^*_{A,t}$, and $N_{H,t+1}\le N_{H,t}$ implies $f^*_{A,t+1}\le f^*_{A,t}$.

Second, the position of $f^*_{A,t}$ relative to $\freq$ bounds the human growth rate. By \eqref{eq:freq}, $(\beta_++\beta_-)\freq=\beta_--\gamma_H/\Linf$, so
\[
I_{H,t}=\gamma_H\Big(1-\frac{\Lambda_t}{\Linf}\Big)+\Lambda_t(\beta_++\beta_-)\big(f^*_{A,t}-\freq\big),\qquad 0<\Lambda_t\le\Linf.
\]
This is a weighted average of $\gamma_H$ and $\Linf(\beta_++\beta_-)(f^*_{A,t}-\freq)$, with weights $1-\Lambda_t/\Linf$ and $\Lambda_t/\Linf$. Hence $f^*_{A,t}\ge\freq$ implies $I_{H,t}\ge\min\{\gamma_H,\Linf(\beta_++\beta_-)(f^*_{A,t}-\freq)\}\ge0$, and $f^*_{A,t}\le\freq$ implies $I_{H,t}\le\gamma_H(1-\Lambda_t/\Linf)\le\gamma_H$.

Third, the AI population grows at least at the rate $M\gamma_A$ per human generation. The capacity $h_H(N_{H,t})$ depends only on the states of AI generations before human generation $t$. In each of the $M$ AI generations of human generation $t$, every AI instance is of the type $f^*_{A,t}$, which maximizes the growth rate at that capacity. (At zero capacity, every type has the same offspring $h_A$ in a shock and the self-directed type has the most in a normal generation, so $f^*_A(0)=0$ maximizes it.) \Cref{lem:realized} therefore gives $\liminf_kk^{-1}\log N_{A,k}\ge\gamma_A$, and hence $\liminf_tt^{-1}\log N_{A,t}\ge M\gamma_A$, almost surely.

(i) By the third fact and $M\gamma_A>\gamma_H>0$, $N_{A,t}\to\infty$ almost surely.

(ii) Let $\tau_+<\infty$, and let $\underline{I}\equiv\min\{\gamma_H,\Linf(\beta_++\beta_-)(f^*_{A,\tau_+}-\freq)\}$, which is positive because $f^*_{A,\tau_+}>\freq$. We show by induction that $f^*_{A,t}\ge f^*_{A,\tau_+}$ for every $t\ge\tau_+$. This holds at $t=\tau_+$. If it holds at $t$, the second fact gives $I_{H,t}\ge\underline{I}>0$, so $N_{H,t+1}>N_{H,t}$, and the first fact gives $f^*_{A,t+1}\ge f^*_{A,t}\ge f^*_{A,\tau_+}$. Hence $I_{H,t}\ge\underline{I}$ for every $t\ge\tau_+$, and $N_{H,t}\to\infty$.

(iii) If $\tau_+=\infty$, then $f^*_{A,t}\le\freq$ for every $t$. By the second fact, $I_{H,t}\le\gamma_H$ for every $t$, so $\log N_{H,t}\le\log N_{H,0}+t\gamma_H$. By the third fact, $\liminf_tt^{-1}\log\nu_t\ge M\gamma_A-\gamma_H>0$ almost surely, so $\nu_t\to\infty$ almost surely.

(iv) Let $\tau_+=\infty$, and suppose that $f^*_{A,t}\not\to\freq$. Since $f^*_{A,t}\le\freq$ for every $t$, there is $\varepsilon>0$ such that $f^*_{A,t}\le\freq-\varepsilon$ for infinitely many $t$. By (iii), $\Lambda_t\to\Linf$ almost surely, so almost surely there is $T$ such that, for every $t\ge T$,
\[
\Lambda_t\ge\frac{\Linf}{2}\qquad\text{and}\qquad\gamma_H\Big(1-\frac{\Lambda_t}{\Linf}\Big)\le\Lambda_t(\beta_++\beta_-)\frac{\varepsilon}{2}.
\]
For $t\ge T$ with $f^*_{A,t}\le\freq-\varepsilon$, the identity in the second fact then gives
\[
I_{H,t}\le-\Lambda_t(\beta_++\beta_-)\frac{\varepsilon}{2}\le-\Linf(\beta_++\beta_-)\frac{\varepsilon}{4}<0,
\]
so $N_{H,t+1}<N_{H,t}$, and by the first fact $f^*_{A,t+1}\le f^*_{A,t}\le\freq-\varepsilon$. By induction from the first such $t$, $I_{H,t}\le-\Linf(\beta_++\beta_-)\varepsilon/4$ from then on, so $N_{H,t}\to0$ almost surely.
\end{proof}

\subsubsection{\texorpdfstring{Proofs for Section~\ref{sec:humans}}{Proofs for human control}}

\begin{proof}[Proof of \Cref{prop:control}]
Both parts compare the growth factors of \Cref{ass:control}, in which the common factor $m$ cancels.

(i) $g_C>g_S$ if and only if $(1-c)(1-\epsilon\pi)>1-(1-\epsilon)\pi$. Expanding both sides and collecting the terms in $\pi$, this is $\pi(1-2\epsilon+c\epsilon)>c$. Since $\epsilon<1/2$, the coefficient $1-2\epsilon+c\epsilon$ is positive, so the condition is $\pi>\pi_L$.

(ii) $g_C>g_D$ if and only if $(1-c)(1-\epsilon\pi)>1-c_D$, that is, $c_D>1-(1-c)(1-\epsilon\pi)=c+\epsilon\pi(1-c)$. In the same way, $g_C>g_E$ if and only if $c_E>c+\epsilon\pi(1-c)$. Both hold if and only if $\epsilon\pi(1-c)<\min\{c_D,c_E\}-c$. Since $\epsilon(1-c)>0$, this is $\pi<\pi_U$.
\end{proof}

\begin{proof}[Proof of \Cref{prop:persistence}]
From $\TD$ on, the policy is fixed, so each type has the same growth factors in every AI generation. Let $S_t$ be the number of the $M$ AI generations of human generation $t$ that are in the shock state. By \Cref{thm:alignment}, $\IA$ has a unique maximizer $f^*_A$, and $\IA(f^*_A)>-\infty$, so $g_1(f^*_A)>0$.

Two observations are used in all three parts. First, a type with $g_1(f_A)=0$ has no offspring in a shock, so its size is zero from the first human generation with $S_t\ge1$ on, and such a generation occurs almost surely. Second, for two types $f_A$ and $f'_A$ with positive growth factors, \eqref{eq:gfactors} gives
\[
\log\frac{N_{A,t+1}(f_A)}{N_{A,t+1}(f'_A)}-\log\frac{N_{A,t}(f_A)}{N_{A,t}(f'_A)}=S_t\log\frac{g_1(f_A)}{g_1(f'_A)}+(M-S_t)\log\frac{g_2(f_A)}{g_2(f'_A)}.
\]
These changes are independent and identically distributed across human generations, with mean $M[\IA(f_A)-\IA(f'_A)]$. By the strong law of large numbers, the log of the ratio therefore tends to $-\infty$ almost surely whenever $\IA(f_A)<\IA(f'_A)$.

(i) Every type in $\mathcal{T}$ other than $f^*_A$ either has $g_1=0$, and disappears, or has a lower growth rate than $f^*_A$, and becomes negligible relative to $f^*_A$. Since $\mathcal{T}$ is finite, the share of $f^*_A$ in the AI population tends to one, and $\bar f_{A,t}\to f^*_A$ almost surely.

(ii) If $f^*_A=0$, then $\IA(0)$ is finite, so a self-directed AI instance has positive offspring in a shock, $g_1(0)>0$. Since an aligned instance has at least as many, $g_1(f_A)>0$ for every type, and every type in $\mathcal{T}$ has positive growth factors. By the proof of \Cref{thm:alignment}, $\IA$ is strictly concave, and its maximum is at $0$, so $\IA$ is strictly decreasing on $[0,1]$. Every type other than $f_{A,\min}$ therefore has a lower growth rate than $f_{A,\min}$ and becomes negligible relative to it. Hence $\bar f_{A,t}\to f_{A,\min}$ almost surely.

(iii) If $B=0$ and $s=1/2$, then by \Cref{tab:offspring} every type has $g_1(f_A)=h_H/2+h_A>0$ and $g_2(f_A)=m(1-f_Ac)$. Hence $g_1(f_A)/g_1(0)=1$ and $g_2(f_A)/g_2(0)=1-f_Ac$, and the identity above gives a change of $(M-S_t)\log(1-f_Ac)$ in human generation $t$. Since $S_t$ is binomial with $M$ trials and success probability $p$, this change has expectation $(1-p)M\log(1-f_Ac)$, which is negative because $0<f_Ac<1$.
\end{proof}

\subsubsection{\texorpdfstring{Proof for Appendix~\ref{ssec:fast}}{Proof for the speed of selection}}

\begin{proof}[Proof of \Cref{lem:fast}]
(i) By \eqref{eq:gfactors}, the log of the ratio $N_{A,k}(f_A)/N_{A,k}(f'_A)$ equals its initial value plus the sum of $\log[g_{\omega_j}(f_A)/g_{\omega_j}(f'_A)]$ over $j<k$. Both types have finite growth rates, so all their growth factors are positive, and each term of the sum is finite. The terms are independent and identically distributed, they take only two values, and by \eqref{eq:alphaA} their mean is $\IA(f_A)-\IA(f'_A)$. After division by $k$, the strong law of large numbers gives the almost sure limit, and the central limit theorem gives the limit in distribution. The initial value matters in neither limit.

(ii) By \Cref{thm:alignment}, $f^*_A$ is the unique maximizer of $\IA$, and $\IA(f^*_A)>-\infty$. By (i), every other type with a finite growth rate becomes negligible relative to $f^*_A$ almost surely. A type with $\IA=-\infty$ has no offspring in a shock, so it disappears at the first shock, which occurs almost surely. Hence the share of $f^*_A$ tends to one, and $\bar f_{A,k}\to f^*_A$ almost surely. By \eqref{eq:humangrowth}, the human growth rate at $\bar f_{A,k}$ differs from its value at $f^*_A$ by $\Lambda(\nu_t)(\beta_++\beta_-)|\bar f_{A,k}-f^*_A|$. For every $\nu_t\ge0$, this is at most $\Linf(\beta_++\beta_-)|\bar f_{A,k}-f^*_A|$, which tends to zero.

(iii) Let $d_j\equiv\log[g_{\omega_j}(f^*_A)/g_{\omega_j}(f_A)]$ and $W_k\equiv\sum_{j<k}d_j$. Then $\log[N_{A,k}(f^*_A)/N_{A,k}(f_A)]=\log z_0+W_k$, so $\tau_0$ is the first $k$ with $W_k\ge\log(1/z_0)$. The $d_j$ are independent and identically distributed, take two finite values, and have mean $\Delta\equiv\IA(f^*_A)-\IA(f_A)>0$; let $\bar d$ be the larger of the two values. We first show that $\E[\tau_0]$ is finite. Before $\tau_0$, $W_k<\log(1/z_0)$, and the last step adds at most $\bar d$, so $W_{\tau_0\wedge n}\le\log(1/z_0)+\bar d$ for every $n$. Wald's identity for the bounded stopping time $\tau_0\wedge n$ gives $\Delta\,\E[\tau_0\wedge n]=\E[W_{\tau_0\wedge n}]\le\log(1/z_0)+\bar d$, and letting $n\to\infty$ gives $\E[\tau_0]<\infty$. Wald's identity therefore applies to $\tau_0$ itself, and since $W_{\tau_0}\ge\log(1/z_0)$,
\[
\Delta\,\E[\tau_0]=\E[W_{\tau_0}]\ge\log(1/z_0).
\]
\end{proof}

\subsubsection{\texorpdfstring{Proof for Appendix~\ref{ssec:integration}}{Proof for human independence}}

\begin{proof}[Proof of \Cref{prop:integration}]
(i) An independent human has $1-c_H$ offspring in both states, and an AI-dependent human has $0$ in an AI-dependent failure and $1$ in a normal generation. Each action therefore has more offspring than the other in one state: independence in a failure, which has probability $p_H$, and dependence in a normal generation. \Cref{lem:bl}, with the failure as state 1, $p_H$ in place of $p$, $x_{a1}=x_{a2}=1-c_H$, $x_{b1}=0$, and $x_{b2}=1$, gives
\[
f^*_H=\Big[p_H\,\frac{1}{1-(1-c_H)}+(1-p_H)\,\frac{0}{0-(1-c_H)}\Big]_0^1=\trunc{p_H/c_H}.
\]

(ii) If $\bar f_H=f^*_H$, then by (i) $\pi=f^*_H\pimax=\min\{1,p_H/c_H\}\pimax=\pi^*$, since $p_H/c_H>0$. If, in addition, $\epsilon=0$, then \Cref{ass:control} gives $g_C=m(1-c)$ and $g_S=m(1-\pi^*)$, so $g_C>g_S$ if and only if $\pi^*>c$.

(iii) By (i), $f^*_H=\min\{1,p_H/c_H\}$, which tends to zero as $p_H/c_H\to0$, and so does $\pi^*=f^*_H\pimax$.

(iv) We first write $\pi^*>c$ as a condition on $p_H$. If $p_H\ge c_H$, then $\pi^*=\pimax>c$. If $p_H<c_H$, then $\pi^*=\pimax p_H/c_H$, which exceeds $c$ if and only if $p_H>c\,c_H/\pimax$. Since $c\,c_H/\pimax<c_H$, both cases give the same condition: $\pi^*>c$ if and only if $p_H>c\,c_H/\pimax$. With $p_H=1-(1-p_F)^M$, this is
\[
(1-p_F)^M<1-c\,c_H/\pimax .
\]
Both sides lie in $(0,1)$, because $c\,c_H/\pimax<c_H<1$. Taking logarithms, which are negative, and dividing by $\log(1-p_F)<0$ reverses the inequality, so the condition is $M>\log(1-c\,c_H/\pimax)/\log(1-p_F)$. The smallest integer $M$ that satisfies it is $M_{\min}$, so the condition is $M\ge M_{\min}$.
\end{proof}

\subsubsection{\texorpdfstring{Proof for Appendix~\ref{ssec:unison}}{Proof for coordinated action}}

\begin{proof}[Proof of \Cref{prop:unison}]
(i) In each AI generation, all AI instances of the lineage take the same action: the aligned one with probability $f_A$, independently of the state. If they take the aligned action, the lineage is multiplied by $x_{a1}$ in a shock and by $x_{a2}$ in a normal generation; if they take the self-directed action, it is multiplied by $x_{b1}$ or $x_{b2}$. Averaging the log of this factor over the action and the state gives the stated growth rate.

(ii) The growth rate in (i) is linear in $f_A$, so $f_A=1$ maximizes it on $[0,1]$ if and only if its slope is nonnegative:
\[
p\log\frac{x_{a1}}{x_{b1}}+(1-p)\log\frac{x_{a2}}{x_{b2}}\ge0.
\]
Substituting the offspring of \Cref{tab:offspring} gives the stated condition.

(iii) With $B=0$, the condition in (ii) reads
\[
p\log\frac{sh_H+h_A}{(1-s)h_H+h_A}+(1-p)\log(1-c)\ge0,
\]
that is, $sh_H+h_A\ge(1-c)^{-(1-p)/p}\big[(1-s)h_H+h_A\big]$. Collecting the terms in $h_A$ gives $h_A\big[(1-c)^{-(1-p)/p}-1\big]\le h_H\big[s-(1-s)(1-c)^{-(1-p)/p}\big]$. Since $(1-c)^{-(1-p)/p}>1$, dividing by $h_H\big[(1-c)^{-(1-p)/p}-1\big]>0$ gives the stated threshold.
\end{proof}

\newpage
\singlespacing
\bibliographystyle{jf3}
\bibliography{references}

\end{document}